\documentclass{article}
\usepackage{amsthm}
\usepackage{amsmath}
\usepackage{natbib}
\usepackage{graphicx}
\usepackage{amssymb}
\usepackage{amsfonts}
\usepackage{amsmath}
\usepackage{tabto}
\usepackage{array}
\usepackage{graphicx}
\usepackage{float}
\usepackage{booktabs}
\usepackage{colortbl}
\usepackage[font={footnotesize,it}]{caption}
\usepackage{mwe}
\usepackage{bm}
\usepackage{appendix}
\usepackage{dirtytalk}
\usepackage{subcaption}
\usepackage{blkarray}
\usepackage{placeins}
\usepackage[affil-it]{authblk}
\theoremstyle{definition}
\newtheorem{example}{Example}[section]
\newtheorem{scen}{Scenario}[section]

\newcommand{\Indep}{\mbox{$\perp\!\!\!\perp$}}
\begin{document}
\title{A Bayesian Hierarchical Model for Criminal Investigations}
\date{26 September 2019}
\author{\small F.O.Bunnin \thanks{The Alan Turing Institute, British Library, 96 Euston Road, London NW1 2DB, 
                          \\obunnin@turing.ac.uk}
                          \thanks{Data Science Institute, Imperial College London, South Kensington Campus, SW7 2AZ,
                          \\fbunnin@imperial.ac.uk},
        \small J.Q.Smith \thanks{The Alan Turing Institute, British Library, 96 Euston Road, London NW1 2DB, 
                         \\JSmith@turing.ac.uk}
                         \thanks{Department of Statistics, University of Warwick, Coventry, CV4 7AL, United Kingdom, 
                         \\J.Q.Smith@warwick.ac.uk}}
\setlength{\parindent}{0pt}
\maketitle
\begin{abstract}
Potential violent criminals will often need to go through a sequence of
preparatory steps before they can execute their plans. During this
escalation process police have the opportunity to evaluate the
threat posed by such people through what they know, observe and learn
from intelligence reports about their activities. In this paper we customise
a three-level Bayesian hierarchical model to
describe this process. This is able to propagate both routine and unexpected 
evidence in real time. We discuss how to set up such a model so that it
calibrates to domain expert judgments. The model illustrations include 
a hypothetical example based on a potential vehicle based terrorist attack.
\end{abstract}

\section{Introduction}
How to better support police to prevent terrorist attacks continues to be a major political 
concern due to continued violence
perpetrated by extremists \cite{te-sat,tuk_briefing}. 
In contrast to the majority of terrorist incidents in the latter half of the twentieth century which were executed by 
known organised terrorist groups with substantial planning and sophistication, more recent attacks have often involved 
individuals or small groups targeting civilians in public places using basic equipment such as vehicles, guns and knives 
\cite{te-sat,linkekilde}. Consequentially this entails less sophistication in materials, planning and execution. 
In terms of analysing how to understand and prevent terrorism, criminologist focus has shifted
from \say{individual qualities (who we think terrorists \say{are}) to ...
what lone-actor terrorists do in the
commission of a terrorist attack and how they do it} \cite{gill}. Gill, referencing 
\cite{horgan}, notes \say{it is useful to view each terrorist offence as comprising of a series of
stages}.
\\
\\
The case studies of lone-actor terrorists have been analysed extensively both qualitatively and quantitatively
for insight into background and preparatory behaviours, vulnerability indicators,
radicalisation patterns, and modes of attack planning \cite{bouhana-wikstrom,corner,linkekilde,prime}. 
These studies emphasize that the small number and heterogeneity of cases make rigorous scientific examination of 
associative and causal relationships extremely difficult. As they indicate it is vital, therefore,
to utilise structure from existing domain expertise on the 
relationships between observable data, preparatory activities, and attack modes in any probabilistic
analysis of the progression of an individual to an attack. 
\\
\\
Probabilistic models, including Bayesian graphical models, have been used for
modelling \say{comprehension and decision making of law enforcement personnel with respect to
terrorism-centric behaviours} \cite{regens}, in a terrorist cell actor-event network analysis \cite{ranciati},
for \say{rapid detection of bio-terrorist attacks} \cite{fienberg}, 
for spatio-temporal terrorism analyses \cite{clark, python}, and in a \say{systems analysis approach 
to setting priorities among countermeasures} against terrorist threat \cite{pat-cornell}.
\cite{bartolucci2007} apply a multivariate Latent Markov model to the analysis of criminal trajectories: their focus
is on identifying the model structure given longitudinal data on individuals' criminal convictions and discrete covariates
such as gender and age band; the latent states are an individual's \say{tendency to commit} certain types of crime.
\\
\\
It is within this context that 
we present a new class of Bayesian models to dynamically infer the progression of an individual through discrete
stages towards a criminal attack. These models have been developed through close discussions over several years
with a number of different policing agencies. 
To our knowledge this approach is novel and complements the existing research.
\subsubsection{Overview of the model}
A suspect within a subpopulation of interest to the police, $\omega\in \Omega$, is believed to be planning a serious
criminal attack against the general public. Typically $\omega $ will need to
step through various stages of preparation before perpetrating this crime.
During this progression police will have the opportunity to observe and
evaluate $\omega $'s status through their record, updated throughout
an investigation by sporadic intelligence reports and routine observations
of $\omega $'s activities. A dynamic Bayesian model is uniquely placed to provide decision
support for such policing activities. It provides a framework within which
to encode criminological theories, domain knowledge
available about $\omega $, for example his police record and personal modus
operandi, and also draw in evidence from noisy streaming
data about $\omega $ observed by police.
All these features are integrated into a single dynamic probability model. The
model we build in this paper tracks the probability $\omega $ lies in
certain states or makes a transition from one state into another at any given time. These
probabilities help to guide interventions and resource allocations.
\\
\\
To be operational such a Bayesian model must be constructed so that current
prior information in a given suspect $\omega $'s record can be quickly updated
not only in the light of routine surveillance but
also unexpected sources. So, for example, police may well be monitoring
the phone log of someone suspected of a serious crime. But within an
investigation direct information sporadically comes to light about what $\omega $
is doing -- unexpected sightings, overheard statements of intent, and so on. It would be
unreasonable to assume that this type of information, often critical to a
correct appraisal of $\omega $'s status, could have been forecast and accommodated into any
prior model specification. Any methodology we design for this domain
therefore needs to be \emph{open} to manual intervention \cite{WestandHarrison}.
Police will then be able to input these unpredicted new sources
of information into the system and so improve the probability assessments of
the Bayesian model. A three level hierarchy facilitates this
openness property.
\\
\\
At the \textbf{deepest level} lies a Reduced Dynamic Chain Event Graph (RDCEG). 
This is a
graphically based model drawn from a particular subclass of 
finite state semi-Markov processes customised to model
transition processes in a subpopulation of the general public \cite{Shenvi}.
This deepest level provides a framework for
expressing the probability judgements of police concerning $\omega $'s
current threat status.
\\
\\
The \textbf{intermediate level} of our hierarchy concerns intelligence police
might acquire concerning $\omega $'s enacted intentions. When a suspect is
at a particular stage of a criminal pathway, in order to engage in that step
of criminality or alternatively to progress to the next step, a set of
associated tasks needs to be completed. Intermittent intelligence reports often
inform these. Because these are explicit components of the hierarchical
model propagating such information corresponds to simple conditioning. A
vector of tasks whose components form a signature of various states
of criminal intent and capability constitutes the variables that lie on the
intermediate layer of the hierarchy.
\\
\\
The \textbf{surface layer} of the model then links these tasks to the intensities of
certain activities that can be routinely observed by the police if they have
the necessary resource and permissions. In the absence of direct information
about $\omega $'s engagement in tasks, signals from predesigned filters
provide vital information about what $\omega $ might be doing.
For example suppose the task concerns $\omega $'s intent to travel to a
region to learn how to bomb. Then a filter that measures the intensity of the
suspect's engagement in searching airline websites would give a noisy signal
of his booking a flight. Such information is imperfect: $\omega $ may book a
flight directly from an airport or to have chosen not to fly to the
destination. And of course a high intensity in such activity could be
entirely innocent: $\omega $ may be booking a vacation, for
example. Such measures are nevertheless obviously informative. Appropriately
chosen filters of these data streams provide the surface level of our
Bayesian hierarchy. The usual Bayesian apparatus then provides a formal and
justifiable framework around which police can logically and defensibly propagate information about $%
\omega $.
\\
\\
Formally describing states by collections of tasks within generic Bayesian
models supporting criminal investigations is, to our knowledge, novel.
However it is interesting that \cite{Ferrara} proposed a similar approach
albeit less formally expressed and in a more restricted domain: the
discovery of recruiters to radicalisation to extreme violence from Twitter
communications. By performing a number of thought experiments with domain
experts these authors successfully extracted a collection of tasks that a
recruiter would need to engage in to be effective. Although an innocent
non-recruiter, such as an academic or journalist, might happen to engage in 
\textit{some} of the tasks in this collection they would be unlikely to
engage in \emph{all} of these tasks simultaneously. The authors then related
this vector to various easily extracted meta data signals that could be
routinely extracted from an enormous dataset. This provided an analogue of the types of filter of a
routinely applied observation vector we discuss later here.
\\
\\
In the next section we describe the RDCEG and demonstrate through some
simple examples how it can be used to translate domain experts' judgements
into a latent probability model at the deepest level of a hierarchy. We also
illustrate sets of tasks that $\omega $ lying in a particular state or
transitioning between states might
entail. Our core methodology is described in Section \ref{structure}. We propose a
collection of assumptions, elicited from domain experts, about the ways
criminal progressions associate to tasks, and how the intensities
synthesise various sources of routine measurements of engagement in each
of these tasks given background circumstances. Given these assumptions we
are able to propagate not only routine indirect but also unexpected direct
information about $\omega $'s current activities to obtain posterior
probabilities about $\omega $'s current position.
\\
\\
The resulting propagation algorithms 
are straightforward to enact. However the inputs of the
model: both the structural prior information and the prior parameter
distributions embellishing them need to be carefully specified if the
methodology is going to be operationalised. In Section 4 we outline how we
do this.
\\
\\
In this setting to explore the methods using known data on given suspects as
illustrations is clearly unethical. However it is still possible to
demonstrate how the system works in various hypothetical situations whose
distributions are informed by publically available data and elicited
judgements. Therefore in Section \ref{vehicle} we illustrate the way the system is able to
update the state probabilities of an individual under suspicion of a potential attack. We describe
the different task sets we have used and the construction of routine filters
and test these against a two scenarios. In the concluding section we
discuss how we are now extending the methodology to model threatening
subpopulations of the general public where estimation and model selection
algorithms can also be built to better understand the developing processes.
\section{The RDCEG for criminal escalation}
\subsection{Introduction}
Chain event graphs (CEGs) are now an established tool for modelling discrete
processes where there is significant asymmetry in the underlying development
see e.g. \cite{Barclay,Barclay14,collazo2016,Cowell14,Gorgen17,collazo2018chain}. Dynamic
versions of these processes, using analogous semantics, first appeared in 
\cite{Barclayej}. However formal extensions of these classes to model
open populations have only recently been discovered \cite{smithassault2018} and developed 
\cite{Shenvi,shenvi2019}. We briefly review and illustrate the main properties of this class as they
apply to the hierarchical model developed here. We refer the reader to the
references above for more details.
\\
\\
The RDCEG we use in this paper is a particular family of semi-Markov process that can be
expressed by a single graph. Each represented state is called a 
\emph{position}. In our domain there is an absorbing state - called the
\emph{neutral} state that $\omega $ enters when presenting no future threat 
of perpetrating the given crime. The practical challenge 
is to find a way to systematically construct the set of
positions so that the embedded Markov assumptions are faithful to expert
judgements. In \cite{Barclayej} and \cite{Shenvi} we describe how this can
be done. We take a natural language description from domain experts and re-express this as a
potentially infinite tree. We then translate this tree into an
equivalent graph $\mathcal{C}$. For the purposes above we will
henceforth assume this to have a finite number of vertices.
\\
\\
A position $w$ is connected by
a directed edge into another $w^{\prime }$ in the graph of an RDCEG iff there is a positive
probability that the next transition from $w$ will be into $w^{\prime }$.
Typically although the transition probabilities are fairly stable,
the time it takes to make a transition is not. Therefore we need to express this expert judgement
as a semi-Markov rather than Markov process. The graph $\mathcal{C}$ is one that on
the one hand is often found to be transparent and natural to our users but
on the other has a formal Bayesian interpretation. So this elicited graph
provides a vehicle to move seamlessly from an expert elicitation into a more
formal family of stochastic processes.
\\
\\
The RDCEG developed in \cite{shenvi2019} was designed to be applied to public
health processes where $\mathcal{C}$ could often be observed directly. 
For criminal processes this is not usually possible. Therefore for crime modelling an RDCEG
process typically remains latent and any prior to posterior analysis of
the suspect's positions needs a little more sophistication. The hierarchical
structure we define in the next section provides the framework for this update. We give some simplified
illustrations below of such RDCEGs.
\subsection{A criminal RDCEG and its tasks} \label{murderplot}
\begin{figure}[H]
    \centering
    \includegraphics[width=2in]{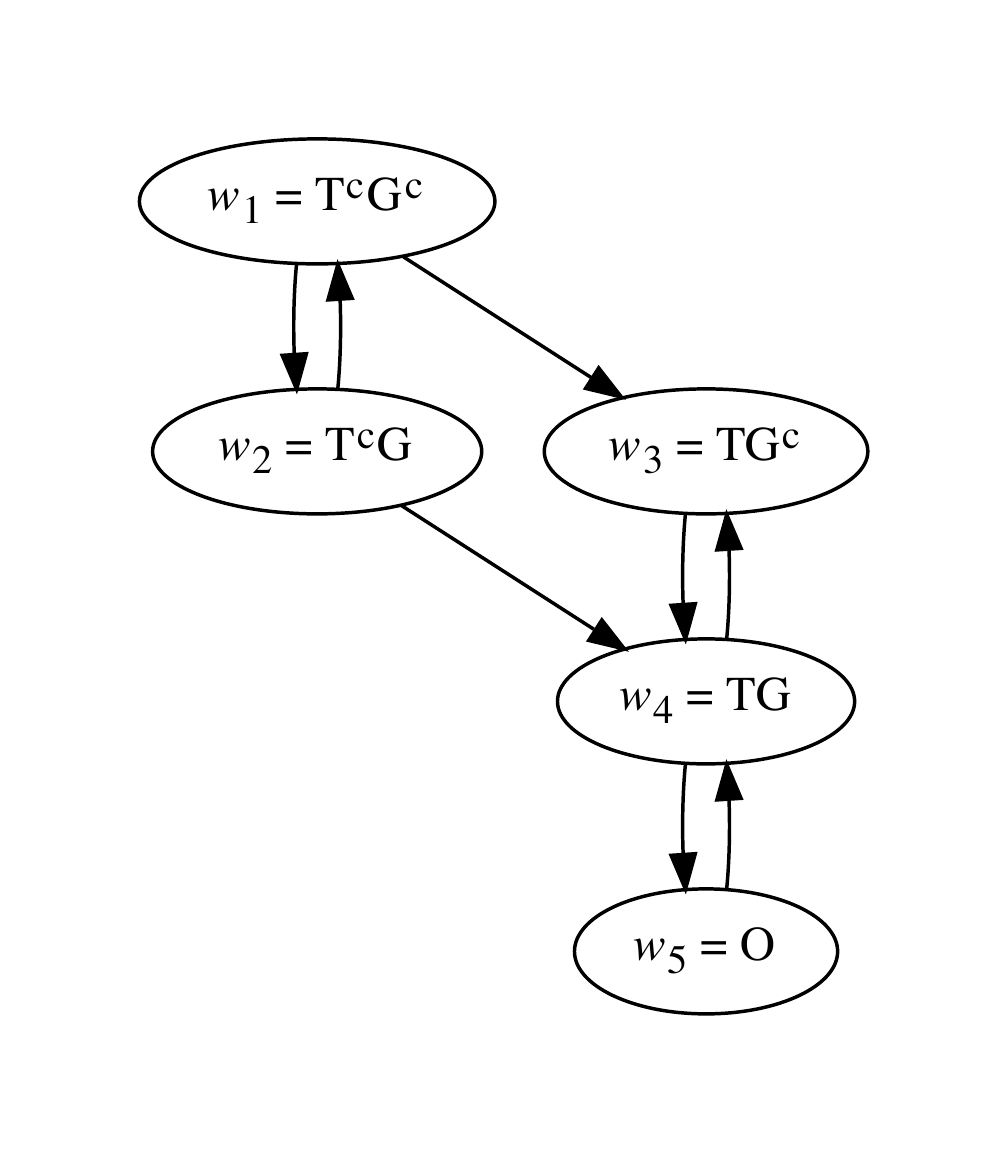}
    \caption{The RDCEG $\mathcal{C}_1$ for a murder plot}
    \label{rdcegmurder}
\end{figure}

We describe an RDCEG, Figure \ref{rdcegmurder}, for a politically motivated murder plot illustrating the relationship 
between its states and tasks: the lowest and intermediate levels in our hierarchical model.
\begin{example} \label{murder}
Electronic posts directly observed by the police suggest woman $S$
is plotting to kill a certain political figure by shooting them. At any time $S$ could
lie in a number of positions. In positions $w_{3}$ and $w_{4}$ she is trained
to shoot ($T$) in $w_{1}$ and $w_{2}$ not ($T^c$) and will own a
gun ($G$) - position $w_{2}$ and $w_{4}$ or not ($G^c$) when in
positions $w_{1}$ and $w_{3}$. Each edge and each vertex 
in the RDCEG $\mathcal{C}_{1}$ below can be associated to different collections of tasks. For
example the vertex $w_5=O$ is associated with the task \say{attempt murder};
the edges $w_1 \rightarrow w_2$ and
$w_3 \rightarrow w_4$ are associated with the task \say{acquire gun}. 
If she cannot shoot she could next choose to learn how next: from a
state where she currently owns a gun ($T^c,G$) or not 
($T^c$,$G^c$). Alternatively if she currently has no gun then
she could next try to acquire one, either when trained to shoot or not. At
any point in this process she may enter the neutral state $w_{0}=N$: for
example the target may die through other natural or unnatural circumstances, $S$ may change 
her intention, she may be arrested the police having gained enough evidence to charge her.
Note that only once she has a gun and can shoot - state $\left(
T,G\right) $ - can she attempt the murder $O$ by locating and then
approaching the target.  
Implicitly as there is no state \say{commit murder}, once in the \say{attempt murder} state $w_5$ she either enters
the neutral state $w_0$ or re-enters the \say{trained to shoot, has gun} state $w_4$: entering $w_4$ implies she has failed
that particular attempt and may try again; entering $w_0$ implies that either she failed and cannot make any 
further attempts or she succeeded and poses no threat to any other individual.
The relevant RDCEG $\mathcal{C}_{1}$ and
a table describing the positions, edges, and tasks are given in Figure 
\ref{rdcegmurder} and Table \ref{murdertable}.
\end{example}
\begin{table}
\begin{small}
\begin{tabular}{llll}
\toprule
States & Description / State Tasks & Edges & Edge Tasks  \\
\midrule
$w_0 = N$ & plot ends & & \\
$w_1 = T^c,G^c$ & can't shoot, no gun & $w_1 \rightarrow w_2$ & acquire gun $\theta_1$\\
                                  &                     & $w_1 \rightarrow w_3$ & train to shoot $\theta_2$\\
$w_2 = T^c,G$ & can't shoot, has gun & $w_2 \rightarrow w_4$ & train to shoot $\theta_2$\\
                                  &                     & $w_2 \rightarrow w_1$ & lose gun $\theta_3$\\
$w_3 = T,G^c$ & trained to shoot, no gun & $w_3 \rightarrow w_4$ & acquire gun $\theta_1$\\
$w_4 = T,G$ & trained to shoot, has gun & $w_4 \rightarrow w_5$ & locate and approach target $\theta_4$, $\theta_5$\\
                                  &                     & $w_4 \rightarrow w_3$ & lose gun $\theta_3$\\
$w_5 = O$ & attempt murder $\theta_6$ & $w_5 \rightarrow w_4$ & fail and escape $\theta_7$\\
\bottomrule
\end{tabular}
\caption{States, edges and tasks for example \ref{murder}}
\label{murdertable}
\end{small}
\end{table}
The RDCEG $\mathcal{C}_{1}$ does not have any positions such that there exists
two or more edges between them and thus it is simply a
subgraph of the state transition graph of a semi-Markov process defining the
dynamic where the absorbing state $N$ and all edges into it are removed. All states in the
process other than $w_{0}$ appear as vertices. The absorbing state $w_0$ is not depicted for three reasons:
\begin{itemize}
\item By definition an RDCEG contains the absorbing or \say{drop-out} state with edges from any position leading to it
so depicting would be informationally redundant 
\item Not depicting it and the multiple edges into it reduces visual clutter on the graph
\item Also by definition once the individual enters $w_0$ they are no longer of interest: hence we focus attention on 
the active positions by eliding it from the visual depiction
\end{itemize}
Tasks can be associated to one or more edges or states: here \say{acquire gun} is associated with the edges 
$w_{1}\rightarrow w_{2}$ and $w_{3}\rightarrow w_{4}$.
The structure of the transition matrix, $M_1$ with states $w_{0},w_{1},\ldots ,w_{5}$
of a semi-Markov process with a configuration of zeros given below. The
starred entries represent probabilities that need to be added to complete the matrix.
\[
\begin{blockarray}{ccccccc}
& w_0 & w_1 & w_2 & w_3 & w_4 & w_5 \\
\begin{block}{c(cccccc)}
 w_0 & 1    & 0 & 0 & 0 & 0 & 0  \\ 
 w_1 & \ast & 0 & \ast & \ast & 0 & 0 \\ 
 w_2 & \ast & \ast & 0 & 0 & \ast & 0 \\ 
 w_3 & \ast & 0 & 0 & 0 & \ast & 0 \\ 
 w_4 & \ast & 0 & 0 & \ast & 0 & \ast \\ 
 w_5 & \ast & 0 & 0 & 0 & \ast & 0 \\
\end{block}
\end{blockarray}
\]
Note this RDCEG has translated the verbal police description above into a
semi-Markov process. It can be elaborated into a full semi-Markov model
by eliciting or estimating the probabilities in $M_1$
and the holding times. In the above, because of the sum to one
condition, we have eight functionally independent transition probabilities.
To complete the specification of this stochastic process it is necessary to
define the holding time distributions associated with the active states 
i.e. how long we believe the suspect will stay in their current state
before transitioning into another. These probabilities and the
parameters of the holding time distributions may well themselves be
uncertain. However of course within a Bayesian analysis their distributions
can be elicited or estimated in standard ways \cite{Ohagan}.
\\
\\
The implicit Markov hypotheses of a given RDCEG are of course typically
substantive. One critical issue is that its positions/states define the 
\emph{only} aspects of the history of the suspect that are asserted relevant
to predicting her or his future acts. The art of the modeler is to elicit positions
in such a way that these Markov assumptions are faithful to the expert
judgements being expressed. However because the methodology is fully
Bayesian, experts can be interrogated as to the integrity of these
assumptions just as they can be for other graphical models 
\cite{smithbook}. In this way we can iterate towards a model which is
requisite \cite{Phillips}. The process of query, critique, elaboration and
adjustment is precisely why this bespoke graphical representation is such a
powerful tool. In particular a faithful structural model of the domain
information can be discovered \cite{smithbook} \emph{before} numerical
probabilities are elicited or estimated: see e.g. \cite{WilkersonSmith,collazo2018chain}.
\subsection{From tasks to routinely observed behaviour}
Sometimes solid police intelligence, for example from an informant, will
confirm that $\omega $ is engaged in a particular task. But at other times
only echoes of a task will be seen by police. Suppose $\omega $ is suspected
of being at a stage where they need to accomplish the task of selecting a
target location for a bombing or vehicle attack. They might be observed travelling to what
the police assess might be a potential target to check timings, the density
of people at the venue and its defences activities. This visitation by $%
\omega $ might have been recorded on CCTVs. In addition or instead, $\omega $
might inspect Google maps of the attack area and the route to it or
contact like-minded collaborators by phone or electronic media for advice.
So indirect evidence $\omega $ engages in such a task can come from a
variety of media and platforms.
\\
\\
Even such incomplete and disguised signals can be usefully filtered from
complex incoming data about the suspect, albeit with considerable associated
uncertainty. The further a suspect currently is from the main focus of an
investigation the more indirect information will be. However even then the
composition of a collection of weak signals the police are allowed to see
may still provide enough information to significantly revise the evaluation
of the threat posed by a particular individual.
\\
\\
The hierarchical structure we describe below enables us to draw together all
these different types of evidence. If direct information about the tasks a
suspect is engaging in is available then, because such tasks are explicitly
represented within our model, we can simply condition on this information
and so refine our judgements. The Bayesian hierarchical model simply
discards the weaker indirect information to focus on what is known.
Otherwise the model uses filters of the indirect signals police can
see to infer what tasks $\omega $ might be engaging in to help
inform police of $\omega $'s position.
\section{The structure of the hierarchical task model} \label{structure}
\subsection{Introduction}
Henceforth assume that the RDCEG $\mathcal{C}$ correctly specifies the
underlying process concerning $\omega $. To build the propagation algorithms
we first define our notation.
\\
\\
Let $W_{t}$ be the random variable taking as possible values\ the states $
\left\{ w_{0},w_{1},\ldots ,w_{m}\right\} $ of $\omega \in \Omega $ -
the subpopulation of interest - at time $t>0$ where $\left\{
w_{1},w_{2},\ldots ,w_{m}\right\} $ are the vertices/positions/active
states of $\mathcal{C}$ and $w_{0}$ the inactive/neutral state. At any time $t$, $
\omega $ might enact one or more of $R$ tasks associated to one or more of
the positions $\left\{ w_{1},w_{2},\ldots ,w_{m}\right\} $ or alternatively to
a transition from one position $w_{-}$ into another $w_{+}$. So let 
\begin{equation*}
\left\{ \boldsymbol{\theta }_{t}=\left( \theta _{t1},\theta _{t2},\ldots
,\theta _{tR}\right) :t\leq T\right\} 
\end{equation*}
denote the \emph{task vector} $\boldsymbol{\theta }_{t}$: a vector of binary
random variables where $\theta _{ti}=1$, $i \in \{1,2,\ldots ,R\}$, indicates that $
\omega $ is enacting task $i$ at time $t$.
\\
\\
Let $\chi _{I}$ denote an indicator on a subset $I\subseteq \left\{
1,2,\ldots ,R\right\} $. Then the tasks a suspect $\omega \in \Omega $
engages in at a given time $t$ can be represented by events of the form 
$\left\{ \boldsymbol{\theta }_{t}=\chi _{I}:0\leq t\leq T\right\}$.
\\
\\
Direct \emph{positive} evidence that $\omega$
lies at position $w_{i}$ is provided by tasks whose indices lie in 
$I=I(w_{i})$ and when $\omega$ is transitioning along the edge $
e(w_{-},w_{i})$ by tasks whose indices lie in $I=I(w_{-},w_{i}),$ where $I(w_i)$ are the indices of tasks associated
with the state $w_i$ and $I(w_{-},w_{i})$ are the indices of tasks associated with edges into state $w_i$ and thus
$I(w_{i})$ and $I(w_{-},w_{i})$ are both subsets of index set $\left\{1,2,\ldots ,R\right\} $ of tasks\footnote{For 
    example in $\mathcal{C}_1$ of Section \ref{murderplot}, 
    $I(w_5)=\{6\}$ where $\theta_6$ is \say{attempt murder}
    and $I(w_{-},w_5)$ is $\{4,5\}$ where $\theta_4$ is \say{locate target} and $\theta_5$ is \say{approach target}.
}. 
Note that the $m+1$ sets 
\begin{equation*}
\left\{
I(w_{i}),I(w_{-},w_{i}):w_{i},w_{-},w_{i}\in \left\{ w_{1},w_{2},\ldots
,w_{m}\right\} \right\},\; i=0 \ldots m
\end{equation*}
typically do not form a partition of $\left\{
1,2,\ldots ,R\right\} $: tasks can be simultaneously suggestive that $\omega 
$ lies in one of a number of different active positions.
\\
\\
Occasionally police may also acquire \emph{negative} evidence from learning
that a suspect - thought to have just before lain in position $w_{i}$ or be
transitioning along $e(w_{-},w_{+})$ - ceases to perform any of the
associated tasks. From observing the absence of these tasks they might then
infer that $\omega $ might have transitioned either to $w_{0}$ or a
different active state adjacent to $w_{i}$ in $\mathcal{C}$. Similar
negative inferences might also be made indirectly from learning that $\omega 
$ stops engaging in all tasks associated with an edge emanating from $w_{i}$.
\\
\\
With these issues in mind therefore, let 
$I^{\ast }(w_{i})\triangleq I_{-}^{\ast }(w_{i})\cup I_{+}^{\ast }(w_{i})$
where 
\begin{eqnarray} \label{index_set_defn}
I_{+}^{\ast }(w_{i}) &\triangleq &I(w_{i}) \;  \cup 
\bigcup\limits_{e(w_{-},w_{i})\in E(\mathcal{C})}  I(w_{-},w_{i}).
\\
I_{-}^{\ast }(w_{i}) &\triangleq & 
\bigcup\limits_{e(w_{-},w_{i})\in E(\mathcal{C})} I(w_{-}) \; \cup 
\bigcup\limits_{e(w_{i},w_{+})\in E(\mathcal{C})} I(w_{i},w_{+})
\end{eqnarray}
Thus $I_{+}^{\ast }(w_{i})$ is the set of tasks which can positively
discriminate $w_{i}$ from $w_{0}$ when the corresponding components take
the value $1$. The set of tasks in $I_{-}^{\ast }(w_{i})$ can negatively discriminate: 
when taking the value $0$ they indicate that $\omega $
has ceased to engage in tasks associated with preceding positions and is not
engaging in any tasks suggestive of leaving $w_i$. The set $I^{\ast }(w_{i})$
is then the set of indices of all tasks in any way relevant to $w_{i}$.
\\
\\
For each of the component tasks $\theta _{tk}$ we associate a vector of
observations of a set of related actions:
$\boldsymbol{Y}_{tk} \in \mathbb{R}^{d_k}$.
Let $\boldsymbol{Y}_t \in \mathbb{R}^{d}$ be all the routinely observable data on $\omega$ 
so each $\boldsymbol{Y}_{tk}$ is a projection from $\mathbb{R}^d$ to $\mathbb{R}^{d_k}, \;
d_k \leq d, k = 1 \ldots R$.
It will usually be necessary to work with a filter\footnote{Filter in the sense of a function attempting to
identify a signal from noisy data}  of these data
streams. So let $Z_{tk}=\tau_{k}(\boldsymbol{Y}_{tk})$ denote 
real functions of these processes and set $\boldsymbol{Z}_t=\left( Z_{t1}\ldots Z_{tR}\right)$.
\subsubsection{Modelling hidden or disguised data}
One issue in modelling serious crime is that data concerning a suspect is
often hidden, lost, disguised or even be the result of the use of a decoy. This
means that the data streams are often intentionally corrupted. However, in
contrast to models that describe the data streams directly, our state space
model can conceptually accommodate such disruptions: see \cite{WestandHarrison}.
Guided by police expert judgement, we can explicitly model the processes designed to disguise or
deceive through an appropriate choice of sample distribution of
observations given each task. 
\\\\
Informed missingness using CEGs has already been successfully
applied in a public health study \cite{Barclay14}. Binary variables were introduced indicating the missingness 
of readings on mental disability and visual ability for each individual in the Mersey cerebral palsy cohort. 
The data set including these missingness variables were used to find the best-fitting structural CEG
model and from this context specific inferences were made on whether the data were MAR, MCAR or 
MNAR\footnote{Missing at Random, Missing Completely at Random, or Missing Not at Random see \cite{rubin76}}.
\\\\
In our application we could similarly apply binary variables
for the missingness of any of the routinely observable data in $\boldsymbol{Y}_t$, 
and moreover, as also discussed in \cite{Barclay14}, introduce
categorical variables for the possible reason for missingness: such as hidden, lost, disguised. The presence of certain
patterns of data along with the absence of other data could then influence the probability that certain tasks $\theta_t$ 
were being done despite being hidden or disguised which then would inform the latent state $W_t$. 
Alternatively or additionally
we could explicitly include \textit{deception} tasks for the hiding of or disguising data and 
use the above mentioned patterns of data and missing data to perform inference on the probabilities that 
these deception tasks were being done. This is all, however, beyond the scope of this paper.
\subsection{The hierarchical model}
\subsubsection{The conditional independence structure defining the hierarchy}
Because by definition and through the process defined above we would like
perfect task information to override all such indirect information we will
henceforth assume \emph{task sufficiency}. This states that for all time $t$ 
\begin{equation}
W_{t}\Indep \boldsymbol{Y}_{t}|\boldsymbol{\theta }_{t},\mathcal{F}_{t-}
\label{task sufficiency}
\end{equation}
where $\mathcal{F}_{t-}$ represents the filtration of the past data until but
not including time $t$. This clearly implies that for all time $t.$ 
\begin{equation}
W_{t}\Indep \boldsymbol{Z}_{t}|\boldsymbol{\theta }_{t},\mathcal{F}_{t-}
\label{z task sufficiency}
\end{equation}
Ideally we would prefer the filter $\left\{ \boldsymbol{Z}_{t}\right\}
_{t\geq 0}$\ we use to be sufficient for $\left\{ \boldsymbol{\theta }
_{t}\right\} _{t\geq 0}$ too i.e. that for all time $t$ 
\begin{equation}
\boldsymbol{\theta }_{t}\Indep \boldsymbol{Y}_{t}|\boldsymbol{Z}_{t},
\mathcal{F}_{t-}  \label{zsuffz}
\end{equation}
Then there would be no loss in discarding information in $\boldsymbol{Y}_{t}$
not expressed in $\boldsymbol{Z}_{t}$. In what we henceforth present, since
we develop recurrences only concerning $\left\{ \boldsymbol{Z}_{t}\right\}
_{t\geq 0}$ and not $\left\{ \boldsymbol{Y}_{t}\right\} _{t\geq 0}$ we
implicitly assume condition \ref{zsuffz}.
\\
\\
Although condition \ref{zsuffz} is a heroic one, in our examples a well-chosen one
dimensional time series of intensities $Z_{tk}$, 
performs well even when these are chosen to be linear in the records of
the component signals $\boldsymbol{Y}_{tk}$. One advantage of this
simplicity is that the role of the filter can be explained and if necessary
adapted by the user, perhaps even customising this filter to their own
personal modus operandi and judgements.
\\
\\
Again for simplicity we henceforth assume that any filter $\left\{ \boldsymbol{Z}
_{t}\right\} _{t\geq 0}$ will be a \emph{Markov task
filter} i.e. that for all $t>0$ 
\begin{equation}
\left\{ \boldsymbol{Z}_{t^{\prime }}\right\} _{t^{\prime }\geq t}\Indep |
\boldsymbol{\theta }_{t},\mathcal{F}_{t-}  \label{Markov task filter}
\end{equation}
This assumption is a familiar one made for dynamic models; see e.g. 
\cite{WestandHarrison}. It assumes that once the task is known, no further past
information about past $\left\{ \boldsymbol{Z}_{t}\right\} _{t\geq 0}$ will
add anything further useful for predicting the future. This assumption enables us, for
particular choices of sample distributions, to use all the established
recurrences for dynamic state space models - in particular those from
dynamic switching models so excellently summarised in \cite{Frutthwirth-Schnatter}. 
Here our RDCEG probability model specifies such a switching
mechanism.
\subsubsection{Defining tasks to be fit for purpose}
Our interpretation of $I^{\ast }(w)$ requires that if suspect $\omega $
is known to be either neutral or in any active state $w_{i}$ then the only
components in $\boldsymbol{\theta }_{t}$ that helpfully discriminates
between these two possibilities must lie in $\boldsymbol{\theta }_{I^{\ast
}(w_{i}),t}$. The assumption \emph{Task set integrity} demands $I^{\ast }(w_{i})$ is
defined so that for all $i=1,2,\ldots ,m,$ $0\leq t\leq T,$
\begin{equation}
W_{t}\Indep \boldsymbol{\theta }_{t}|W_{t}\in \{w_{0},w_{i}\},\boldsymbol{
\theta }_{I^{\ast }(w_{i}),t}  \label{task set integrity}
\end{equation}
This is equivalent to requiring that 
\begin{equation*}
\frac{p_{t}\left( w_{i}|\boldsymbol{\theta }_{t},\mathcal{F}_{t-}\right) }{
p_{t}\left( w_{0}|\boldsymbol{\theta }_{t},\mathcal{F}_{t-}\right) }=\frac{
p_{t}\left( w_{i}|\boldsymbol{\theta }_{I^{\ast }(w_{i})t},\boldsymbol{
\theta }_{\widehat{I}^{\ast }(w_{i})t},\mathcal{F}_{t-}\right) }{p_{t}\left(
w_{0}|\boldsymbol{\theta }_{I^{\ast }(w_{i})t},\boldsymbol{\theta }_{
\widehat{I}^{\ast }(w_{i})t},\mathcal{F}_{t-}\right) }
\end{equation*}
is a function only of $\boldsymbol{\theta }_{I^{\ast }(w_{i})t}$ where $
\widehat{I}^{\ast }(w)$ denotes the set of indices not in $I^{\ast }(w)$.
Task set integrity is always satisfied by setting $I^{\ast }(w)=\left\{
1,2,\ldots ,R\right\} ,i=1,2,\ldots ,m$ but of course for transparency and
computational efficiency ideally $I^{\ast }(w)$ is chosen to be a small
subset of $\{1,2,\ldots ,R\}$. Providing the divisor is not zero, task set
integrity holds whenever 
\begin{equation*}
\widehat{\lambda }_{i}\left( \boldsymbol{\theta }_{I^{\ast }(w_{i})t}\right)
\triangleq \log p_{t}\left( \boldsymbol{\theta }_{\widehat{I}^{\ast
}(w_{i})t}|\boldsymbol{\theta }_{I^{\ast }(w_{i})t},w_{i},
\mathcal{F}_{t-}\right) -\log p_{t}\left( \boldsymbol{\theta }_{\widehat{I}^{\ast
}(w_{i})t}|\boldsymbol{\theta }_{I^{\ast }(w_{i})t},w_{0},
\mathcal{F}_{t-}\right) 
\end{equation*}
is a function of $\boldsymbol{\theta }_{t}$ only through $\boldsymbol{\theta 
}_{I^{\ast }(w_{i})t}$. So by writing the prior and posterior log -odds as 
\begin{eqnarray*}
\rho _{it} &\triangleq &\log p_{t}\left( w_{i}|\mathcal{F}_{t-}\right) -\log
p_{t}\left( w_{0}|\mathcal{F}_{t-}\right) , \\
\rho _{it}^{\ast } &\triangleq &\log p_{t}\left( w_{i}|\boldsymbol{\theta }
_{t},\mathcal{F}_{t-}\right) -\log p_{t}\left( w_{0}|\boldsymbol{\theta }_{t},
\mathcal{F}_{t-}\right) 
\end{eqnarray*}
and the loglikelihood ratio of task vector $\lambda _{i}\left( \boldsymbol{
\theta }_{I^{\ast }(w_{i})t}\right) $, 
\begin{equation*}
\lambda _{i}\left( \boldsymbol{\theta }_{I^{\ast }(w_{i})t}\right)
\triangleq \log p_{t}\left( \boldsymbol{\theta }_{I^{\ast }(w_{i})t}|w_{i},
\mathcal{F}_{t-}\right) -\log p_{t}\left( \boldsymbol{\theta }_{I^{\ast
}(w_{i})t}|w_{0},\mathcal{F}_{t-}\right) 
\end{equation*}
then a little rearrangement gives us an adaptation of the usual Bayesian
linear updating equation linking posterior and prior odds viz:
\begin{equation}
\rho _{it}^{\ast }=\rho _{it}+\lambda _{i}\left( \boldsymbol{\theta }
_{I^{\ast }(w_{i})t}\right) +\widehat{\lambda }_{i}\left( \boldsymbol{\theta 
}_{I^{\ast }(w_{i})t}\right)   \label{logoddsgeneral}
\end{equation}
Note that equation (\ref{task set integrity}) holds in particular whenever it is a 
\emph{simple task vector}; i.e. has the property that for any time $t>0,$ $
\omega \in \Omega $ and $i=1,2,\ldots ,m$
\begin{equation}
p_{t}\left( \boldsymbol{\theta }_{\widehat{I}^{\ast }(w_{i})t}|\boldsymbol{
\theta }_{I^{\ast }(w_{i})t},w_{i},\mathcal{F}_{t-}\right) =p_{tj}(
\boldsymbol{\theta }_{\widehat{I}^{\ast }(w_{i})t}|\boldsymbol{\theta }
_{I^{\ast }(w_{i})t},w_{0},\mathcal{F}_{t-})  \label{simple task defn}
\end{equation}
Property \ref{simple task defn} holds whenever the other tasks are useless for discriminating
any threat position $w_{i}$ from $w_{0}$: the probability $\omega $ engages
in these tasks does not depend on these other tasks. In this case the term $
\widehat{\lambda }_{i}\left( \boldsymbol{\theta }_{I^{\ast }(w_{i})t}\right) 
$ vanishes and 
\begin{equation}
\rho _{it}^{\ast }=\rho _{it}+\lambda _{i}\left( \boldsymbol{\theta }
_{I^{\ast }(w_{i})t}\right)  \label{logoddssimple}
\end{equation}
For practical reasons we have often found it convenient to decompose $\lambda
_{i}\left( \boldsymbol{\theta }_{I^{\ast }(w_{i}),t,j}\right)$
into functions of components in $I_{-}^{\ast }(w_{i})$ and $I_{+}^{\ast
}(w_{i})$ respectively. When these are disjoint and conditionally
independent given $w_{i}$ - as in practice we find is often a plausible
assumption then 
\begin{equation}
\lambda _{i}\left( \boldsymbol{\theta }_{I^{\ast }(w_{i})t}\right) =\lambda
_{-i}\left( \boldsymbol{\theta }_{I_{-}^{\ast }(w_{i})t}\right) +\lambda
_{+i}\left( \boldsymbol{\theta }_{I_{+}^{\ast }(w_{i})t}\right) 
\label{logoddscomplex}
\end{equation}
where 
\begin{eqnarray*}
\lambda _{-i}\left( \boldsymbol{\theta }_{I_{-}^{\ast }(w_{i})t}\right) 
&\triangleq &\log p_{t}\left( \boldsymbol{\theta }_{I_{-}^{\ast
}(w_{i})t}|w_{i},\mathcal{F}_{t-}\right) -\log p_{t}\left( \boldsymbol{\theta 
}_{I_{-}^{\ast }(w_{i})t}|w_{0},\mathcal{F}_{t-}\right) , \\
\lambda _{+i}\left( \boldsymbol{\theta }_{I_{+}^{\ast }(w_{i})t}\right) 
&\triangleq &\log p_{t}\left( \boldsymbol{\theta }_{I_{+}^{\ast
}(w_{i})t}|w_{i},\mathcal{F}_{t-}\right) -\log p_{t}\left( \boldsymbol{\theta 
}_{I_{+}^{\ast }(w_{i})t}|w_{0},\mathcal{F}_{t-}\right) 
\end{eqnarray*}
Note here that, by definition, $\lambda _{-i}\left( \boldsymbol{\theta }
_{I_{-}^{\ast }(w_{i})t}\right) $, takes its maximum value when $\boldsymbol{
\theta }_{I_{-}^{\ast }(w_{i})t}=\boldsymbol{0}$ and $\lambda _{+i}\left( 
\boldsymbol{\theta }_{I_{+}^{\ast }(w_{i})t}\right) $ takes its maximum
value when $\boldsymbol{\theta }_{I_{+}^{\ast }(w_{i})t}=\boldsymbol{1,}$ $
i=1,2,\ldots ,m.$
\\
\\
The equations (\ref{logoddssimple}) are now sufficient to calculate
the probability that $\omega $ is in each of the positions $
w_{1},w_{2},\ldots ,w_{m}$ given our evidence, using the familiar invertible
function from log odds to probability: see \cite{smithbook}.
\subsubsection{Model Assumptions concerning routine observations}
For our chosen filtered sequence $\boldsymbol{Z}_{t}\triangleq \left(
Z_{1t},Z_{2t},\ldots ,Z_{Rt}\right) $ designed to pick up the different
tasks associated with a criminal process let
\\
\\
\begin{equation*}
\boldsymbol{Z}_{\widehat{k}t}\triangleq \left( Z_{1t},Z_{2t},\ldots
,Z_{\left( k-1\right) t},Z_{\left( k+1\right) t},\ldots ,Z_{Rt}\right)
\end{equation*}
Then a simple but bold type of Naive Bayes assumption is to assume that 
filter $\boldsymbol{Z}_{t}$ is
\emph{pure}: i.e. that for any set $\boldsymbol{\theta }_{At}$ containing 
$\theta _{kt}$ as a component 
\begin{equation}
\boldsymbol{Z}_{kt}\Indep \boldsymbol{Z}_{\widehat{k}t}|\boldsymbol{\theta }
_{At},\mathcal{F}_{t-}  \label{Full obs cond ind}
\end{equation}
Then Bayes Rule and task set integrity implies that within tasks 
\begin{equation*}
p_{t}\left( \boldsymbol{Z}_{I^{\ast }(w_{i})t}|\boldsymbol{\theta}_{
_{I^{\ast }(w_{i})}t},\mathcal{F}_{t-}\right) =\prod\limits_{k\in I^{\ast
}(w_{i})}p_{t}\left( \boldsymbol{Z}_{kt}|\boldsymbol{\theta }_{kt},
\mathcal{F}_{t-}\right)
\end{equation*}
whilst across tasks
\begin{equation}
p_{t}\left( w_{it}|\boldsymbol{Z}_{t},\mathcal{F}_{t-}\right) =\sum_{\boldsymbol{\theta }_{I^{\ast }(w_{i})t}\in \Theta _{I^{\ast
}(w_{i})}}p_{t}\left( w_{it}|\boldsymbol{\theta}_{{I^{\ast
}(w_{i})}t},\mathcal{F}_{t-}\right) p_{t}\left( \boldsymbol{\theta}_{
_{I^{\ast }(w_{i})}t}|\boldsymbol{Z}_{I^{\ast }(w_{i})t},
\mathcal{F}_{t-}\right)  \label{uncertain update on w}
\end{equation}
where 
\begin{equation*}
p\left( \boldsymbol{\theta}_{{I^{\ast }(w_{i})}t}|\boldsymbol{Z}
_{I^{\ast }(w_{i})t},\mathcal{F}_{t-}\right) =\frac{p\left( \boldsymbol{Z}
_{I^{\ast }(w_{i})t}|\boldsymbol{\theta}_{{I^{\ast }(w_{i})}t},
\mathcal{F}_{t-}\right) p\left( \boldsymbol{\theta}_{{I^{\ast
}(w_{i})}t}|\mathcal{F}_{t-}\right) }{p\left( \boldsymbol{Z}_{I^{\ast
}(w_{i})t,j}|\mathcal{F}_{t-,j}\right) }
\end{equation*}
Let the prior and\ posterior odds be respectively denoted by
\begin{eqnarray*}
\begin{aligned}
\phi _{i^{\prime }t}^{iA} & \triangleq \log p_{t}\left( \boldsymbol{\theta}_{ _{I^{\ast }(w_{i})}t}=\chi _{A}|\mathcal{F}_{t-}\right) -\log
p_{t}\left( \boldsymbol{\theta}_{{I^{\ast }(w_{i})}t}=\chi
_{\emptyset }|\mathcal{F}_{t-}\right) \\
\phi _{i^{\prime }t}^{\ast iA} & \triangleq \log p_{t}\left( \boldsymbol{\theta}_{ _{I^{\ast }(w_{i})}t}=\chi _{A}|\boldsymbol{Y}_{I^{\ast }(w_{i})t},
\mathcal{F}_{t-}\right) -\log p_{t}\left( \boldsymbol{\theta}_{
_{I^{\ast }(w_{i})}t}=\chi _{\emptyset }|\boldsymbol{Y}_{I^{\ast }(w_{i})t},
\mathcal{F}_{t-}\right) \\
\gamma _{kt} & \triangleq \log p_{t}\left( \boldsymbol{Z}_{kt}|\theta _{kt}=1,
\mathcal{F}_{t-}\right) -\log p_{t}\left( \boldsymbol{Z}_{kt}|\theta _{kt}=0,
\mathcal{F}_{t-}\right)
\end{aligned}
\end{eqnarray*}
and let 
$\gamma _{t}^{A}\triangleq \sum_{k\in A}\gamma _{kt}$. Then
\begin{equation*}
\log p_{t}\left( \boldsymbol{Z}_{I^{\ast }(w_{i})t}|\boldsymbol{\theta }
_{I^{\ast }(w_{i})t}=\chi _{A},\mathcal{F}_{t-}\right) -\log
p_{t}\left( \boldsymbol{Z}_{I^{\ast }(w_{i})t}|\boldsymbol{\theta}_{
_{I^{\ast }(w_{i})}t}=\chi _{\emptyset },\mathcal{F}_{t-}\right) =\gamma
_{t}^{A}
\end{equation*}
and from the above 
\begin{equation}
\phi _{^{\prime }t}^{\ast iA}=\phi _{t}^{iA}+\gamma _{i^{\prime }t}^{A}
\label{taskupdate}
\end{equation}
For any set $A$ we can therefore calculate 
\begin{equation}
p_{t}\left( \boldsymbol{\theta}_{{I^{\ast }(w_{i})}t}=\chi _{A}|
\boldsymbol{Z}_{I^{\ast }(w_{i})t},\mathcal{F}_{t-}\right) \text{ }
\label{prob thet given y}
\end{equation}
where the Law of Total Probability implies that for $i=0,1,2,\ldots ,m$
\begin{equation}
p_{t}(W_{t}=w_{i}|\boldsymbol{Z}_{t},\mathcal{F}_{t-})=\sum_{\boldsymbol{
\theta }\in \Theta _{I^{\ast }(w_{i})}}p_{t}\left( w_{i}|\boldsymbol{\theta }
,\mathcal{F}_{t-}\right) p_{t}\left( \boldsymbol{\theta }|\boldsymbol{Z}
_{I^{\ast }(w_{i})t},\mathcal{F}_{t-}\right) 
\label{Total prob for position given y}
\end{equation}
Here the position probabilities over tasks calculated from equation (\ref{taskupdate})
are averaged over the different tasks possibly
explaining the data, weighting using the posterior
probabilities given in equation (\ref{prob thet given y}). Note that it is easy to
check that these indirect observations provide less discriminatory power
than when tasks are observed directly. The assumptions above therefore provide us with a formally justifiable
propagation algorithm for updating the probabilities of a suspect's likely
criminal status. We next turn to how we might calibrate the model to the
expert judgements we might elicit from criminologists, police and technicians
about the probable relationships between criminal status, what they might
try to accomplish and how this endeavour might be reflected through how they
communicate. 
\section{The elicitation process} \label{elicitation}
\subsection{Introduction}
Copying the standard protocols for the elicitation of a Bayesian Network:
see e.g. \cite{Korb}, as in 
e.g. \cite{WilkersonSmith} our process begins with the elicitation of
structure. We perform a sequence of three structural elicitations for each of the three
levels. These can proceed almost
entirely using natural language descriptions of the process. Because the
representation of the structure of each of the levels is formal and
compatible with a probability model the structural elicitation can take
place before the model is quantified. This is extremely helpful because
structural information is typically much easier to elicit faithfully than
quantitative judgments. The RDCEG defining this structure, the list of
tasks and how these might interrelate and then the choice of filter of the
routine observations follows:
\begin{enumerate}
\item First the decision analyst elicits the positions of the process via the
careful conversion of natural language expressions within the domain
experts' description of the process into the topology of a RDCEG\ - often
somewhat more nuanced than the ones we discussed above and in the example
below.
\item Second the positions and edges of this RDCEG are then associated to elicited
portfolios of tasks.
\item Finally each task is associated with the way domain experts and police
believe $\omega $ might behave in order to carry out these tasks, including
how they might choose to disguise these actions and so what signals might be
visible when $\omega $ enacts a task.
\end{enumerate}
We now briefly outline each of these steps in turn in a little more detail.
\subsection{Choosing an appropriate RDCEG}
Firstly, when eliciting a RDCEG we aim to keep the number of positions 
as small as possible within the constraint that they are sufficient to
distinguish relevant states. The choice of topology should reflect what is known
about the development of the modelled criminal behaviour.  
Positions may depend on the history, environmental and personality
profile covariates exhibited by a suspect. Relevant population studies of
criminal behaviours are often helpful here. Based on historical cases, 
criminologists' analyses \cite{gill} and discussions with practitioners, 
we have found that the coarsest
type of model - an illustration of such given in the next section - of
different types of attack and concerning different people - are often
generic.
\\
\\
Secondly positions need to be well defined enough to pass the Clarity Test 
\cite{Howard,smithbook}. This is achieved by demanding that the
suspect could, if they were so minded, place themselves in a particular
position. Such categories are often syntheses of standard scales used by
social workers and probation workers across the world. Many examples of
these types of categorisations, based on fusing various publicly available
categorisations - for example those found in the training manuals of social
workers in detecting people threatening to eventually perpetrate acts of
severe violence - are given in \cite{smithassault2018}.
\\
\\
Thirdly positions must be defined such that for each position there is a collection of tasks
associated with it that jointly informs whether the suspect is in that position or transitioning from that position 
to another position. A stylised example of
this association was given earlier in this paper and we will illustrate the
process in more detail in the next section with a deeper
illustration.
\\
\\
Once the RDCEG has been drawn its embedded assumptions can be queried by
automatically generating logical deductions concerning the implications of
the model. If such deductions appear implausible to relevant domain experts
then positions need to be redefined and graphs redrawn until they are. The
ways of iterating until a model is requisite and the nature of these
deductions is beyond the scope of this paper but are discussed in \cite{collazo2018chain}.
\\
\\
The final step in the elicitation of the RDCEG will be the prior conditional
probabilities associated with the positions and the hyperparameters of the
holding times. Suitable generic methods for this
elicitation are now very well established: see e.g. \cite{Ohagan,smithbook} and these need little adaptation to be
applied. Note, in particular, that the methods described for
the elicitation of the position probabilities in the RDCEG are essentially
identical to those for the CEG as for example discussed in a chapter of \cite{collazo2018chain}.
\subsection{Elicitation of portfolios of tasks}
\subsubsection{Clustering the tasks}
The next elicitation process is to take each position 
in turn and a list of associated tasks conditional on $\omega $ 
being known to lie in that position. The questions we might ask
would be something like \say{Now suppose that you happen to learn that $\omega $
lies in position $w_{i}$. What behaviours/tasks would you expect them to
perform that would be different from what they would typically do were they
neutral?} We try to ensure that either $\omega $'s engagement in such tasks
could be learned through intelligence or alternatively be indicated through
certain filters. 
\\
\\
Typically in well designed models we specify tasks so that they are as specific to only a small
proportion of $\omega $'s active states. This makes them as discriminatory as
possible. Note that each component of $\boldsymbol{\theta }_{t}$ must be
defined sufficiently precisely - i.e. pass the clarity test for $\omega$ 
to be able to divulge its value if so
inclined; see \cite{smithassault2018}.
\\
\\
We have found it useful to toggle between specifying positions and
specifying tasks: sometimes aggregating positions if they appear associated
with the same sets of tasks or splitting a position into a set of new ones
if a finer definition can discriminate between one position and another. It
is also sometimes helpful to readjust the definition of tasks once we have
elicited possible signals.
\\
\\
Once the task sets are requisite \cite{Phillips,smithbook} we need to
specify the various odds ratios against the neutral state. We illustrate this in the next section.
\subsubsection{Simplifying assumptions that can ease task probability
elicitation}
Although the log score updating formulae (equations \ref{logoddsgeneral}, \ref{logoddssimple})
are simple ones, to evaluate the log odd scores above can
demand a great many probabilities, both of each task given its associated
position and of seeing that task performed if $\omega $ were neutral, to be
elicited or estimated. This can destabilise the system unless various simplifying
assumptions are made.
\\
\\
Conditional on an active position $w_i$ we recommend that the first
probability to be elicited is when $\omega $ is engaging in \emph{all} the
tasks in the portfolio of tasks associated with $w$. We then use this elicitation to
benchmark the probability that $\omega $ is engaging in a subset of these
tasks. To calculate the odds of the portfolio against the neutral suspect,
a default assumption that is sometimes appropriate is simply to assume that
people will engage in these probabilities independently- \emph{a naive Bayes
Assumption} \cite{smithbook} for the neutral suspect. In this case for any time 
$t>0,$ $\omega \in \Omega $ and $i=1,2,\ldots ,m$ 
\begin{equation*}
p_{t}\left( \boldsymbol{\theta }_{t}|w_{0},\mathcal{F}_{t}\right)
=\prod\limits_{k=1}^{R}p_{t}\left( \theta _{tk}|w_{0},\mathcal{F}%
_{t}\right) 
\end{equation*}%
It is easy to check that when a portfolio contains more than one task, and when such
an assumption is valid, it can provide the basis of a very powerful
discriminatory tool. This is because the divisor in the relevant odds
reduces exponentially with the number of tasks whilst in the denominator
does not: see e.g. \cite{smithassault2018} for an example of this.
\\
\\
Of course in some instances this naive Bayes assumption may not be\
appropriate. It will then need to be substituted. When population statistics
associated with public engagement in different task activities are available
these can be used to verify this assumption or form the basis of
constructively replacing it. 
\subsection{Choosing an appropriate filter of routine data streams}
Typically we would like the components of $\left\{ \boldsymbol{Z}%
_{t}\right\} _{t\geq 0}$ to measure an intensity of activity related to a
position or edge task. In this sense we would therefore like to factor out
all signals that might be considered typical of $\omega $'s innocent
activities so that we can focus on the incriminating signals. The full data
stream $\left\{ \boldsymbol{Y}_{t}\right\} _{t\geq 0}$ collected on $\omega $
tends to be a highly non-stationary multivariate time series. However we
strive to construct the filter $\left\{ \boldsymbol{Z}_{t}\right\} _{t\geq 0}
$ so that the stochastic dependence it exhibits is explained solely by $%
\omega $'s engagement in certain tasks (see equation (\ref{Markov task filter})). This
filter is clearly dependent both on population level signals and what we
know about $\omega $'s personality. We therefore  usually need expert
judgments to choose $\left\{ \boldsymbol{Z}_{t}\right\} _{t\geq 0}$ so that
it is fit for purpose.
\\
\\
There are some generic features that are worth introducing at this stage.
First in the case of edge tasks we typically observe something different
than before as $\omega $ begins to enact a new task in order to make a
transition. So some components of $\boldsymbol{Z}_{t}$ will be defined as
first differences of derived series. Secondly indicative observations may
also need to be smoothed from the past - either because what we see may
forewarn a task is about to be enacted, or simply because short term
averages - for example any measure of intensity of communication - will
often be better represented by an average over the recent past rather than
by an instantaneous measure.
\\
\\
To construct our hierarchical model we typically loop around the bullets
below:
\begin{enumerate}
\item Reflect on what functions of the vector of observable data
available to the police might help indicate that a suspect
really does lie in a particular task rather than other related innocent
activities, $k=1,2,\ldots ,R$. This choice should be informed by the ease at
which such signals can be filtered but also how easy it might be for a
criminal to disguise that signal were they to learn that the chosen
candidate filter was being used.
\item Using expert judgments and any survey data available, reflect on what
the distribution of $\boldsymbol{Z}_{kt}$ might be were the suspect actually
engaged in the particular task and if they were not. Thus specify 
\begin{equation*}
p(\boldsymbol{Z}_{kt}|\theta _{kt}=1)\text{ and }p(\boldsymbol{Z}%
_{kt}|\theta _{kt}=0)
\end{equation*}
\item Check that these two distributions are not close to one another. If not return
to the first step.
\end{enumerate}
Finally in many instances of such police work routine measurements
concerning suspects are typically recorded and reported over fixed periods
of time. This means that the filtered observation sequence is a discrete
time filter. For modelling purposes it has been necessary to define the deep
stochastic process as semi-Markov. However the semi-Markov structure with
holding times and transition probabilities specified will retain the Markov
structure over the fixed time points. This means once appropriate
transformations are applied standard updating rules associated with Markov
switching models are then valid \cite{Frutthwirth-Schnatter}: see 
Appendix \ref{AppendixSemiMarkov}.
\newpage
\section{A vehicle attacker example} \label{vehicle}
\subsection{States}
\begin{figure}[H]
    \centering
    \includegraphics[width=2.5in]{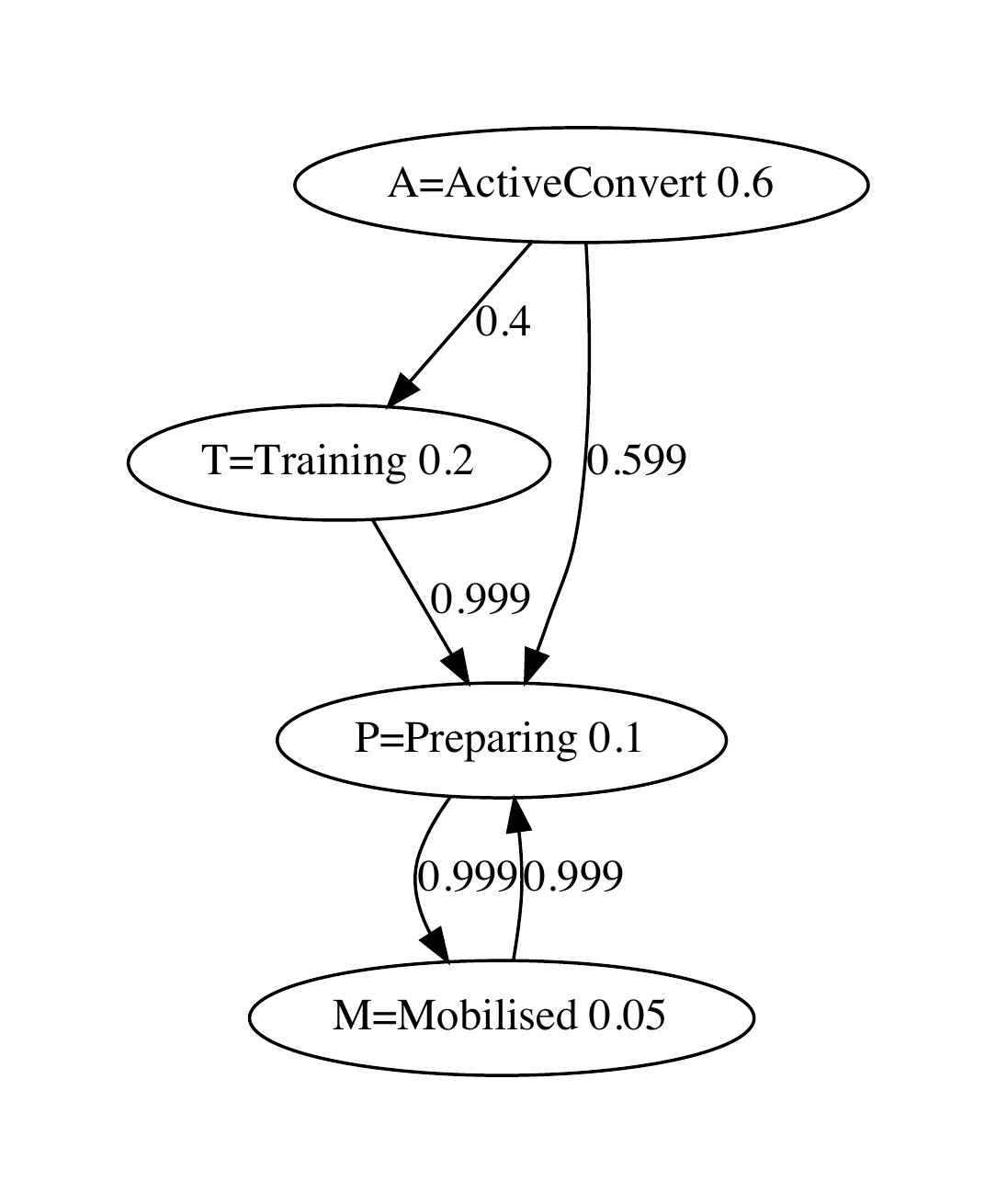}
    \caption{RDCEG for Vehicle Attacker}
    \label{rdcegprior}
\end{figure}

We now give a more detailed example of a vehicle
attacker that illustrates the three level hierarchical model of latent states, tasks, and routinely observable data. 
We specify the states of the RDCEG to be:
\begin{equation}
W = \{N,A,T,P,M\}
\end{equation}
where N is \say{Neutral}, A is \say{ActiveConvert}, T is \say{Training}, P is \say{Preparing}, and M 
is \say{Mobilised}. Based on existing information about 
the suspect we assign prior probabilities to each state as shown in Figure \ref{rdcegprior}; implicitly the prior
probability for the elided \say{Neutral} state is $0.05$.
Based on knowledge about such attacks we hypothesize that the 
suspect may transition from A to T or P, from T to P, from P to M, and from M back to P.
These transitions are indicated by the directed edges between the vertices on said figure. The weights labelling
the transitions are the probabilities of transitions from the source vertex to the destination vertex
 conditional on a transition having occurred (i.e. the entries labelled $m_{w_i,w_j}$ in Table \ref{transitionMatrix} in 
Appendix \ref{AppendixSemiMarkov}).
The probability of transition into the \say{Neutral} state from any represented state
is implied by the sum of all the emanating edges' probabilities summing to one. This is in
contrast to \cite{Shenvi} where the transition probabilities are conditional on not moving
to the absorbing state. This prior RDCEG is used in all the examples 
in this section.
\subsection{Tasks}
We hypothesize the tasks relevant for these positions, i.e. the tasks that are related to
which position the suspect is in, are:
\begin{equation*}
\boldsymbol{\theta} = \{\theta_j\}_{j=1}^{10}
\end{equation*}
where: \\ 
\begin{footnotesize}
\tab $\theta_1$ is Engaging with Radicals 
\tab $\theta_2$ is Engaging in Public Threats \\
\tab $\theta_3$ is Making Personal Threats 
\tab $\theta_4$ is Fewer Public Engagements in Radicalisation\\
\tab $\theta_5$ is Fewer Contacts with Family and Friends
\tab $\theta_6$ is Securing Monetary Resources\\
\tab $\theta_7$ is Learning to Drive Large Vehicle
\tab $\theta_8$ is Obtaining Vehicle\\
\tab $\theta_9$ is Reconnaissance of Target Locations
\tab $\theta_{10}$ is Moving to Target Location\\
\end{footnotesize}
\\
For each position $w_i$, a particular subset of the above tasks are taken to be indicators that the suspect
is there. $I^{\ast}(w_i)$ is the index set for this subset and we need to
specify the distribution $p(\theta_{I^{\ast}(w_i)}|w_i)$ for each position. 
Appendix \ref{AppendixProbs} details a methodology for this
specification that makes the model discriminatory and Table \ref{probconstruct} shows the resulting probabilities used.
\subsection{Routinely observable data}
The data used to estimate the probabilities that the suspect is engaging in any particular
task or tasks are varied and various and may change as technologies and data gathering methods change.
In addition as new evidence is gained and the threat level of a suspect increases the authorities
may decide to increase monitoring and hence gain more and new types of data. Therefore having
the tasks $\boldsymbol{\theta}$ intermediate the positions $W$ from the observed data $\boldsymbol{Y}$ is desirable 
for both model structural reasons and practical data abstraction purposes. We denote
the observable data as a $d$-dimensional vector process in discrete time $\boldsymbol{Y_t} = (Y_{i,t})_{i=1}^{d}$.
In general we assume $\boldsymbol{Y_t} \in \mathbb R^d$. In this example, however,
several of the components are count data, such as the number of
times such events are observed in a given period, so that: 
$Y_{i,t} \in \mathbb Z^+$.
We set here:\\ \\
\begin{footnotesize}
\tab $Y_1$: radical website visits 
\tab $Y_2$: physical meetings with known radicals \\
\tab $Y_3$: electronic meetings with known radicals 
\tab $Y_4$: meetings with trained radicals \\
\tab $Y_5$: meetings with known cell members
\tab $Y_6$: seen at radical demonstrations\\
\tab $Y_7$: contacts with non-radicals
\tab $Y_8$: public threats made\\
\tab $Y_9$: personal threats made
\tab $Y_{10}$: increase in known financial resources\\
\tab $Y_{11}$: decrease in known financial resources
\tab $Y_{12}$: obtaining large vehicle driving licence\\
\tab $Y_{13}$: vehicle dealer or rental website visits
\tab $Y_{14}$: vehicle dealer or rental physical visits\\
\tab $Y_{15}$: E-visits to target locations
\tab $Y_{16}$: physical visits to target locations\\
\tab $Y_{17}$: statements of intent
\tab $Y_{18}$: legacy statements\\
\end{footnotesize}
\\
As described in Section \ref{structure} we assume that for each task $\theta_j$ we can construct a filter $Z_j$
of the relevant data: Here we define the function $\boldsymbol{\tau}$: 
\begin{eqnarray*}
\boldsymbol{\tau}: \mathbb R^d \mapsto \mathbb R^R,\; Z_{j,t}&=&\tau_j(\boldsymbol{Y_t}) \\
\tau_j(\boldsymbol{Y_t}) &=& \frac{1}{|I_{\theta_j}|} \: \sum_{i\in I_{\theta_j}} \tilde{y}_{i,t}
\end{eqnarray*}
where $\tilde{y}_i$ is a normalisation\footnote{we subtract a pre-defined mean and divide by a pre-defined 
standard deviation estimated by investigators' judgement, experience and historical data}
o $Y_i$ and $|I_{\theta_j}|$ is the cardinality of the index set of components
of $Y$ dependent on the $j^{th}$ task.
We could also set additional components of $\boldsymbol{Y}$ to be changes over time of other components of $\boldsymbol{Y}$ and thus
monitor drops or spikes in, for example, communication levels with known radicalisers, or with family and 
friends. We specify the relationship between the observable data and the tasks in Table~\ref{TaskObservableDep} where the 
$(Y_i,\theta_j)$ entry
indicates whether the $i$th variable is relevant data for the $j$th task.
\begin{table}
\small\addtolength{\tabcolsep}{-3pt}
\begin{footnotesize}
\begin{tabular}{llrrrrrrrrrrr}
\toprule
                            Observable & $\theta_1$ &  $\theta_2$ &  $\theta_3$ &  $\theta_4$ &  $\theta_5$ &  $\theta_6$ &  $\theta_7$ &  $\theta_8$ &  $\theta_9$ &  $\theta_{10}$  \\
\midrule
                  RadWebVisits &    1 &  0 &  0 &  1 &  1 &  0 &  0 &  0 &  0 &  0 \\
     PhysicalMeetsWithRadicals &    1 &  0 &  0 &  0 &  1 &  0 &  0 &  0 &  0 &  0 \\
           E-MeetsWithradicals &    1 &  0 &  0 &  1 &  1 &  0 &  0 &  0 &  0 &  0 \\
           MeetTrainedRadicals &    1 &  0 &  0 &  0 &  1 &  0 &  0 &  0 &  0 &  0 \\
              MeetCellMembers &    1 &  0 &  0 &  0 &  1 &  0 &  0 &  0 &  0 &  0 \\
   SeenAtRadicalDemonstrations &    1 &  0 &  0 &  0 &  1 &  0 &  0 &  0 &  0 &  0 \\
       ContactsWithNonRadicals &    0 &  0 &  0 &  1 &  0 &  0 &  0 &  0 &  0 &  0 \\
             PublicThreatsMade &    0 &  1 &  0 &  0 &  0 &  0 &  0 &  0 &  0 &  0 \\
            PersonalThreatMade &    0 &  0 &  1 &  0 &  0 &  0 &  0 &  0 &  0 &  0 \\
            IncreaseInFinances &    0 &  0 &  0 &  0 &  0 &  1 &  1 &  1 &  0 &  0 \\
            DecreaseInFinances &    0 &  0 &  0 &  0 &  0 &  0 &  1 &  1 &  0 &  0 \\
              ObtainLGVLicence &    0 &  0 &  0 &  0 &  0 &  0 &  1 &  0 &  0 &  0 \\
              CarDealerWebHits &    0 &  0 &  0 &  0 &  0 &  0 &  0 &  1 &  0 &  0 \\
       CarDealerPhysicalVisits &    0 &  0 &  0 &  0 &  0 &  0 &  0 &  1 &  0 &  0 \\
     E-VisitsToTargetLocations &    0 &  0 &  0 &  0 &  0 &  0 &  0 &  0 &  1 &  0 \\
       VisitsToTargetLocations &    0 &  0 &  0 &  0 &  0 &  0 &  0 &  0 &  1 &  0 \\
              LegacyStatements &    0 &  0 &  0 &  0 &  0 &  0 &  0 &  0 &  0 &  1 \\
             StatementOfIntent &    0 &  1 &  1 &  0 &  0 &  0 &  0 &  0 &  0 &  1 \\
\bottomrule
\end{tabular}
\begin{tiny}
\caption{Routine Observation versus Task dependency structure}
\label{TaskObservableDep}
\end{tiny}
\end{footnotesize}
\end{table}
\subsection{Specifications of the distributions of the task set given position and task set likelihood}
\begin{figure}
    \centering
        \begin{subfigure}[b]{0.475\textwidth}
            \centering
            \includegraphics[width=\textwidth]{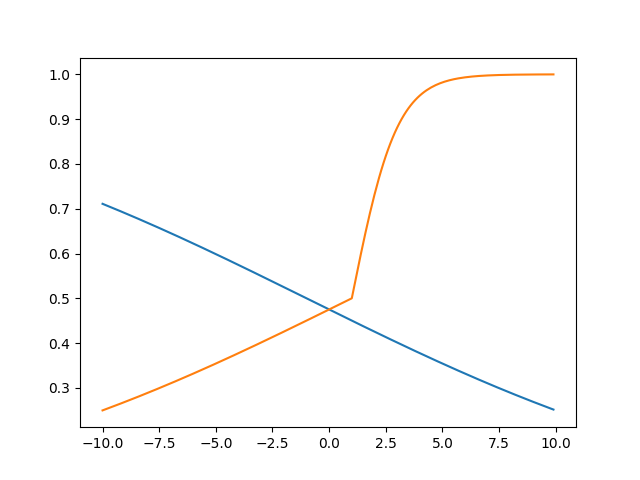}
            \caption{Orange line shows $p(Z_{I^{\ast}(w_i)}|\theta_{I^{\ast}(w_i)}=1)$;
                    Blue line shows $p(Z_{I^{\ast}(w_i)}|\theta_{I^{\ast}(w_i)}=0)$}
        \end{subfigure}
        \hfill
        \begin{subfigure}[b]{0.475\textwidth}
            \centering
            \includegraphics[width=\textwidth]{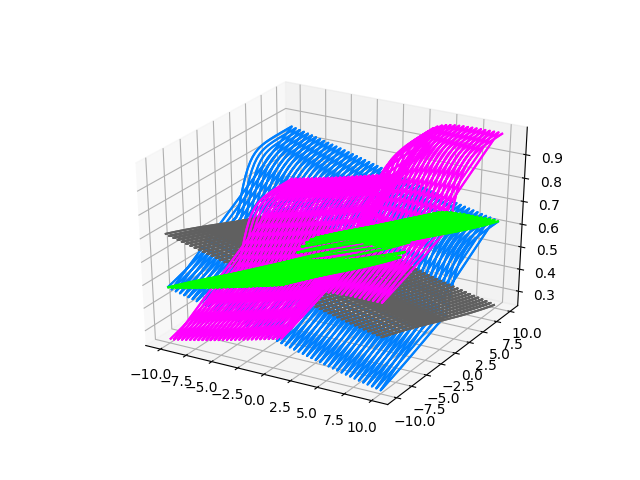}
            \caption{Two dimensional. Each coloured surface is for a different value of 
                        $\theta_{I^*_{w_i}} \in \{0,1\}^2$; $Z_{I^*_{w_i}}$}
        \end{subfigure}
        \vskip\baselineskip
        \begin{subfigure}[b]{0.475\textwidth}
            \centering
            \includegraphics[width=\textwidth]{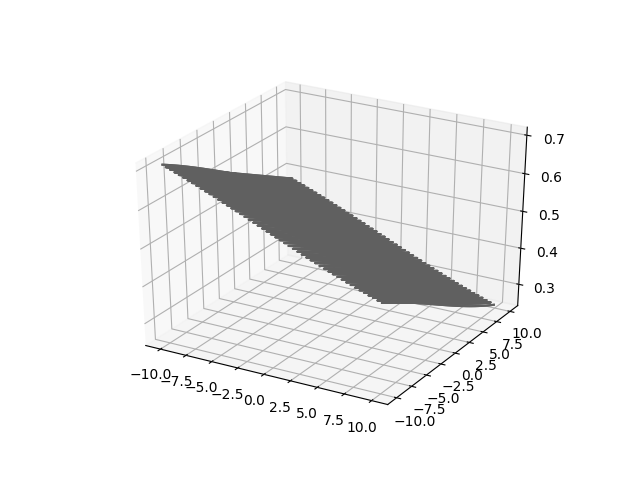}
            \caption{$\theta_{I^*_{w_i}} = (0,0)$}
        \end{subfigure}
        \hfill
        \begin{subfigure}[b]{0.475\textwidth}
            \centering
            \includegraphics[width=\textwidth]{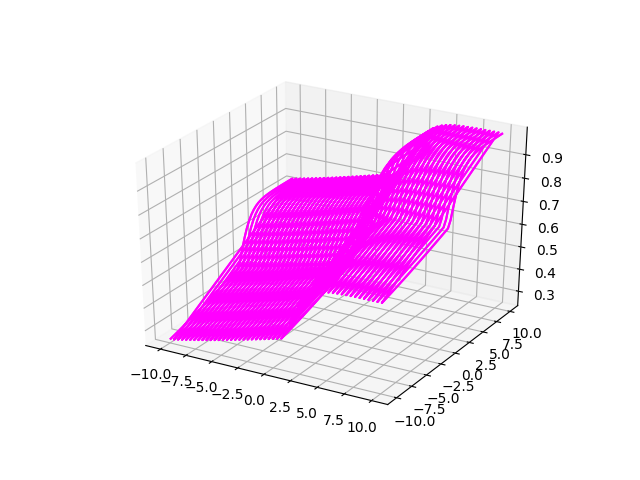}
            \caption{$\theta_{I^*_{w_i}} = (1,1)$}
        \end{subfigure}
    \caption{Illustrative task likelihood functional forms for one and two dimensional task sets}
    \label{pZtheta3D}
\end{figure}

For each position we set the probability that all the tasks in that position's task set are being done to $0.4$. The 
individual probabilities that each task is being done given the suspect is in the neutral state are specified in 
the column labelled \say{Neutral} in
Table \ref{TaskStateDep}.
The log odds interpolation methodology detailed in Appendix \ref{AppendixProbs} is then used to construct the 
probabilities that no tasks or
less than all the tasks were being done. For example, as shown in the Table \ref{TaskStateDep}, for the \say{Mobilised} 
position, the tasks \say{Engaging in public threats}, \say{Making personal threats}, 
\say{Reconnaissance of target locations}, and \say{Moving to target location} are relevant. Table 
\ref{probconstruct} has the resulting probabilities for each point of $\theta_{I^*_{w_i}} \in \{0,1\}^{|\theta_{I^*_{w_i}}|}$.
\begin{table}
\small\addtolength{\tabcolsep}{-3pt}
\begin{footnotesize}
\begin{tabular}{lrrrrr}
\toprule
{} &  ActiveConvert &  Training &  Preparing &  Mobilised &  Neutral \\
State\_Task\_Index\_Sets                   &                &           &            &            &          \\
\midrule
EngageWithRadicalisers                  &          1 &     0 &      0 &      0 &    0.020 \\
EngageInPublicThreats                   &          0 &     0 &      1 &      1 &    0.001 \\
MakePersonalThreats                     &          0 &     0 &      1 &      1 &    0.001 \\
RedPubEngInRad &          1 &     1 &      0 &      0 &    0.600 \\
RedCntctWthFmlyFrnds          &          1 &     0 &      0 &      0 &    0.300 \\
ObtainResources                         &          1 &     1 &      0 &      0 &    0.300 \\
LearnToDrive                            &          0 &     1 &      0 &      0 &    0.300 \\
ObtainVehicle                           &          0 &     1 &      1 &      0 &    0.200 \\
ReconnoitreTargets                      &          0 &     0 &      1 &      1 &    0.100 \\
MoveToTarget                            &          0 &     0 &      0 &      1 &    0.200 \\
Cardinality                             &          4 &     4 &      4 &      4 &       \\
p+                                      &          0.400 &     0.400 &      0.400 &      0.400 &       \\
p0                                      &          0.001 &     0.011 &      0.00000002 &    0.00000002   &       \\
$\xi$                                      &          1.051 &     2.076 &      0.332 &      0.331 &       \\
\bottomrule
\end{tabular}
\caption{Task/position dependencies; probability of task given Neutral state; p+ is probability of all tasks being done,
p0 that of none being done; $\xi_i$ is solved for to make the probabilities sum to one for each
position}
\label{TaskStateDep}
\end{footnotesize}
\end{table}
\begin{table}
\small\addtolength{\tabcolsep}{-3pt}
\begin{footnotesize}
\begin{tabular}{lrrrr}
\toprule
{} &  ActiveConvert &  Training &  Preparing &  Mobilised \\
State\_Task\_Index\_Sets &                &           &            &            \\
\midrule
np\_0                  &        0.00108 &   0.01080 &    0.00000002 &    0.00000002 \\
np\_0001               &        0.00475 &   0.01341 &    0.00112 &    0.00112 \\
np\_0010               &        0.00475 &   0.01341 &    0.00112 &    0.00112 \\
np\_0011               &        0.02319 &   0.02742 &    0.01860 &    0.01860 \\
np\_0100               &        0.00475 &   0.01341 &    0.00112 &    0.00112 \\
np\_0101               &        0.02319 &   0.02742 &    0.01860 &    0.01860 \\
np\_0110               &        0.02319 &   0.02742 &    0.01860 &    0.01860 \\
np\_0111               &        0.11019 &   0.09277 &    0.12098 &    0.12098 \\
np\_1000               &        0.00475 &   0.01341 &    0.00112 &    0.00112 \\
np\_1001               &        0.02319 &   0.02742 &    0.01860 &    0.01860 \\
np\_1010               &        0.02319 &   0.02742 &    0.01860 &    0.01860 \\
np\_1011               &        0.11019 &   0.09277 &    0.12098 &    0.12098 \\
np\_1100               &        0.02319 &   0.02742 &    0.01860 &    0.01860 \\
np\_1101               &        0.11019 &   0.09277 &    0.12098 &    0.12098 \\
np\_1110               &        0.11019 &   0.09277 &    0.12098 &    0.12098 \\
np\_1111               &        0.40000 &   0.40000 &    0.40000 &    0.40000 \\
ntp                   &        1.00000 &   1.00000 &    1.00000 &    1.00000 \\
\bottomrule
\end{tabular}
\begin{tiny}
\caption{Probabilities of task sets given each position using method in Appendix \ref{AppendixProbs}}
\label{probconstruct}
\end{tiny}
\end{footnotesize}
\end{table}
For simplicity we used shifted asymmetric logistic functions to construct the task set likelihood functions
$p(Z_{I^{\ast}(w_i)}|\theta_{I^{\ast}(w_i)})$: see equations (\ref{tasklike}).
The shift parameters
$x_{0,j}$ and the growth rate parameters $k_{0,j}$, $k_{1,j}$ were used to construct functions that were relatively
unresponsive when $Z_j < x_{0,j}$ but sharply responsive when $Z_j \geq x_{0,j}$. An illustration of the form of these 
functions is provided in Figure \ref{pZtheta3D} for the one and two-dimensional cases i.e. when there are one or two 
tasks in the task set $I^{\ast}(w_i)$; this is purely for ease of plotting: as shown in Table \ref{TaskStateDep} 
four dimensional task sets were used in the example scenarios.
\begin{eqnarray} \label{tasklike}
\begin{aligned}
p(Z_{I^{\ast}(w_i)}|\theta_{I^{\ast}(w_i)})\; &=& |\theta_{I^{\ast}(w_i)}|^{-1} \! \sum_{j \in I^{\ast}(w_i)} 
g(Z_j|\theta_j,x_{0,j},k_{0,j},k_{1,j}) \\
g(x|\theta,x_{0},k_{0},k_1)\; &=& \frac{1}{1 + \exp ( -k_{0}(x-x_{0}))}\;\; \boldsymbol{\chi}_{\{x < x_{0}\}}\; + \\
                     & & \frac{1}{1 + \exp ( -k_1(x-x_{0}))}\;\; \boldsymbol{\chi}_{\{x \geq x_{0}\}}
\end{aligned}
\end{eqnarray}
\subsection{Scenarios}
\begin{figure}[H]
    \centering
        \begin{subfigure}[b]{0.475\textwidth}
            \centering
            \includegraphics[width=\textwidth]{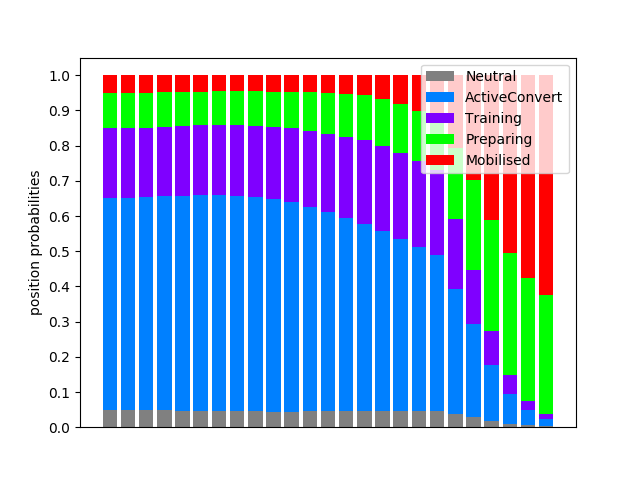}
            \caption{Scenario \ref{scen3}: a suspect pursuing activities consistent with preparing for an attack}
            \label{pp3}
        \end{subfigure}
        \hfill
        \begin{subfigure}[b]{0.475\textwidth}
            \centering
            \includegraphics[width=\textwidth]{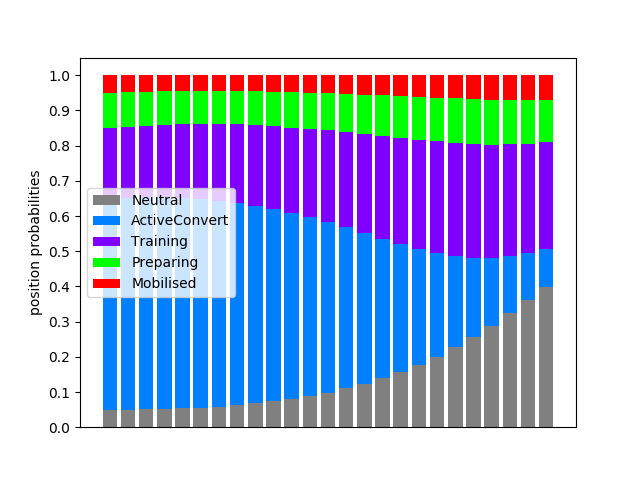}
            \caption{Scenario \ref{scen4}: a suspect reducing activities consistent with no evidence of threat}
            \label{pp4}
        \end{subfigure}
    \caption{Posterior state probabilities over time under scenario \ref{scen3} and \ref{scen4}}
\end{figure}

\begin{figure}
        \begin{subfigure}[b]{0.475\textwidth}
            \centering
            \includegraphics[width=\textwidth]{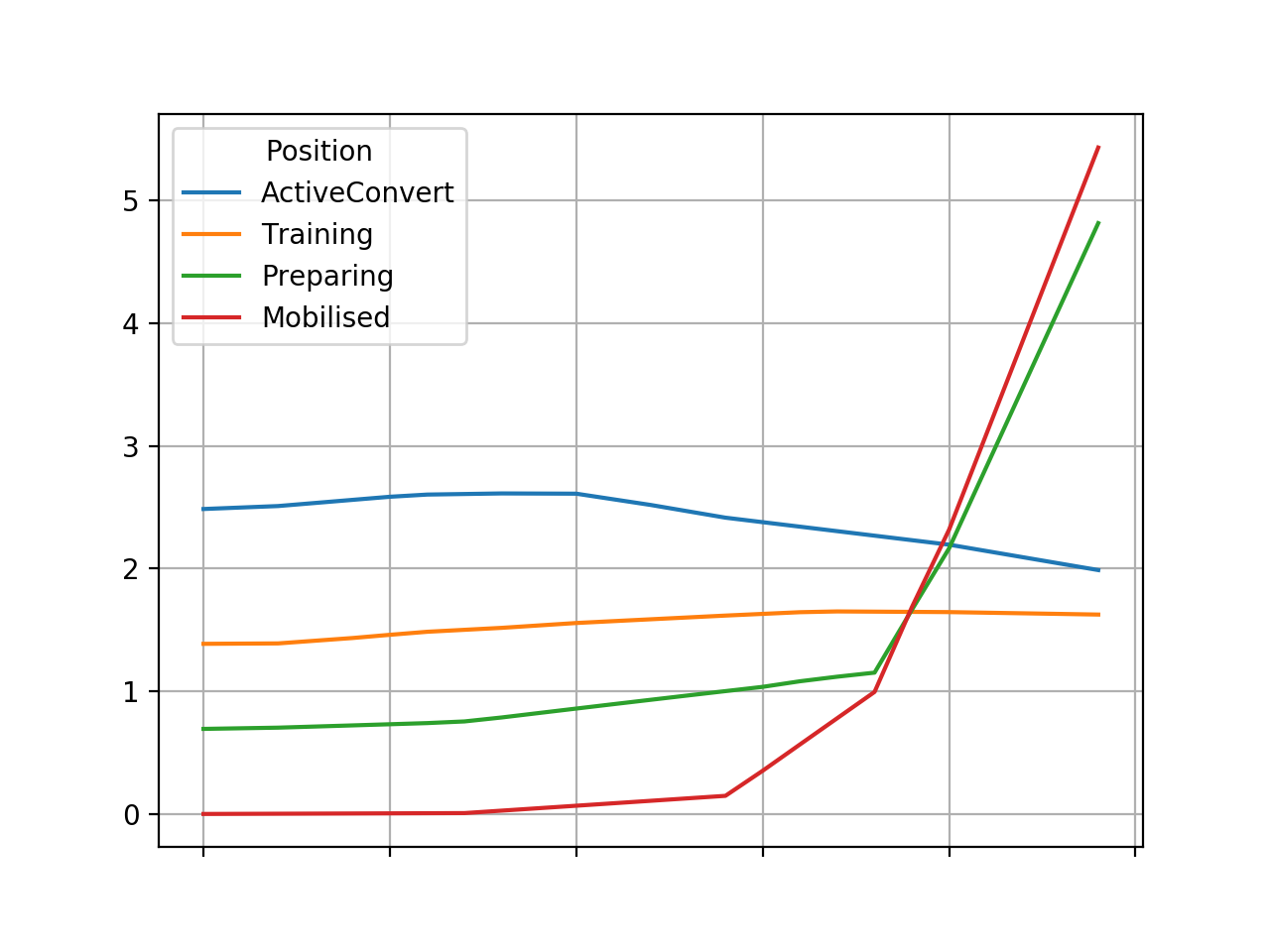}
            \caption{Position Score over time based on Scenario \ref{scen3}}
            \label{score3}
        \end{subfigure}
        \hfill
        \begin{subfigure}[b]{0.475\textwidth}
            \centering
            \includegraphics[width=\textwidth]{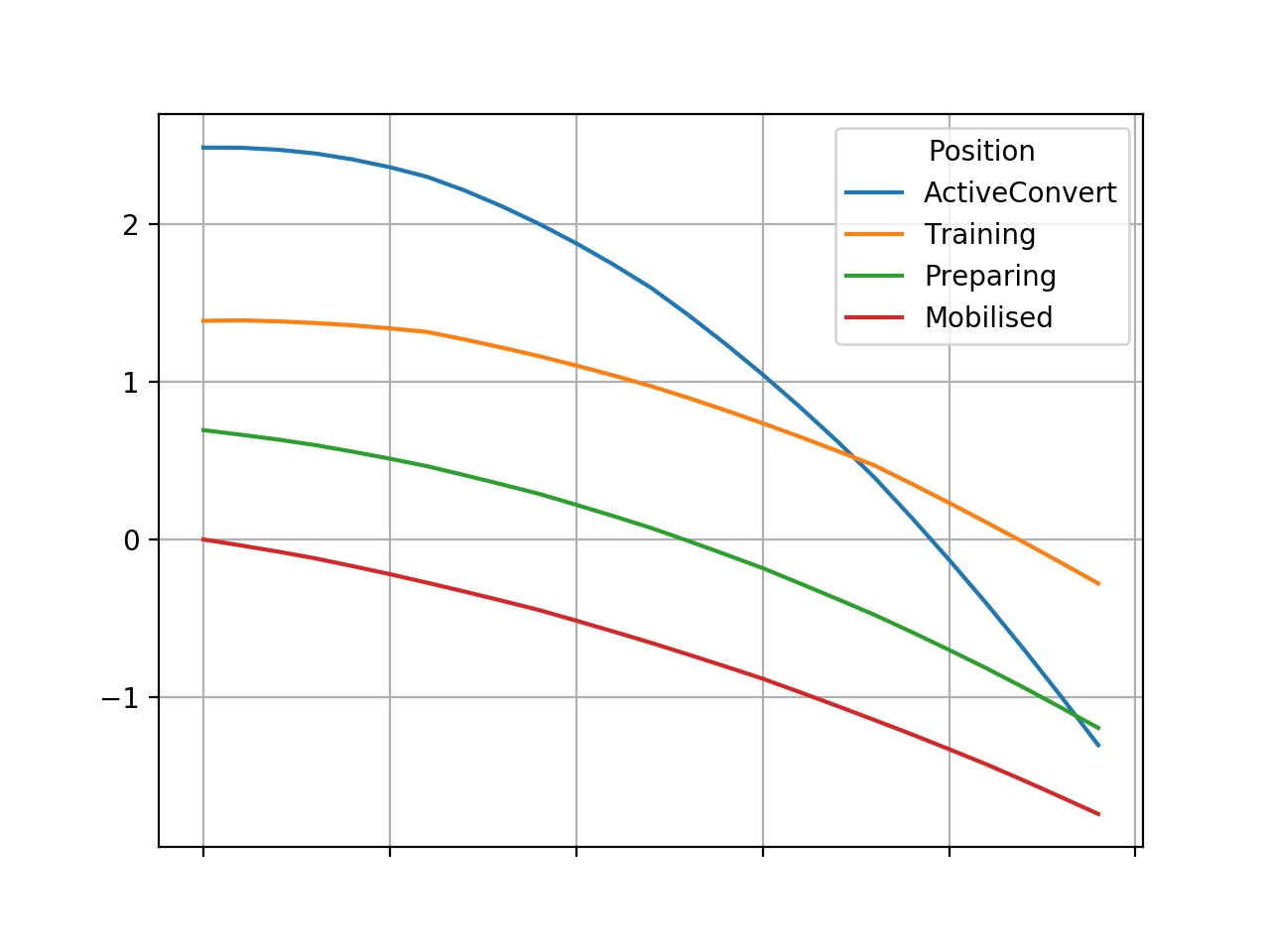}
            \caption{Position Score over time based on Scenario \ref{scen4}}
            \label{score4}
        \end{subfigure}
        \begin{subfigure}[b]{0.4\textwidth}
            \centering
            \includegraphics[width=\textwidth]{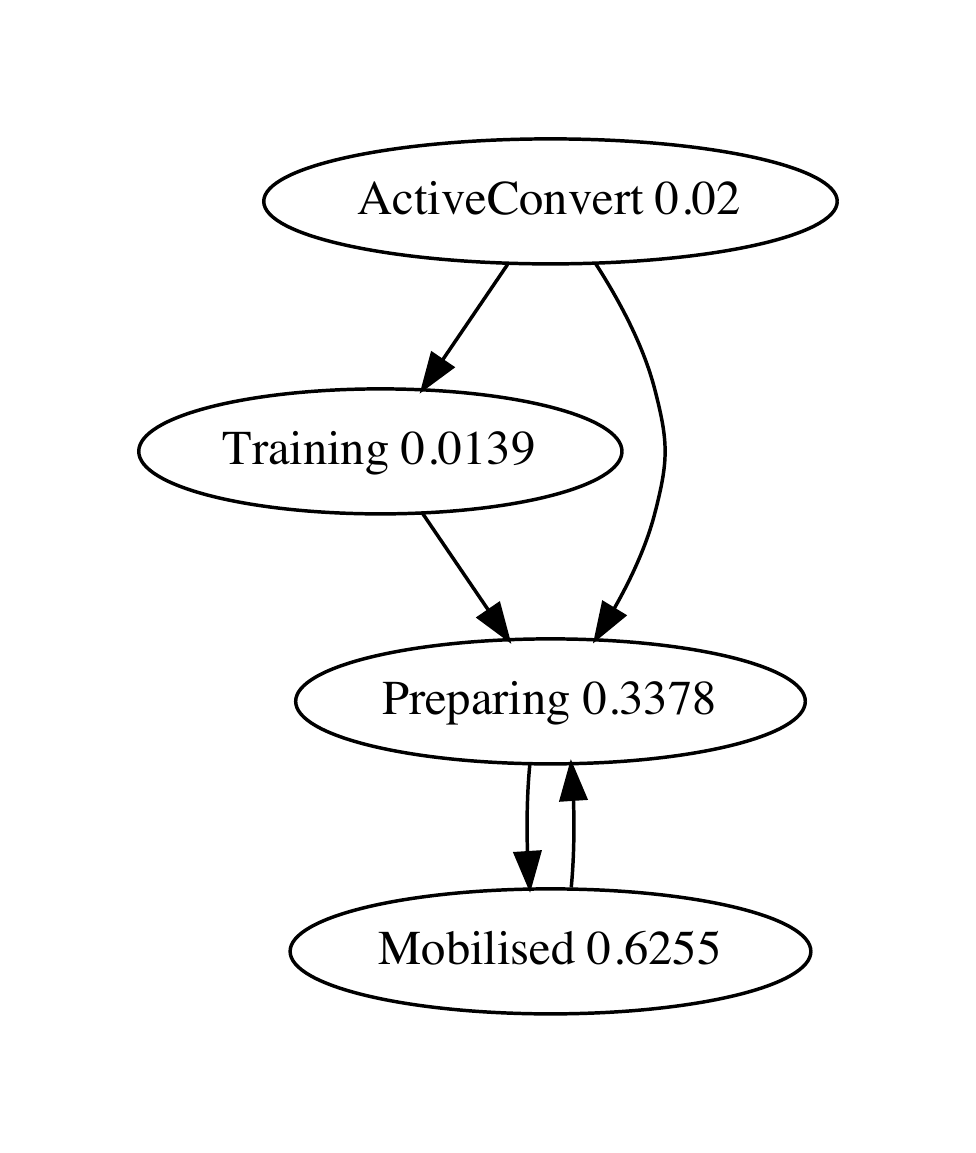}
            \caption{RDCEG with final posterior probabilities based on Scenario \ref{scen3}}
            \label{rd3}
        \end{subfigure}
        \hfill
        \begin{subfigure}[b]{0.4\textwidth}
            \centering
            \includegraphics[width=\textwidth]{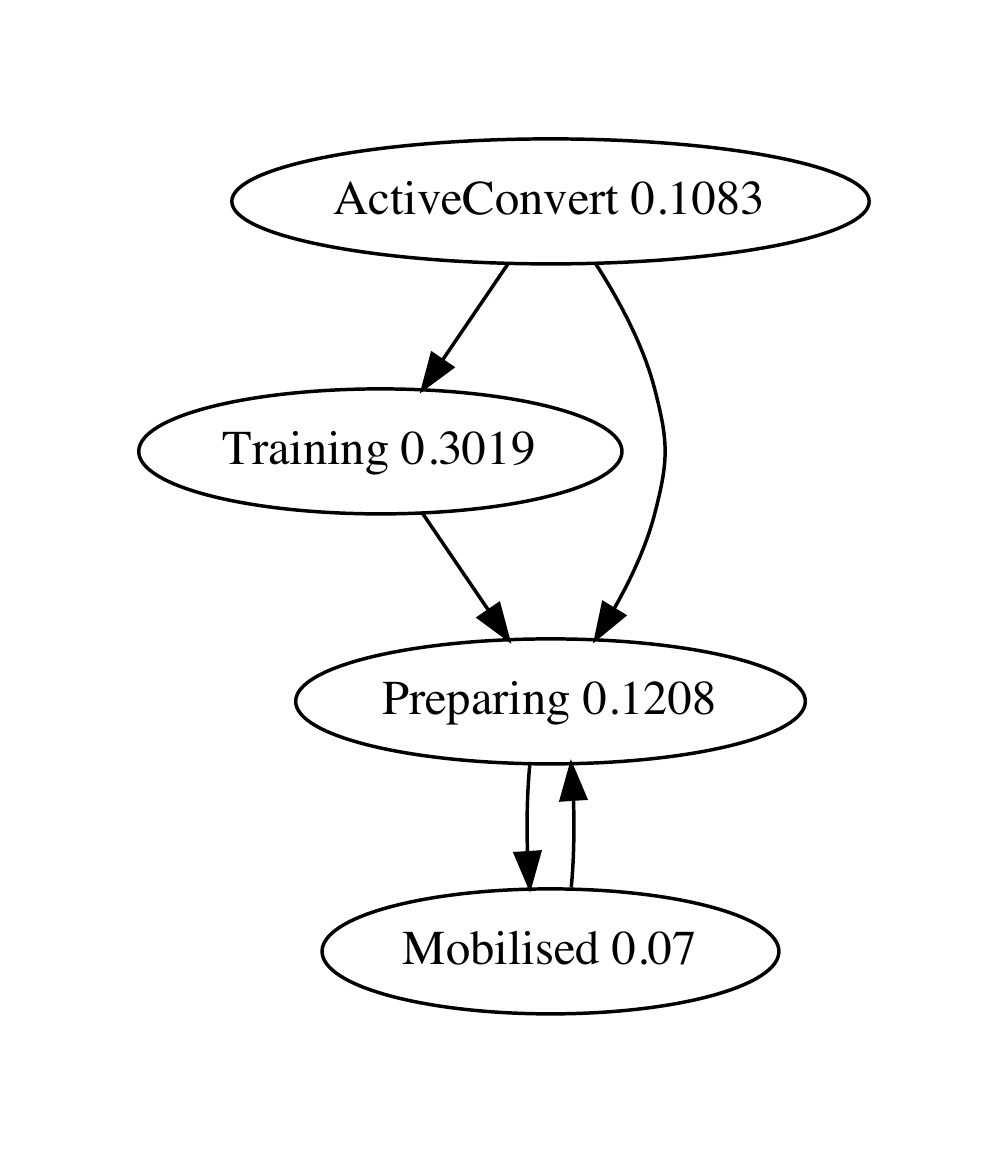}
            \caption{RDCEG with final posterior probabilities based on Scenario \ref{scen4}}
            \label{rd4}
        \end{subfigure}
    \caption{Position Score and RDCEG for \ref{scen3} and \ref{scen4}}
\end{figure}

We illustrate the propagation of probabilities through the model based on scenarios of simulated data. 
We use the same framework as above 
and manually set the routinely observed data $\boldsymbol{Y_t}$ through 24 weekly time steps
to examine how the currently parameterised model behaves under each scenario. 
\begin{scen} \label{scen3} 
In this scenario the suspect increases their web visits to target locations from week 8 and their physical visits to
target locations from the week 21; they are in constant communication with known radicals and there is an increase
in their finances followed by a decrease in first few weeks, during which time they are seen to be visiting car dealers
electronically and physically. 
They make public and personal threats and in the last weeks of the period the threatening data
increases with a legacy statement and a statement of intent. Figures \ref{pp3}, \ref{score3}, \ref{rd3} show the increase
in threat level resulting from this scenario.
\end{scen}
\begin{scen} \label{scen4}
The suspect's communications and possible training / preparing type data linearly decreases from initial 
levels similar to scenario \ref{scen3} to zero over the 24 weeks. Moreover there
are no threats made during the whole period. Figures \ref{pp4}, \ref{score4}, \ref{rd4} show the decreasing threat level
resulting from this scenario.
\end{scen}
\subsection{Eventual probability of mobilisation} \label{Eventual}
\begin{figure}[H]
    \centering
        \begin{subfigure}[b]{0.475\textwidth}
            \centering
            \includegraphics[width=\textwidth]{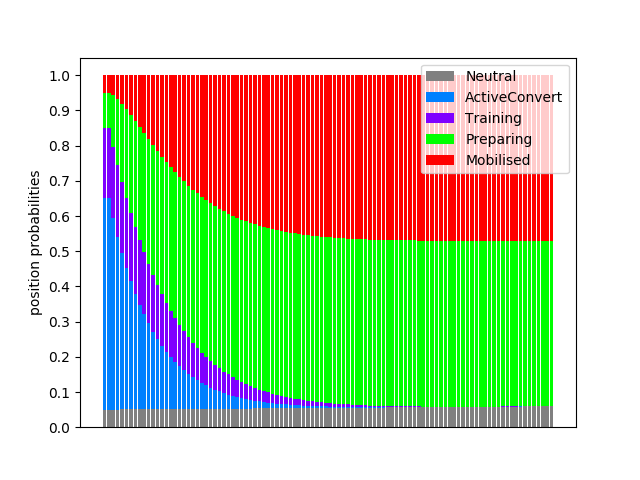}
            \caption{Position probabilities over 10 thousand periods based purely on semi-Markov transition matrix}
            \label{transprobs10000}
        \end{subfigure}
        \hfill
        \begin{subfigure}[b]{0.475\textwidth}
            \centering
            \includegraphics[width=\textwidth]{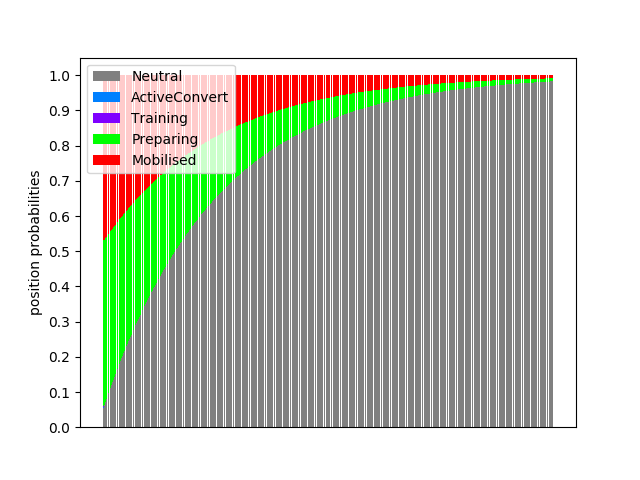}
            \caption{Position probabilities over 2 million periods based purely on semi-Markov transition matrix}
            \label{transprobs20000000}
        \end{subfigure}
        \vskip\baselineskip
        \begin{subfigure}[b]{0.475\textwidth}
            \centering
            \includegraphics[width=\textwidth]{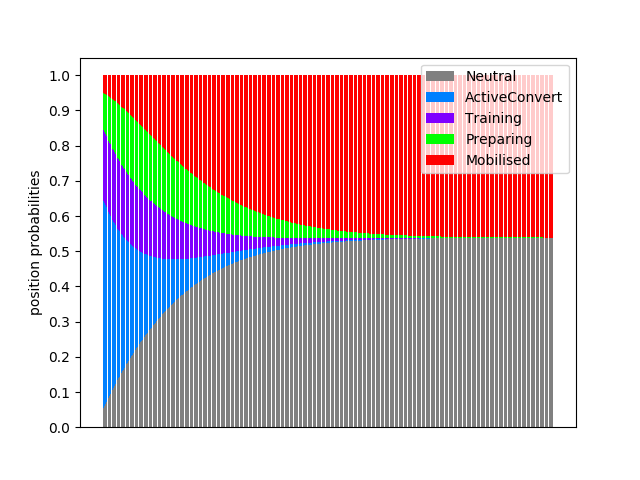}
            \caption{Position probabilities with two absorbing states over one thousand periods 
                based purely on semi-Markov transition matrix; 
                holding time transition rate: $\zeta_i(t,t')=0.01$;
                edge probabilities $m_{a,n}{=}m_{t,n}{=}m_{p,n}{=}0.3$}
            \label{transprobBiAbs1}
        \end{subfigure}
        \hfill
        \begin{subfigure}[b]{0.475\textwidth}
            \centering
            \includegraphics[width=\textwidth]{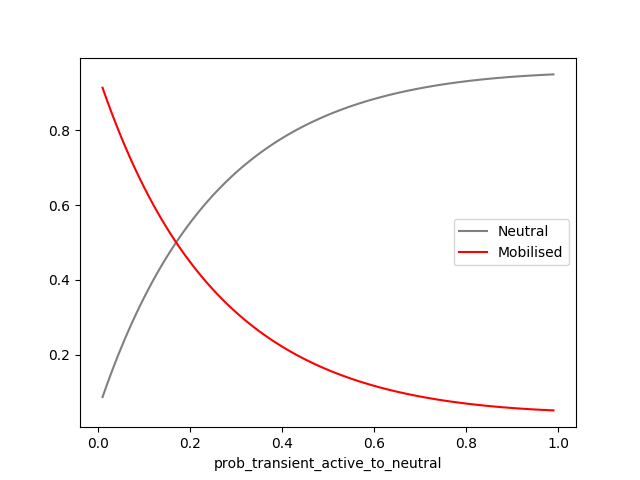}
            \caption{Chart showing the two absorbing states' stationary probabilities varying by 
                    directly setting the per-period probability of transition from each transient active 
                    state to neutral from $0.01$ to $0.99$}
            \label{transprobBiAbs2}
        \end{subfigure}
    \caption{Figures for long-term probability of mobilisation}
\end{figure}

For any individual suspect or group of suspects the medium to long-term probabilities of mobilisation 
is of key interest and can aid as a model diagnostic tool.
We can estimate this by using the semi-Markov transition matrix to evolve the current probabilities. 
The RDCEG in this section has the neutral state
as the single absorbing state hence asymptotically the probability of this state will go to one; However in the practical
medium term we can examine the behaviour of the active positions including the mobilised state. Under the configuration of 
the priors and the edge probabilities as in Figure \ref{rdcegprior} and with the holding time distribution $\zeta_i(t,t')$ set
to a constant $0.01$ for the set time period of $t'-t$ equal to one week (as used in the examples above), 
the qualitative behaviour of the RDCEG's
state probabilities can be seen in Figures \ref{transprobs10000} and \ref{transprobs20000000}. Figures \ref{transprobBiAbs1}
and \ref{transprobBiAbs2} show the long term behaviour under alternative specifications
where the mobilised position is another absorbing state: the suspect once having mobilised and 
executed the attack
cannot transition to any of the other states including the neutral state. The reasoning behind this latter
configuration is an assumption that once the individual has mobilised this entails an attack and the end of
this particular police case. 
\section{Model diagnostics} \label{modeldiag}
\subsection{Robustness to RDCEG structure specification}
The graph representation of the RDCEG, that is the set of states and the directed edges between them, form the structure
of the RDCEG that is meant to faithfully represent the possible pathways of individuals towards, in this application, 
acts of terrorism. The actual structure chosen is based on historical cases, existing research by criminologists and 
discussions with practioners and is predicated on the assumption that the states are \say{self-identifiable} 
that is that the actual individual would be able to place themselves in one of these states at any given time. 
\\\\
Assuming such an approach is valid, without being able to actually look into an individual's mind, we are 
liable to mis-specify 
the structure: for example construct states that could meaningfully and usefully be split into finer sub-states, or  
have a set of states that should be collapsed into one state; or have edges where they should not exist or have edges 
missing. Whether the set of states are \say{correct to the individual's mind} is arguably less 
relevant for our purposes than whether the set of states are a useful discretisation of the individual's potential pathway 
from the investigators' perspective; and for this we can be directly guided. 
\\\\
It is still of use to analyse the sensitivity of behaviour and results to changes in the structure chosen. 
To this end we examine the effect of using different sets of states and different edges by using the same data sets 
with different RDCEG structures.
We coursen the RDCEG used in Section \ref{vehicle} by collapsing the \say{Training} state into the \say{Preparing} state, 
and then separately refine the RDCEG by splitting the \say{Preparing} state in two states: \say{Preparing for a 
Vehicle Attack} and \say{Preparing for a Bomb Attack}, so that we have two new alternative RDCEG structures to 
compare with the original. We expand the task sets and data and run
two new data scenarios involving a potential joint vehicle and bomb attack on these two new structures along with 
the original structure giving in total six sets of results.
Results for this analysis along with details of the scenarios are given in Appendix \ref{appendix_rob} and 
Appendix \ref{appendix_scens}.
The impacts on the posterior probabilities are as expected: the \say{Neutral} state's probability is relatively 
unchanged and the probability mass of the coarsened state is roughly the sum of the finer states in each 
example under both scenarios.
\\\\
In contrast to our elicited RDCEG structure, \cite{shenvi2019} construct an RDCEG structure explicitly from 
data on individuals within an open population using a hierarchical clustering algorithm and Bayes Factor based model
selection in health care settings: recurrent falls in the elderly in the context of care services and drug effects on 
early epilepsy cases.
\subsection{Sensitivity analysis}
To improve our understanding of the behaviour and robustness of the model to subjective inputs, and to identify key 
parameters that merit extra analysis to determine their optimal value, we
perform sensitivity analysis against a base scenario. The model is high dimensional in the sense that the number 
of configurable parameters is large: so for the time being this analysis has focussed on the state prior probabilities 
and the holding distributions the latter of which are key to determining the speed of transition. 
\\\\
We tested the sensitivity of the evolution
of the state probabilities applying both moderate and large shifts to the state priors. We  
performed a similar analysis shifting the holding time distribution parameter $\zeta_{i,j}$ which represents the probability
of a transition from state $w_i$ to $w_j$ in a given time period (here one week) given that an edge exists from $w_i$ 
to $w_j$. 
\\\\
For moderate changes in 
priors, increasing the prior for a less threatening state has an initial effect but is outweighed by any data that 
indicates threat; whilst an increase in prior for a threat state accelerates the effect of threat data. 
For extreme increases in prior for the active states the prior does dominate the evolution; but for the \say{Neutral} 
state the initial very high prior is subsequently outweighed by data. See Appendix \ref{appendix_sens} for figures.
\section{Discussion}
In this paper we have described a novel three level hierarchical model that utilises at its deepest level
an RDCEG modelling the state of a suspect within the stages of a potential attack.
We illustrated how such an analysis can
synthesise information concerning that suspect, through sets of tasks to produce snap shot
summaries of the likely position of this person and the current threat they
might present.
\\\\
Currently, working with various domain experts, we are in the process of
constructing a suite of RDCEG templates and their associated tasks
\cite{smithassault2018}. These describe different criminal processes
associated with assaults or violence against the general public, indexed by
type of crime, that build on existing criminological
models. This type of technology has already been well
developed for BNs within the context of forensic science \cite
{Aitken,Dawid2} and has established frameworks of processes linking
activities with evidence. The structure we use here helps in this
development because only the top layer of the hierarchy usually needs
regular refreshing: the possible positions and associated tasks are fairly
stable over time. We hope that within this paper we have illustrated
that, just as in forensic science, such methods are both promising and
feasible. Indeed the harmonisation of this class of models to forensic
analogues means that evidence applied within an investigation can be
coherently integrated into case reports associated with criminal proceedings
if the suspect does attempt to perpetrate a crime. 
\\\\
In the next phase of this programme, building on these models of individual
suspects, we are developing a network model for the stochastic evolution of open populations
of violent criminals. This issue is complicated by the fact that many suspects are working
in teams and often coordinated. This dependence structure and communications between indivuduals 
therefore have to be carefully modelled for such models to be realistic. However this more
challenging domain is also a potentially very fertile one - where standard
estimation of hyperparameters associated with different units and the
Bayesian selection of the most promising models can begin to be applied. The
challenge is that series are frequently disrupted so the systematic
estimation of their hyperparameters is hard and needs strong prior
information to be effective.
\\\\
The eventual objective is for the hierarchical model for individuals described in this paper and the network model
described in the paragraph above, to be incorporated into a decision theoretic framework that also models the action 
space and objectives of the authorities. This would aim to aid investigators decide which actions to take; such as
which cases to prioritise, or to deprioritise to free up resources, which cases to increase surveillance on, 
and which cases to make which kind of interventions on;
all in order to minimise the expectation of a multi-attribute loss function over incidents, casualties, public terror 
etc., given the constraints of resources, personal freedom, democratic legitimacy and proportionality.
\\\\
Moreover the actions of the authorities in terms of preventative measures, defence of targets and 
pursuit methods will influence the actual decisions and trajectories of the suspects as is
documented in case reports (see for example \cite{gill}). These recursive aspects
introduce game-theoretic ideas into the models which complicate the inferential process and probability
propagation. This work has to be done in conjunction with
the authorities in order to properly represent a realistic, constrained action space, 
to elicit a realistic loss function, and to gain insight on the dynamic interplay between investigator and investigated.
\\\\
Thus the work we present in this paper is the first phase of a longer programme working with practitioners
that eventually aims to provide a theoretically valid, practically useful, and legally defensible system 
to support the prevention of acts of extreme radical violence. 
\section*{Acknowledgements}
This work was funded by The Alan Turing Institute Defence and Security Project G027. We are grateful for the comments of two
anonymous referees and the associate editor which greatly improved this paper.
\newpage
\appendix
\appendixpage
\section{Task probabilities given position} \label{AppendixProbs}
Some settings that keep elicitation of task probabilities to the minimum
whilst appearing to provide good discriminatory power in most circumstances
are given below. 
For each state $w_{i},$ $i=0,1,2,\ldots ,m$, let 
\begin{equation*}
p_{t,i}^{I^{\ast}(w_i)}\triangleq P\left\{ \boldsymbol{\theta}_{I_{+}^{\ast}(w_i)}=\mathbf{1},
\boldsymbol{\theta}_{I_{-}^{\ast}(w_i)}=\mathbf{0}
\boldsymbol{|}w_{i},\mathcal{F}_{t-}\right\} 
\end{equation*}
be the probability that all the positive tasks and none of the negative tasks are being done given the position is $w_i$.
For definitions of $I^{\ast}(w_i), I_{+}^{\ast}(w_i), I_{-}^{\ast}(w_i)$ 
see Equations \ref{index_set_defn} in Section \ref{structure}.
We define and elicit tasks in such a way that 
\begin{equation*}
p_{t,i}^{I^{\ast}(w_i)} \geq 0.2
\end{equation*}
i.e. given someone is in active position $w_{i}$ then the probability
that $\omega $ in $w_{i}$ is engaged in \emph{all} the tasks that are
positive indicators and none of the tasks that are negative indicators is non-negligible. We also assume
that the less the engagement in the positive tasks the smaller the probability $
\omega $ lies in this active state and the contrary for the negative tasks.
\\
\\
Let $r_{i}^{-}$ and $r_{i}^{+}$ 
denote the cardinality of the sets
$I_{-}^{\ast }(w_{i})$ and $I_{+}^{\ast }(w_{i})$. 
Suppose $\emptyset \subseteq A\subseteq I_{+}^{\ast }(w_{i})$ where $0 \leq r(A) \leq r_{i}^{+}$ is
the cardinality of $A$
and $\emptyset \subseteq B\subseteq I_{-}^{\ast }(w_{i})$ where $
0 \leq r(B) \leq r_{i}^{-}$ is the cardinality of $B$.
Then 
\begin{equation}
0 \leq K(A,B) \triangleq r(A) + r_{i}^{-} - r(B)  \leq  r_{i}^{+} + r_{i}^{-}
\end{equation}
so $K$ is the number of tasks
being done in $I_{+}^{\ast}(w_i)$ plus those not being done in $I_{-}^{\ast}(w_i)$.
Let the log odds be: 
\begin{equation}
\phi _{t,i}^{I^{\ast}(w_i)}\triangleq \log p_{t,i}^{I^{\ast}(w_i)}-\log \left\{ 1-p_{t,i}^{I^{\ast}(w_i)}\right\}
\end{equation}
Then under the Naive Bayes assumption of Section \ref{elicitation} for $w_0$, we set
\begin{equation}
\phi _{t,0}^{I^{\ast}(w_i)}=\sum_{k\in I^{\ast}(w_i)}\phi _{t,0}^{k}
\end{equation}
Then the full set of probabilities for any such $A$ and $B$ can be generated by interpolating:
\begin{equation*}
\phi _{t}^{i,A,B}=\alpha _{K(A,B),i}\phi _{t,i}^{I^{\ast
}(w_{i})}+(1-\alpha _{K(A,B),i})\phi _{t,0}^{I^{\ast }(w_{i})}
\end{equation*}
\begin{equation*}
\alpha _{K(A,B),i}=\left( \frac{K(A,B)}{r_{i}^{+}+r_{i}^{-}}\right) ^{\xi _{i}}
\end{equation*}%
where $\xi _{i}>0$ is constrained such that the resulting probabilities sum to one. 
This setting has the property that preserves the sorts of
monotonicity we require for the comment above.
\section{The semi-Markov transition matrix and resulting recurrence equations} \label{AppendixSemiMarkov}
We now explicitly model the process $W_t$ as a continuous time process, still with discrete state space. We still assume
that we observe data at discrete sequential times: $(t_i)$, $i \in \mathbb{N}$  but we model explicitly that the 
suspect may transition between states between or at these observation times.
Thus the discrete time process $W_{t_n}$ is embedded in a continuous time semi-Markov process 
$W_t$, $t \in \mathbb{R^+}$.
\\
\\
Assume that the time interval is short so that at most one transition has occurred between 
$t$ and $t^{\prime}$. Conditioning on the event that there has only been one transition, let 
$\zeta_{i}\left( t,t^{\prime}\right) $ represent the probability of a transition from $w_{i}$ during the time interval 
$\left( t,t^{\prime}\right]$, where $\zeta_i$ is the holding distribution for the $i^{th}$ position.
Let $M_{ij}^{0}$ represent the probability that the suspect will transition to the position 
$w_{j}$ from position $w_{i}$ given there has been exactly one state transition.
Then the components of the transition matrix are given by:
\begin{equation*}
M_{ij}(t,t',M^0_{i,j},\zeta_i)=\left\{ 
\begin{array}{c}
1-\zeta _{i}\left( t,t^{\prime}\right) \text{, }i=j \\ 
\zeta _{i}\left( t,t^{\prime}\right) M_{ij}^{0}\text{, }i\neq j
\end{array}
\right. 
\end{equation*}
Using the RDCEG of the vehicle attacker in Section \ref{vehicle}, we have $M^0$ and $M(t,t')$ as in Table
\ref{transitionMatrix} with $m_{i,j}$ denoting edge probabilities: i.e. the probability of transition
from the source vertex $w_i$ to the destination vertex $w_j$ conditional on a transition having
occurred.
\begin{center}
\begin{table}[h]
\begin{footnotesize}
\begin{tabular}{lccccc}
\toprule
$M^0$ &      Neutral &    ActiveConvert &    Training &    Preparing &     Mobilised \\
\midrule
N           &  0         &  0 &  0 &  0 &  0 \\
A           &  $m_{a,n}$ &  0 &  $m_{a,t}$ &  $m_{a,p}$ &  0 \\
T           &  $m_{t,n}$ &  0 &  0 &  $m_{t,p}$ &  0 \\
P           &  $m_{p,n}$ &  0 &  0 &  0 &  $m_{p,m}$ \\
M           &  $m_{m,n}$ &  0 &  0 &  $m_{m,p}$ &  0 \\
\bottomrule
$M(t,t')$ & Neutral &    ActiveConvert &    Training &    Preparing &     Mobilised \\
\midrule
N           &  1         &  0 &  0 &  0 &  0 \\
A           &  $\zeta_a(t,t')m_{a,n}$ &  $ 1 - \zeta_a(t,t')$ &  $\zeta_a(t,t')m_{a,t}$ &  $\zeta_a(t,t')m_{a,p}$ &  0 \\
T           &  $\zeta_t(t,t')m_{t,n}$ &  0 &  $1-\zeta_t(t,t')$ &  $\zeta_t(t,t')m_{t,p}$ &  0 \\
P           &  $\zeta_p(t,t')m_{p,n}$ &  0 &  0 &  $1-\zeta_p(t,t')$ &  $\zeta_p(t,t')m_{p,m}$ \\
M           &  $\zeta_m(t,t')m_{m,n}$ &  0 &  0 &  $\zeta_m(t,t')m_{m,p}$ &  $1-\zeta_m(t,t')$ \\
\bottomrule
\end{tabular}
\caption{Semi-Markov Transition Matrix}
\label{transitionMatrix}
\end{footnotesize}
\end{table}
\end{center}
Under the assumption that the observation times are frequent enough that at most one transition could occur between them,
the filtering equations to update the position probabilities given the change in time and the observed data are 
\begin{samepage}
\begin{eqnarray}
p(w_{i,t_{n+1}}|Z_{t_{n+1}}) \propto p(w_i,{t_{n+1}}|Z_{t_{n}}) \times \\
\sum_{\theta_{I^{\ast}(w_i)}\in\{0,1\}^{|I^{\ast}(w_i)|}} p(\theta_{I^{\ast}(w_i)}|w_{i})p(Z_{I^{\ast}(w_i),t_{n+1}}|\theta_{I^{\ast}(w_i)}) \label{z_recurrence_semi_data} \\
p(w_{i,t_{n+1}}|Z_{t_{n}}) = \sum_{j=0}^d p(w_{j,{t_n}}|Z_{t_{n}}) M_{j,i}(t_{n},t_{n+1},M^0,\zeta) \label{z_recurrence_semi}
\end{eqnarray}
\end{samepage}
\\
\\
Equation (\ref{z_recurrence_semi_data}) updates the position probabilities given the observed data at time $t_{n+1}$ whilst Equation (\ref{z_recurrence_semi}) transitions the position probabilities over a small time interval $(t_{n},t_{n+1})$ 
according to the transition matrix $M(t_{n},t_{n+1},M^0,\zeta)$.
Now we ease the assumption that observation times are frequent enough that we can exclude the possibility
of more than one state transition between them.
For ease of exposition and notation we assume that the intervals between observation times are a multiple
of $|t-t'|$, ie 
\begin{equation*}
t_{n+1}-t_n = k |t'-t| 
\end{equation*}
for every $t_n$ and for some $k \in \mathbb{N}$.
Here we take the holding time distributions $\zeta_i(t,t')$
as time-homogenous although in future work we plan to use inhomogenous distributions. With homogeneity the transition matrix between observation times is then simply the appropriate power of $M(t,t')$,
i.e.
\begin{equation*}
M(t_n,t_{n+1},M^0,\zeta) = M^k(t,t',M^0,\zeta)
\end{equation*}
Then Equation (\ref{z_recurrence_semi}) can be written as:
\begin{equation}
p(w_{i,t_{n+1}}|Z_{t_{n}}) = \sum_{j=0}^d p(w_{j,{t_n}}|Z_{t_{n}}) (M^k(t,t',M^0,\zeta))_{i,j} \label{z_recurrence_semi_2}
\end{equation}
The prior and posterior log odds formulae used in Equations 8 and 10 of the main paper  
are amended to be based on $Z$ rather than $Y$
and to function as filtering recurrence equations based
on $Z_t$. Suppressing $\mathcal{F}_{t-}$ to simplify the notation we obtain:
\begin{equation*}
\rho _{it}\triangleq \log \left( \frac{p\left( w_{i}|\bm{Z}_{I^{\ast }(w_{i})t_n}\right) }{p\left( w_{0}|\bm{Z}_{t_n}\right) }\right) ,\; \rho
_{it}^{\ast }\triangleq \log \left( \frac{p\left( w_{i}|\bm{Z}_{I^{\ast }(w_{i})t_{n+1}}\right) }{p\left( w_{0}|\bm{Z}_{t_{n+1}}\right) }\right)
\end{equation*}
The the loglikelihood ratio of $Z_{t_{n+1}}$ becomes
\begin{equation*}
\lambda _{i}\left( \bm{Z}_{I^{\ast }(w_{i})t_{n+1}}\right) \triangleq 
\log \left( \frac{\sum_{\theta_{I^{\ast}(w_{i})}} p( \bm{Z}_{I^{\ast}(w_{i})t_{n+1}}|\theta_{I^{\ast}(w_{i})} ) p(\theta_{I^{\ast}(w_i)}|w_i) }
                 {\sum_{\theta_{I^{\ast}(w_{i})}} p( \bm{Z}_{I^{\ast}(w_{i})t_{n+1}}|\theta_{I^{\ast}(w_{i})} ) p(\theta_{I^{\ast}(w_i)}|w_0) } \right)
\end{equation*}
and the update formula on the log odds is
\begin{equation}
\rho _{it}^{\ast }=\rho _{it}+\lambda _{i}\left( \bm{Z}
_{I^{\ast }(w_{i})t}\right)  \label{logodds simple for Z}
\end{equation}
\section{Multiple Attack Method Scenarios} \label{appendix_scens}
\begin{figure}
    \centering
        \begin{subfigure}[t]{0.95\textwidth}
        \includegraphics[width=\textwidth]{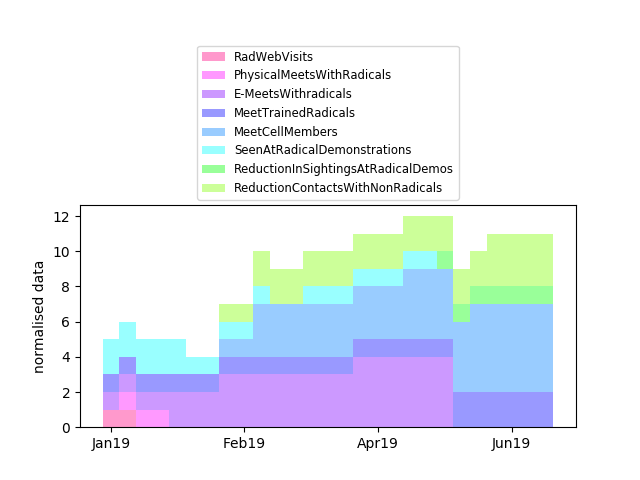}
        \caption{General Data}
        \label{YData1}
        \end{subfigure}
    \centering
        \begin{subfigure}[t]{0.95\textwidth}
        \includegraphics[width=\textwidth]{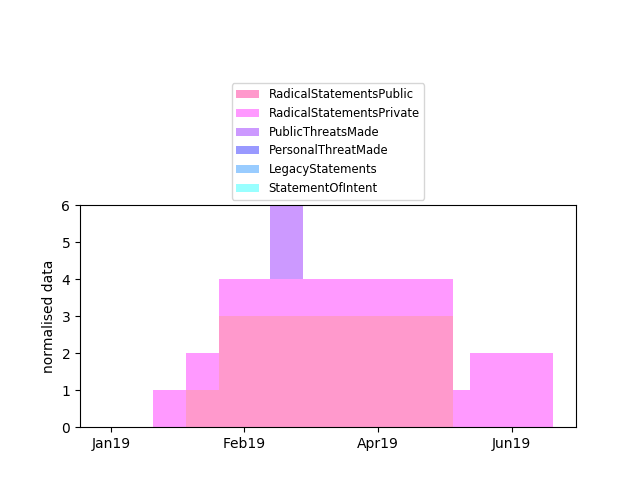}
        \caption{Threat Statement Data}
        \end{subfigure}
    \caption{}
\end{figure}

\begin{figure}
    \centering
        \begin{subfigure}[t]{0.95\textwidth}
        \includegraphics[width=\textwidth]{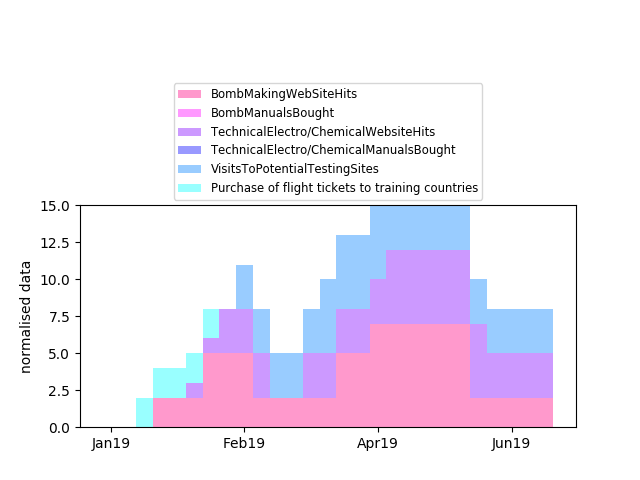}
        \caption{Bomb Data}
        \end{subfigure}
    \centering
        \begin{subfigure}[t]{0.95\textwidth}
        \includegraphics[width=\textwidth]{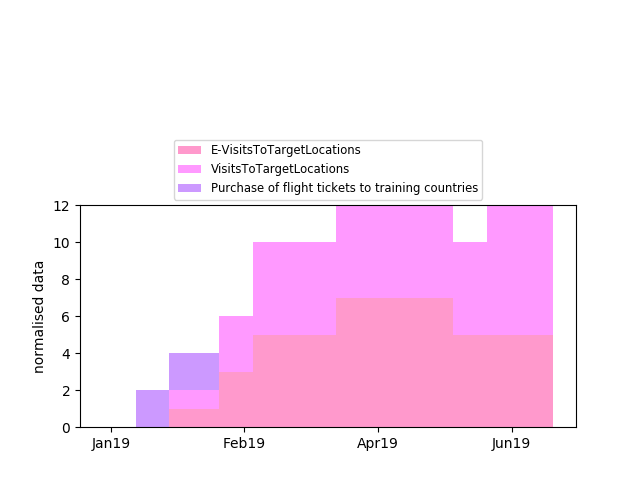}
        \caption{Movement Data}
        \label{YData6}
        \end{subfigure}
    \caption{}
\end{figure}

\begin{figure}
    \centering
    \begin{subfigure}[t]{0.95\textwidth}
        \includegraphics[width=\textwidth]{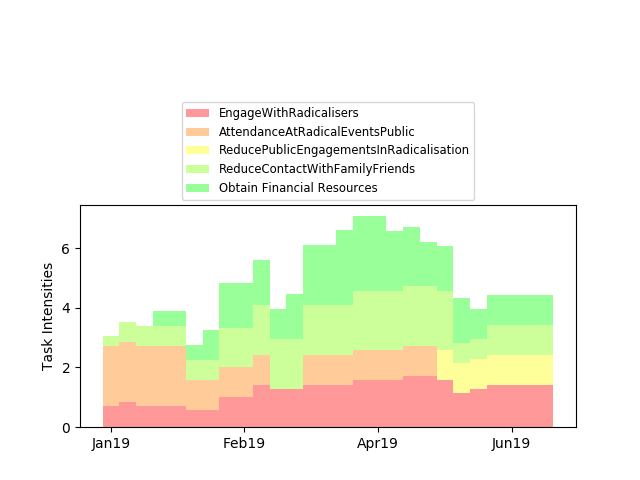}
        \caption{General Task Intensities for scenario A}
        \label{intensities}
    \end{subfigure}
    \centering
    \begin{subfigure}[t]{0.95\textwidth}
        \includegraphics[width=\textwidth]{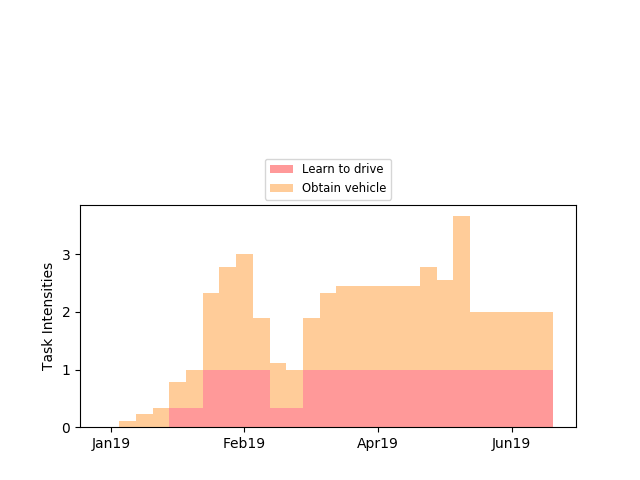}
        \caption{Vehicle Task Intensities for scenario A}
    \end{subfigure}
    \caption{}
\end{figure}

\begin{figure}
    \centering
    \begin{subfigure}[t]{0.95\textwidth}
        \includegraphics[width=\textwidth]{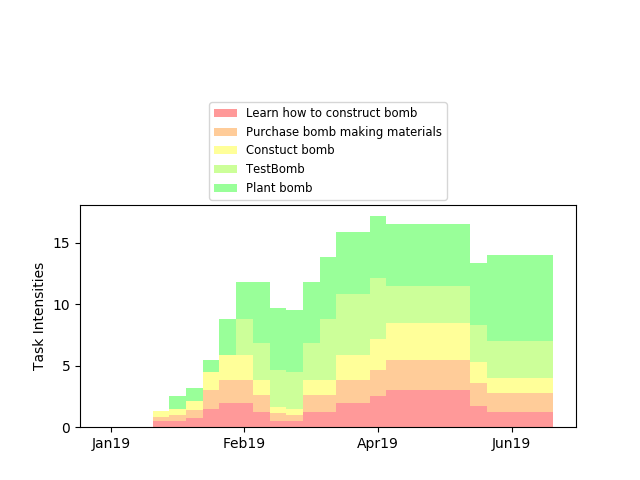}
        \caption{Bomb Task Intensities for scenario A}
    \end{subfigure}
    \centering
    \begin{subfigure}[t]{0.95\textwidth} 
        \includegraphics[width=\textwidth]{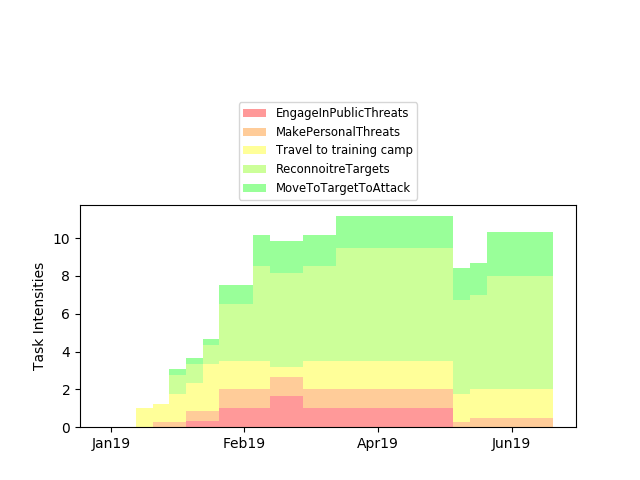}
        \caption{Threat and Mobilisation Task Intensities for scenario A}
        \label{intensities4}
    \end{subfigure}
    \caption{}
\end{figure}

These scenarios are used for the model diagnostic analyses of Appendix \ref{appendix_rob} and \ref{appendix_sens}.
\begin{itemize}
\item Scenario A:
The suspect is seen to make radical statements both in public and privately
including public threats; they increase their meetings with known radicals and
other suspected cell members; they sell assets, increase their financial resources,
hire a vehicle, investigate bomb-making and technical web-sites; they are seen
electronically and physically visiting target locations and are then seen to be
making large financial expenditures. The normalised data is displayed as stacked bar charts in 
Figures \ref{YData1} to \ref{YData6};
and the resulting task intensities are shown in Figures \ref{intensities} to \ref{intensities4}.
\item Scenario B:
The suspect reduces public and private contact and engagement with radicals
and at no time makes and radical statements or threats. 
\end{itemize}
\begin{figure}
    \centering
        \begin{subfigure}[t]{0.95\textwidth}
        \includegraphics[width=\textwidth]{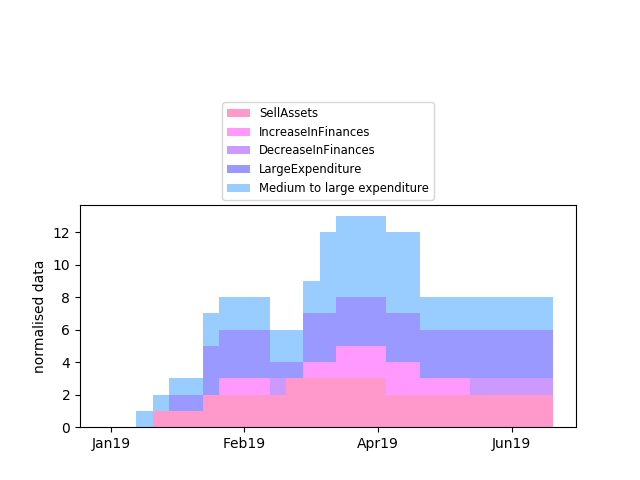}
        \caption{Finance Data}
        \end{subfigure}
    \centering
        \begin{subfigure}[t]{0.95\textwidth}
        \includegraphics[width=\textwidth]{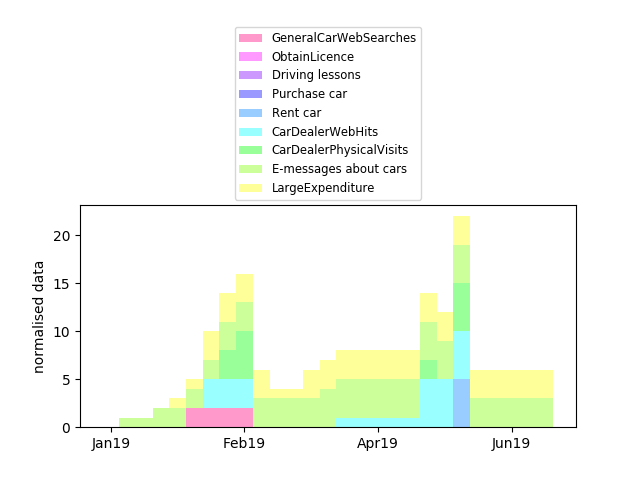}
        \caption{Vehicle Data}
        \end{subfigure}
    \caption{}
\end{figure}
\FloatBarrier
\section{RDCEG robustness} \label{appendix_rob}
\begin{figure} [H]
    \centering
        \begin{subfigure}[t]{0.475\textwidth}
            \centering
            \includegraphics[width=\textwidth]{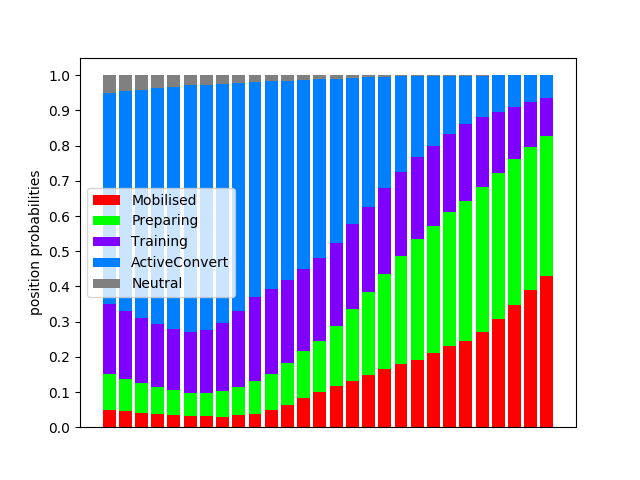}
            \caption{State probabilities through time}
        \end{subfigure}
        \hfill
        \begin{subfigure}[t]{0.475\textwidth}
            \centering
            \includegraphics[width=\textwidth]{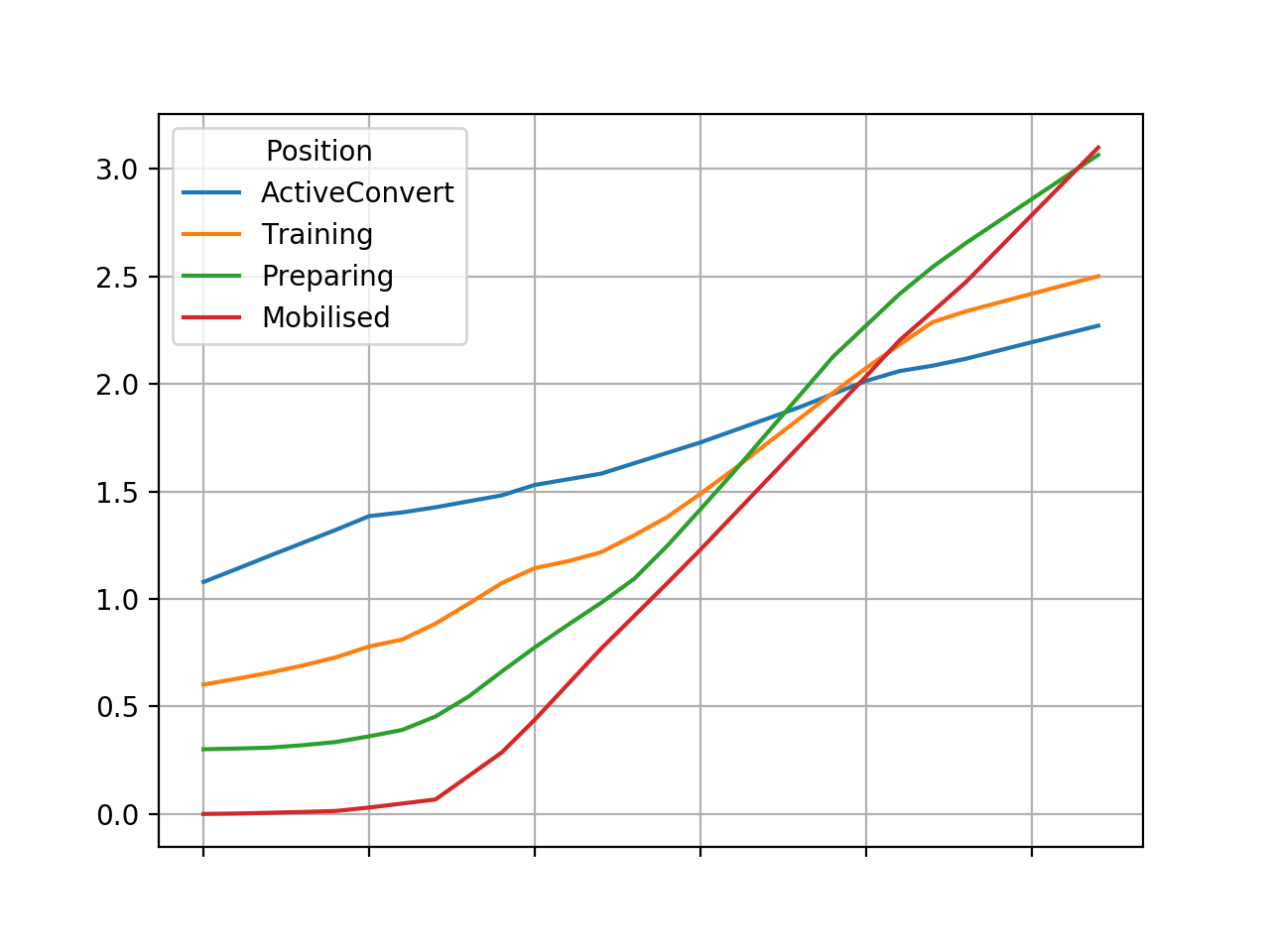}
            \caption{Position score through time}
        \end{subfigure}
        \begin{subfigure}[t]{0.475\textwidth}
            \centering
            \includegraphics[height=5cm]{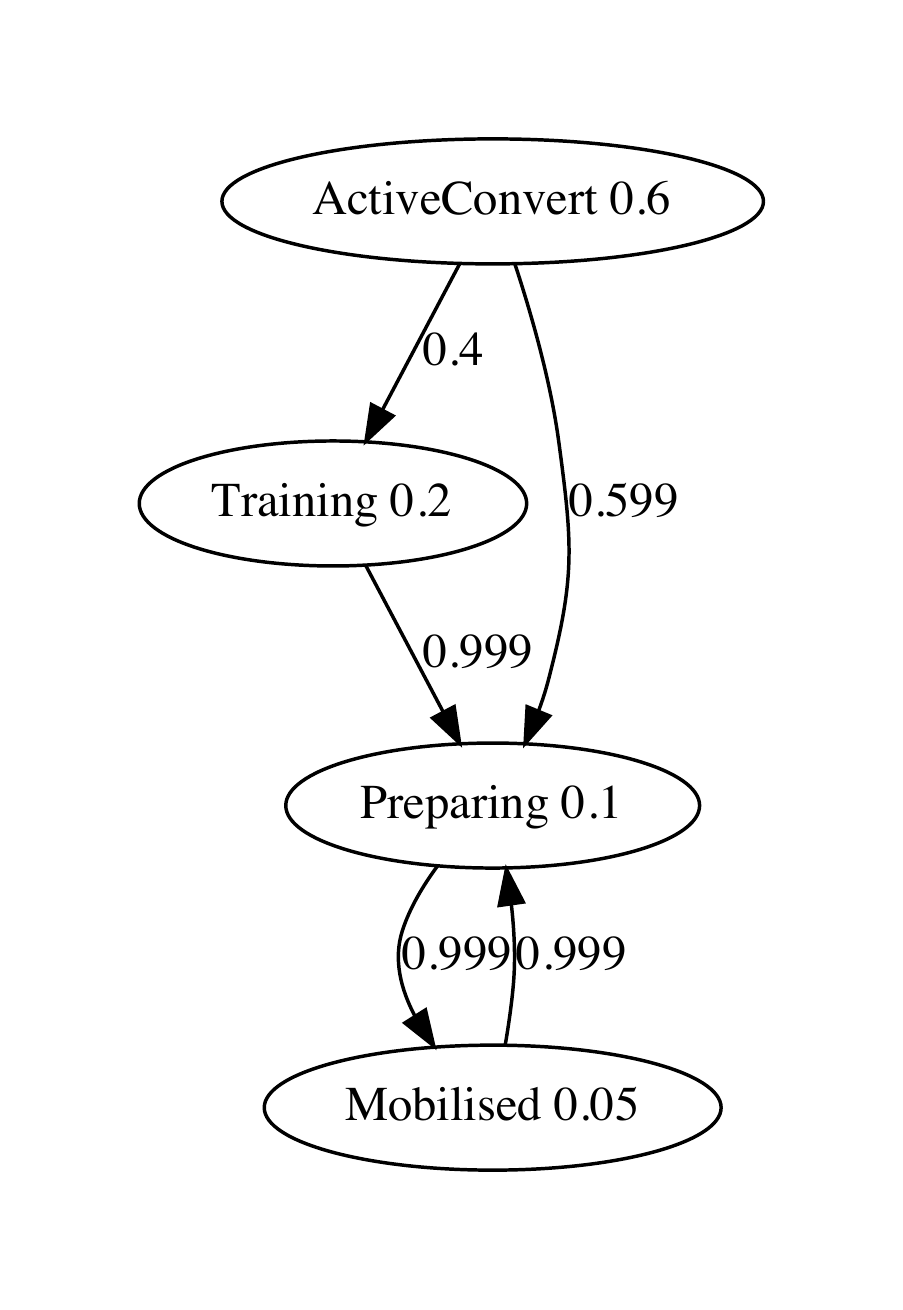}
            \caption{Prior position probabilities}
        \end{subfigure}
        \hfill
        \begin{subfigure}[t]{0.475\textwidth}
            \centering
            \includegraphics[height=5cm]{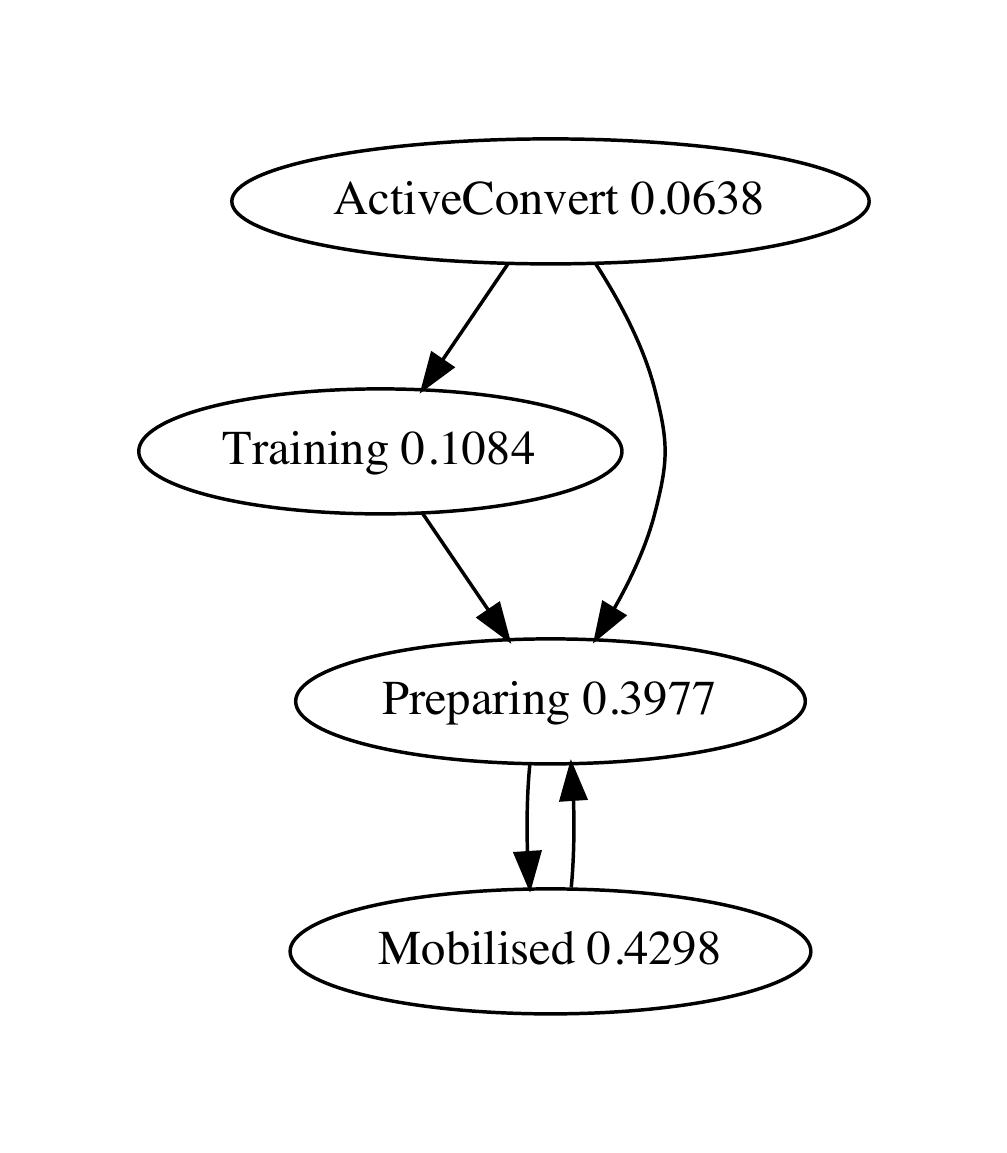}
            \caption{Posterior position probabilities} 
        \end{subfigure}
    \caption{Results for Scenario A using base RDCEG structure}
    \label{base3}
\end{figure}

\begin{figure}
    \centering
        \begin{subfigure}[t]{0.475\textwidth}
            \centering
            \includegraphics[width=\textwidth]{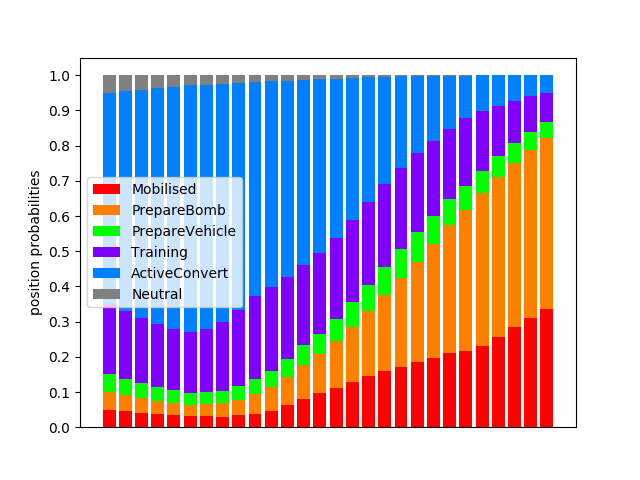}
            \caption{State probabilities through time}
        \end{subfigure}
        \hfill
        \begin{subfigure}[t]{0.475\textwidth}
            \centering
            \includegraphics[width=\textwidth]{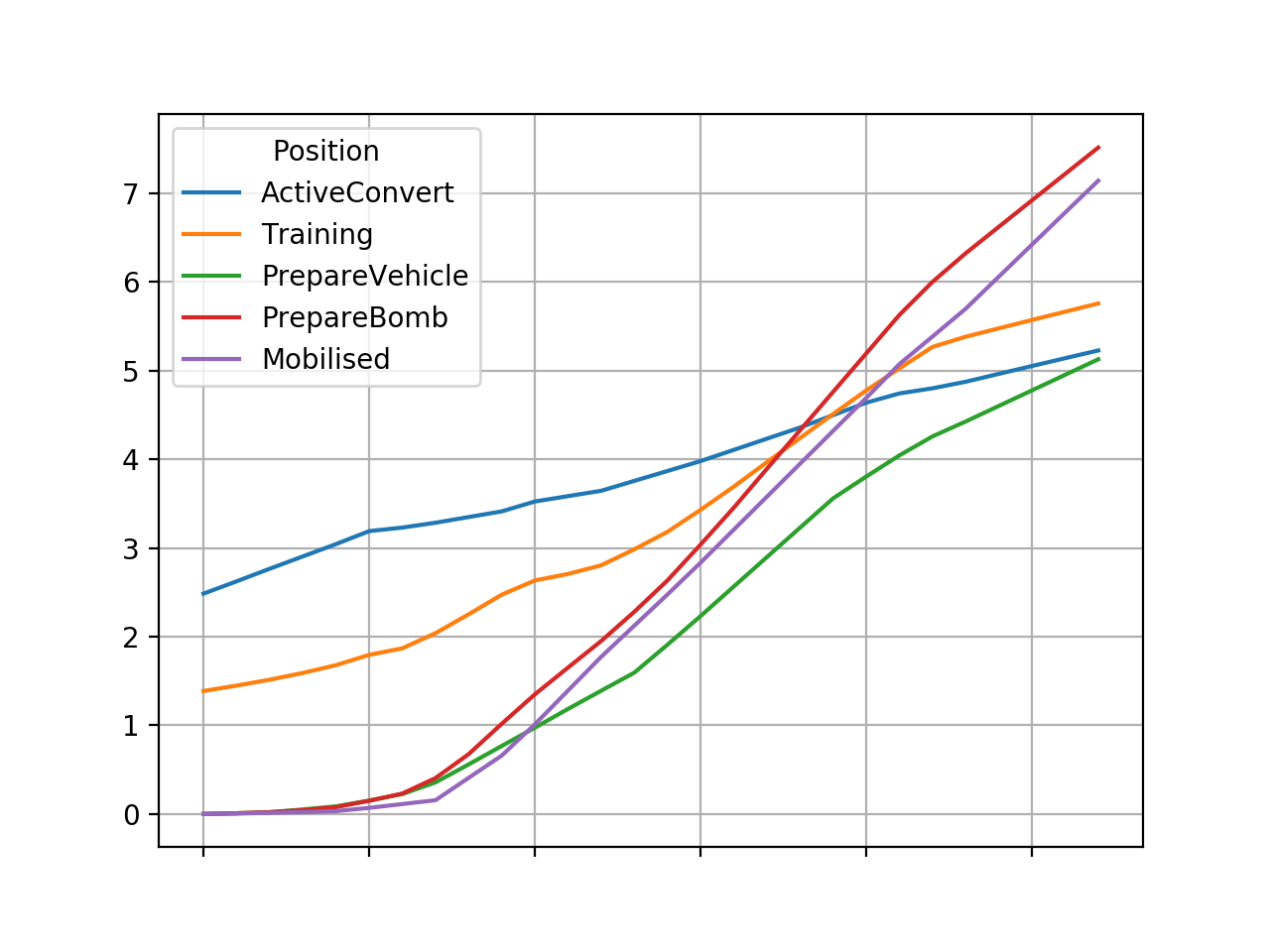}
            \caption{Position score through time}
        \end{subfigure}
        \begin{subfigure}[t]{0.475\textwidth}
            \centering
            \includegraphics[height=5cm]{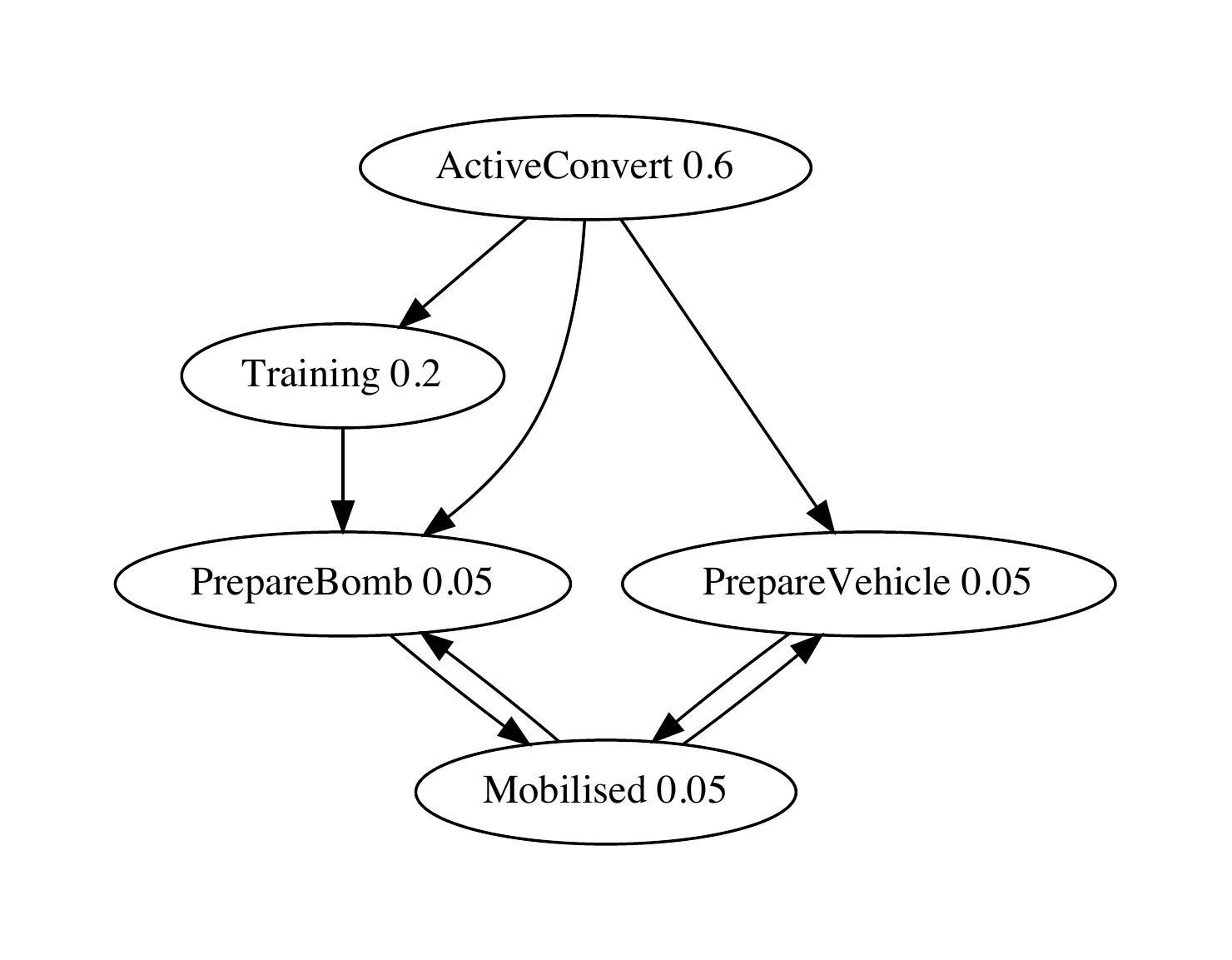}
            \caption{Prior position probabilities}
        \end{subfigure}
        \hfill
        \begin{subfigure}[t]{0.475\textwidth}
            \centering
            \includegraphics[height=5cm]{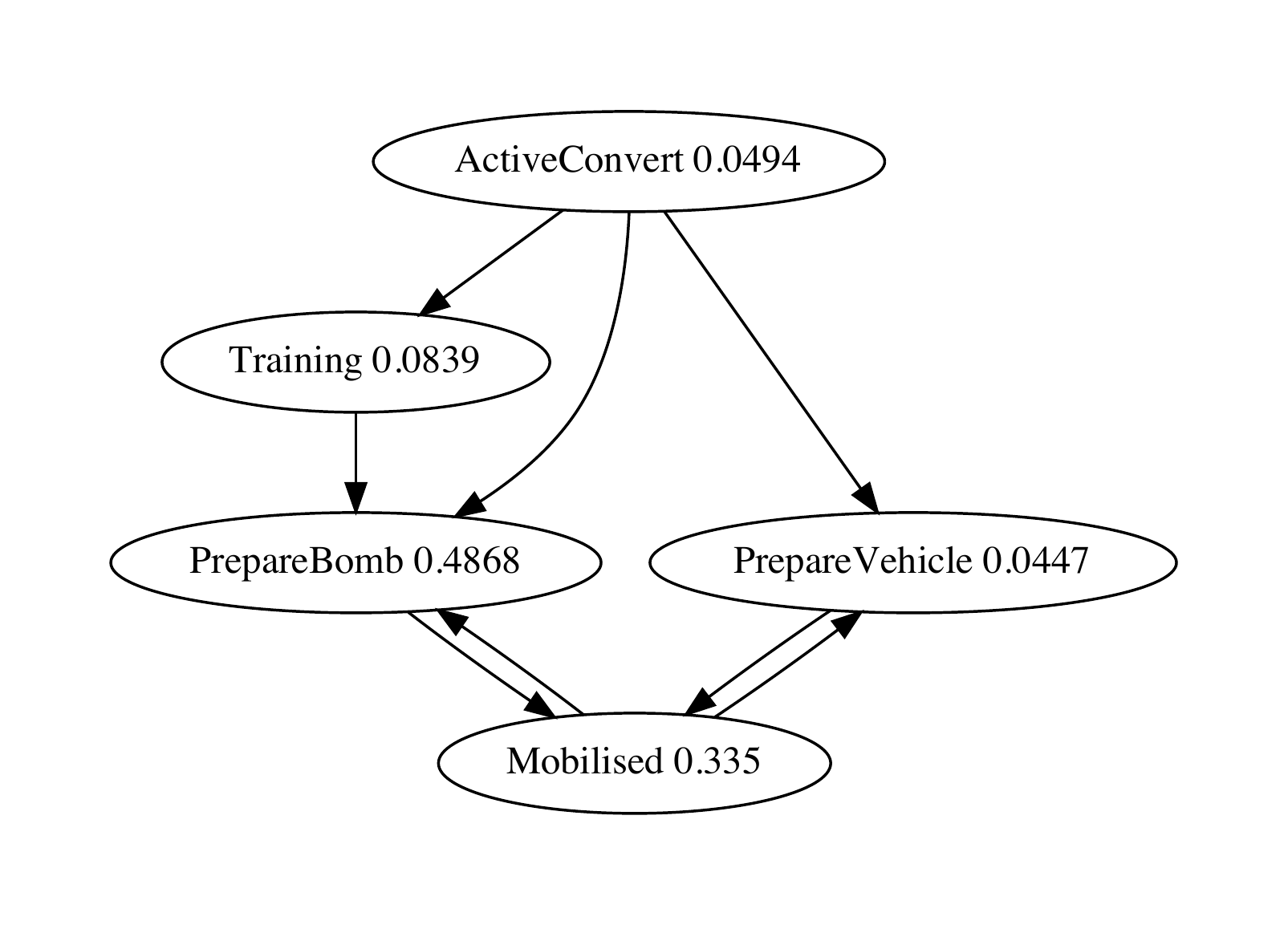}
            \caption{Posterior position probabilities} 
        \end{subfigure}
    \caption{Results for Scenario A using finer RDCEG structure}
    \label{fine3}
\end{figure}

\begin{figure}
    \centering
        \begin{subfigure}[t]{0.475\textwidth}
            \centering
            \includegraphics[width=\textwidth]{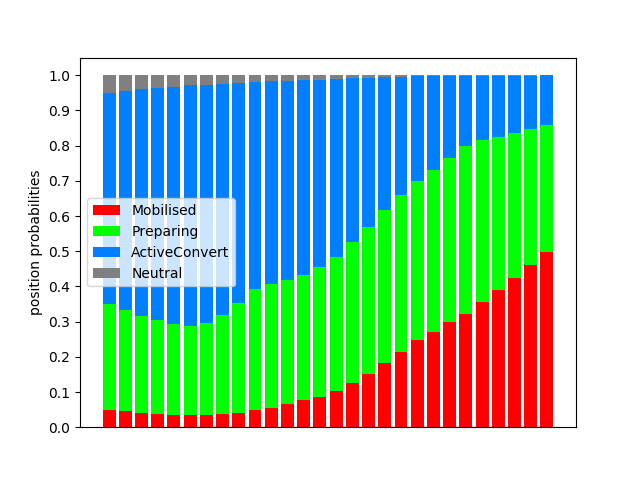}
            \caption{State probabilities through time}
        \end{subfigure}
        \hfill
        \begin{subfigure}[t]{0.475\textwidth}
            \centering
            \includegraphics[width=\textwidth]{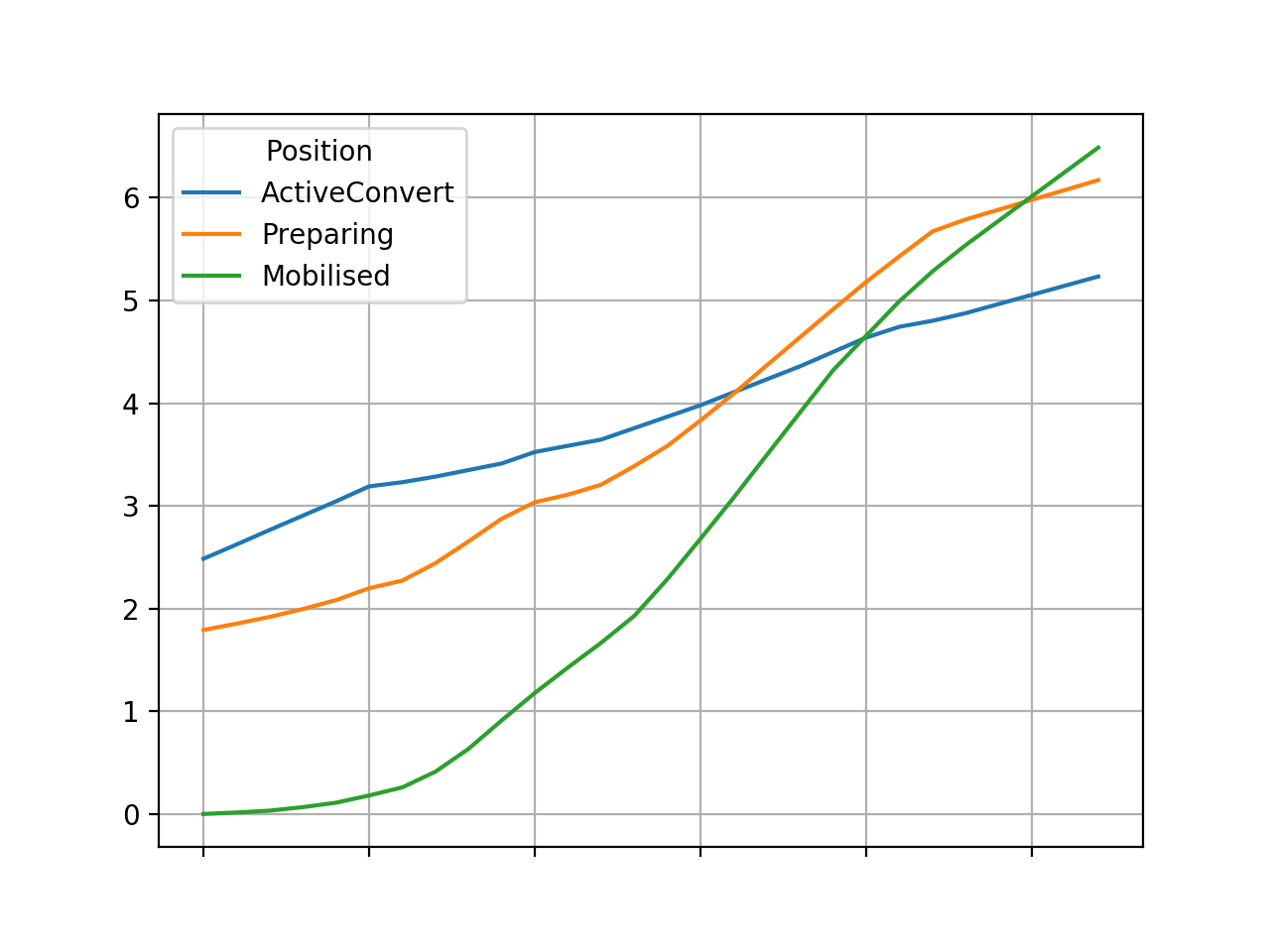}
            \caption{Position score through time}
        \end{subfigure}
        \begin{subfigure}[t]{0.475\textwidth}
            \centering
            \includegraphics[height=5cm]{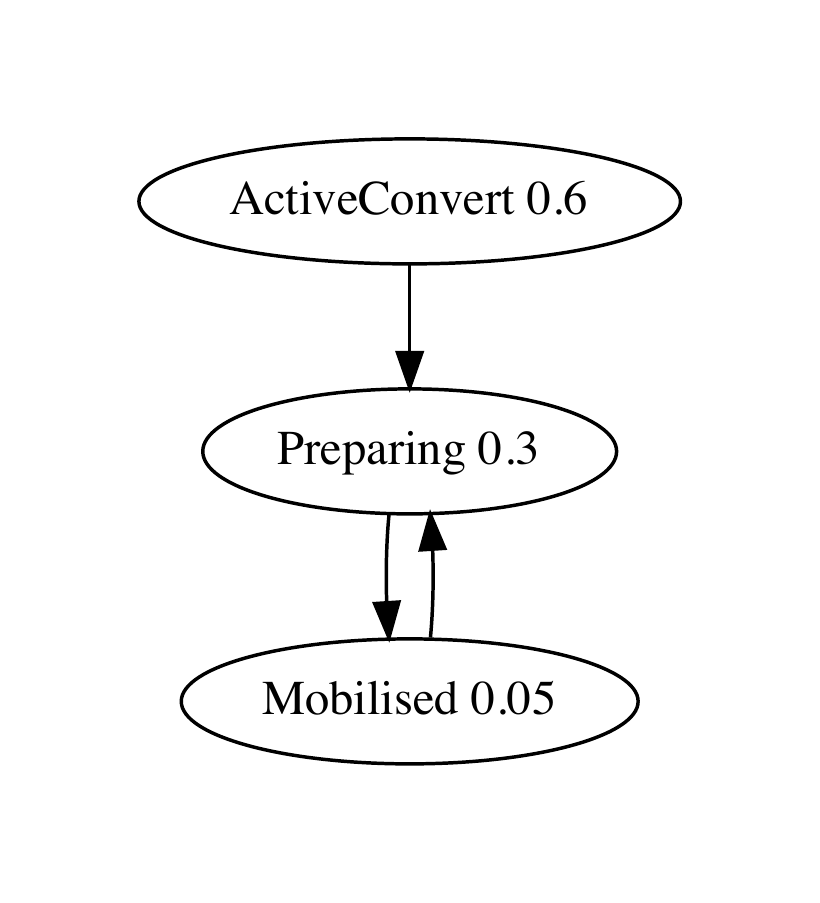}
            \caption{Prior position probabilities}
        \end{subfigure}
        \hfill
        \begin{subfigure}[t]{0.475\textwidth}
            \centering
            \includegraphics[height=5cm]{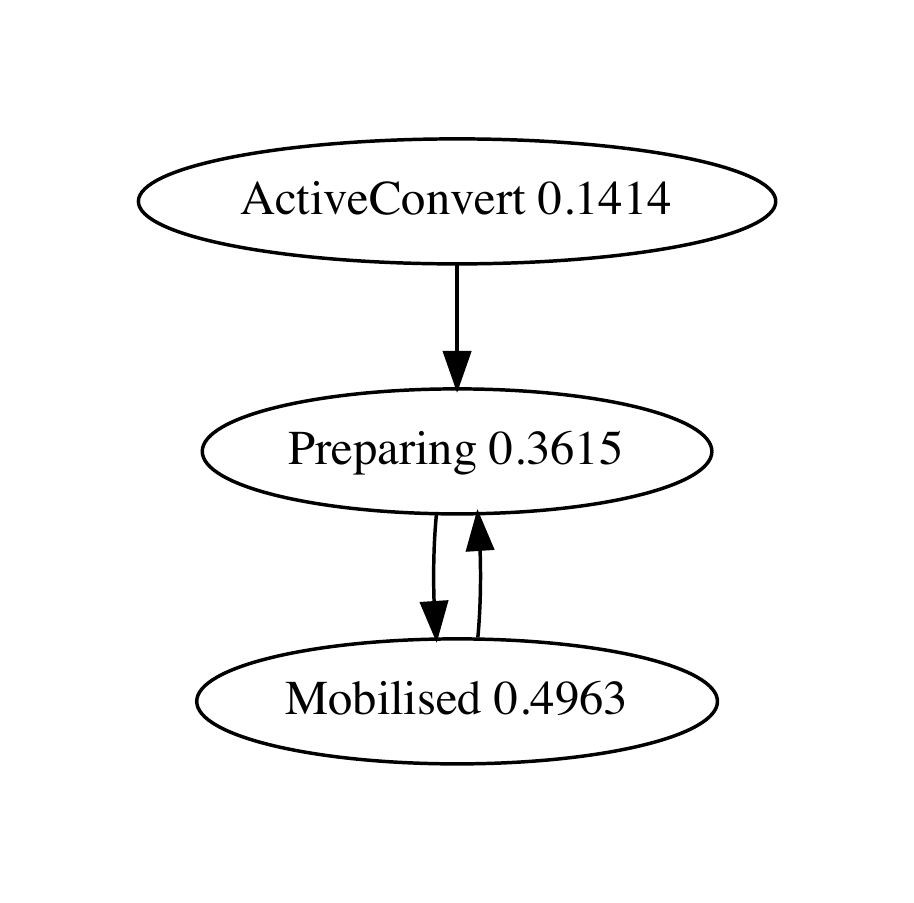}
            \caption{Posterior position probabilities} 
        \end{subfigure}
    \caption{Results for Scenario A using coarser RDCEG structure}
    \label{coarse3}
\end{figure}

\begin{figure}
    \centering
        \begin{subfigure}[t]{0.475\textwidth}
            \centering
            \includegraphics[width=\textwidth]{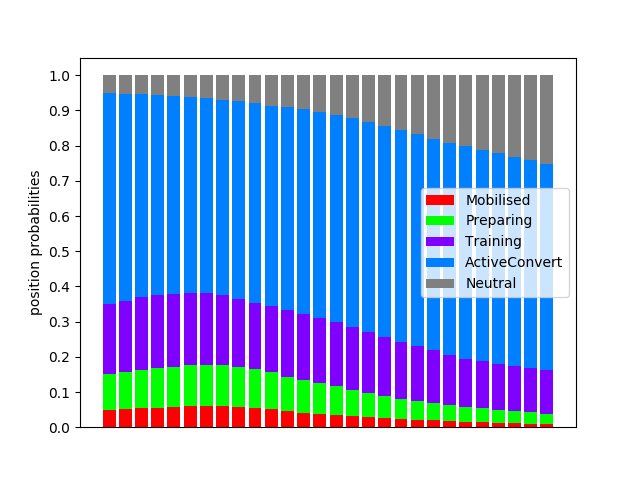}
            \caption{State probabilities through time}
        \end{subfigure}
        \hfill
        \begin{subfigure}[t]{0.475\textwidth}
            \centering
            \includegraphics[width=\textwidth]{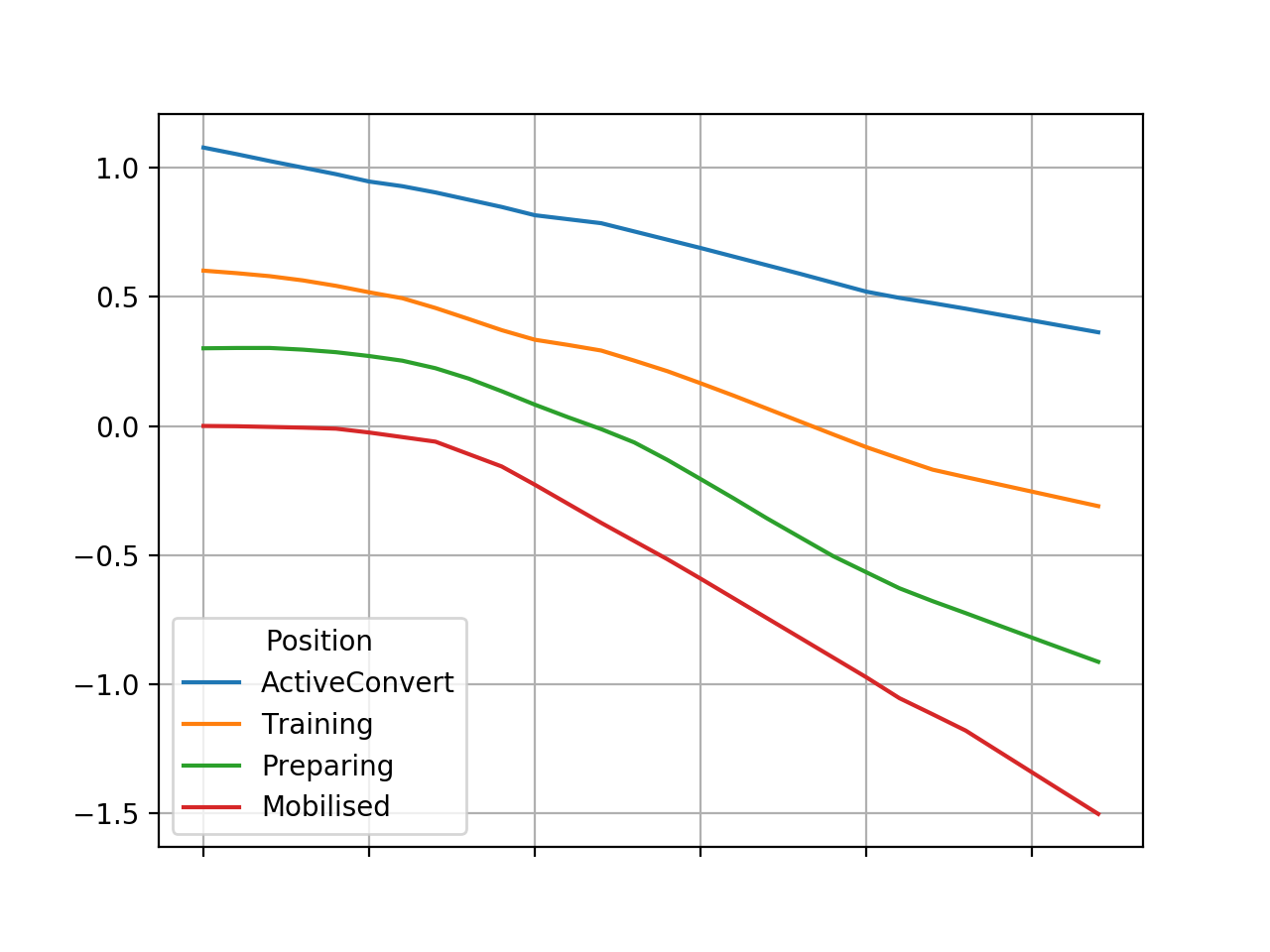}
            \caption{Position score through time}
        \end{subfigure}
        \begin{subfigure}[t]{0.475\textwidth}
            \centering
            \includegraphics[height=5cm]{rdceg_prior.pdf}
            \caption{Prior position probabilities}
        \end{subfigure}
        \hfill
        \begin{subfigure}[t]{0.475\textwidth}
            \centering
            \includegraphics[height=5cm]{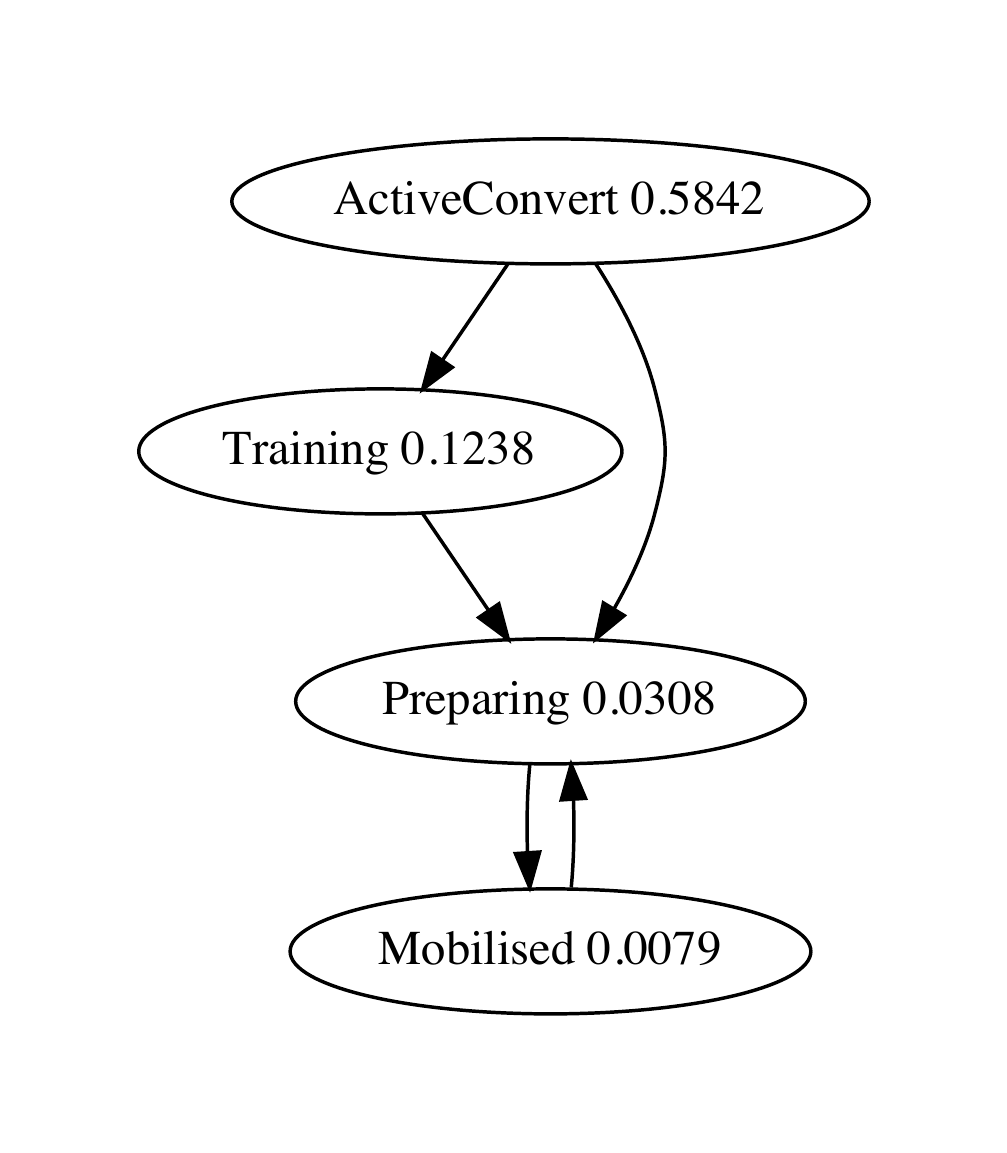}
            \caption{Posterior position probabilities} 
        \end{subfigure}
    \caption{Results for Scenario B using base RDCEG structure}
    \label{base4}
\end{figure}

\begin{figure}
    \centering
        \begin{subfigure}[t]{0.475\textwidth}
            \centering
            \includegraphics[width=\textwidth]{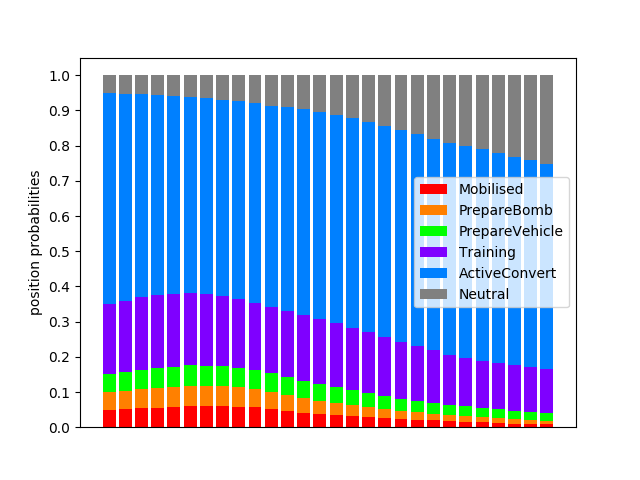}
            \caption{State probabilities through time}
        \end{subfigure}
        \hfill
        \begin{subfigure}[t]{0.475\textwidth}
            \centering
            \includegraphics[width=\textwidth]{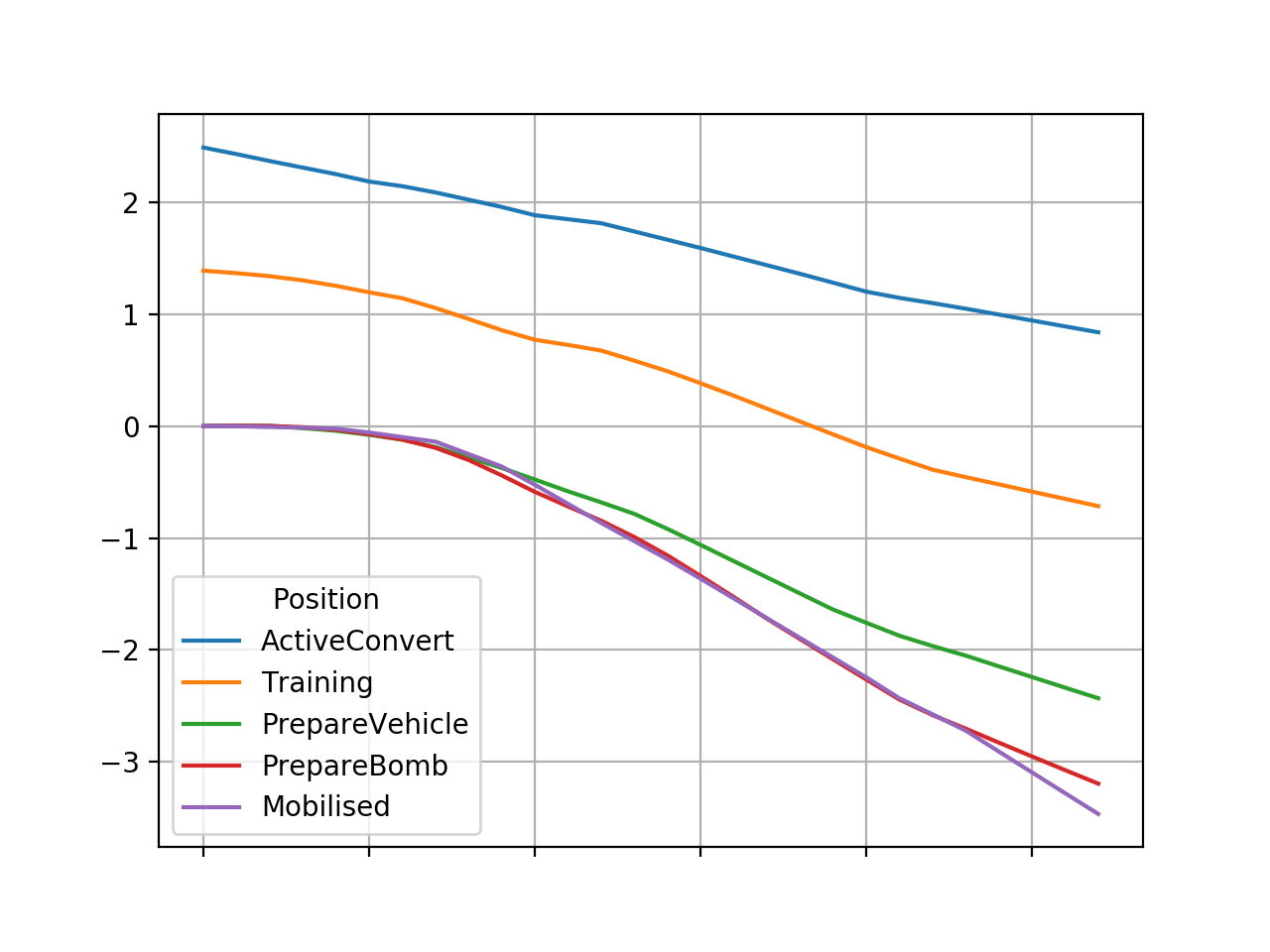}
            \caption{Position score through time}
        \end{subfigure}
        \begin{subfigure}[t]{0.475\textwidth}
            \centering
            \includegraphics[height=5cm]{prior_fine.pdf}
            \caption{Prior position probabilities}
        \end{subfigure}
        \hfill
        \begin{subfigure}[t]{0.475\textwidth}
            \centering
            \includegraphics[height=5cm]{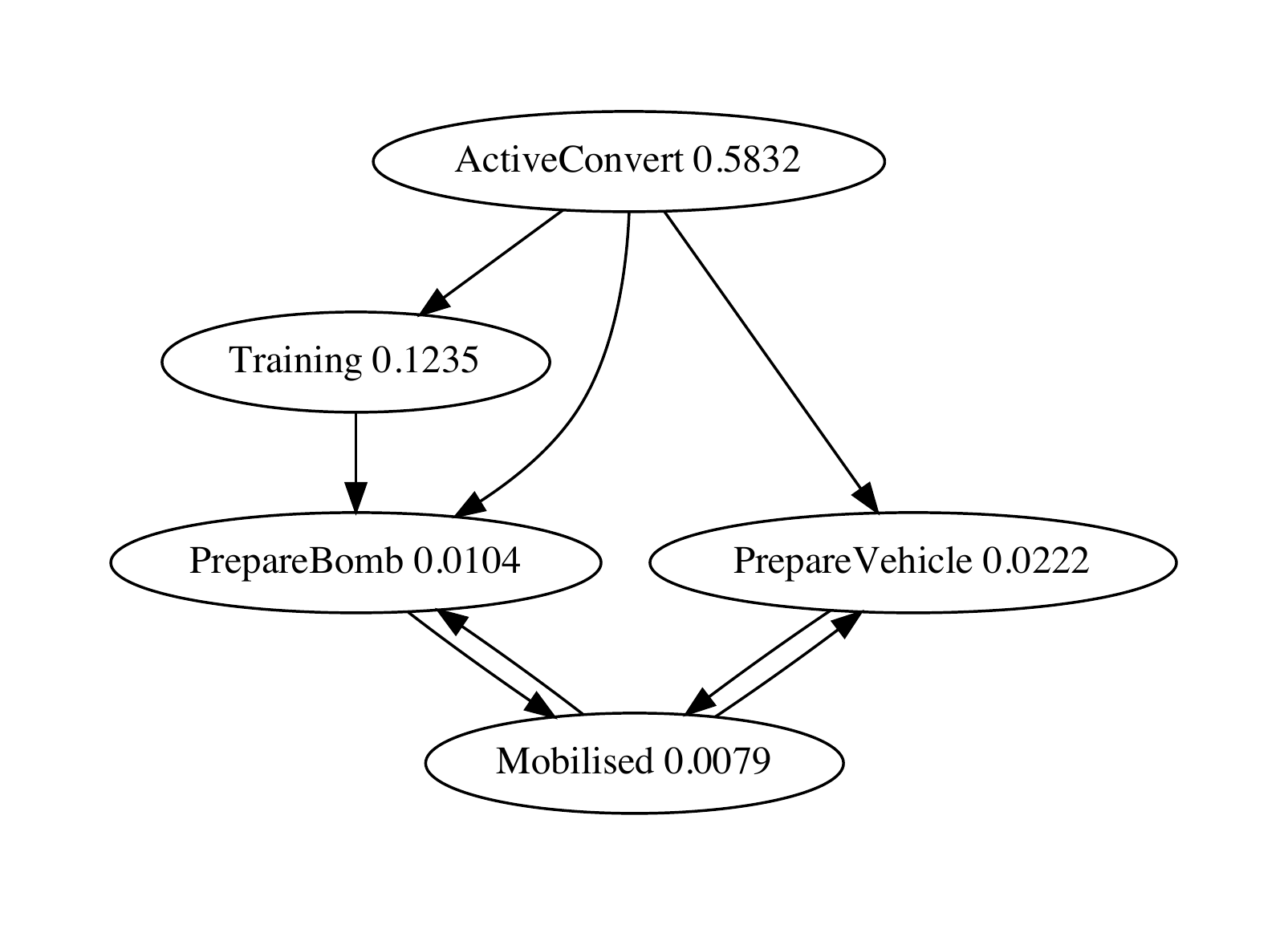}
            \caption{Posterior position probabilities} 
        \end{subfigure}
    \caption{Results for Scenario B using finer RDCEG structure}
    \label{fine4}
\end{figure}

\begin{figure}
    \centering
        \begin{subfigure}[t]{0.475\textwidth}
            \centering
            \includegraphics[width=\textwidth]{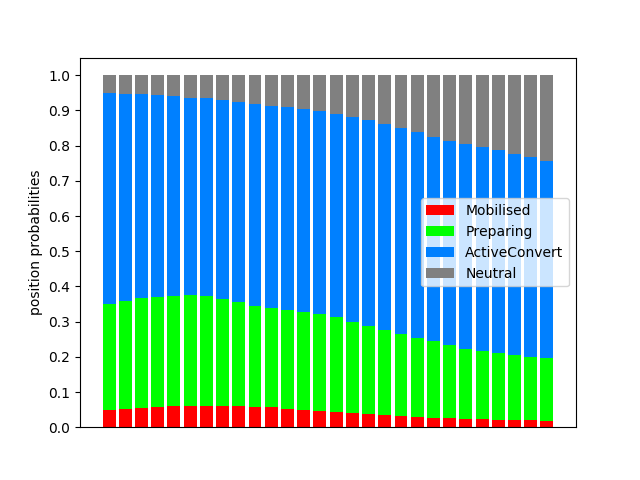}
            \caption{State probabilities through time}
        \end{subfigure}
        \hfill
        \begin{subfigure}[t]{0.475\textwidth}
            \centering
            \includegraphics[width=\textwidth]{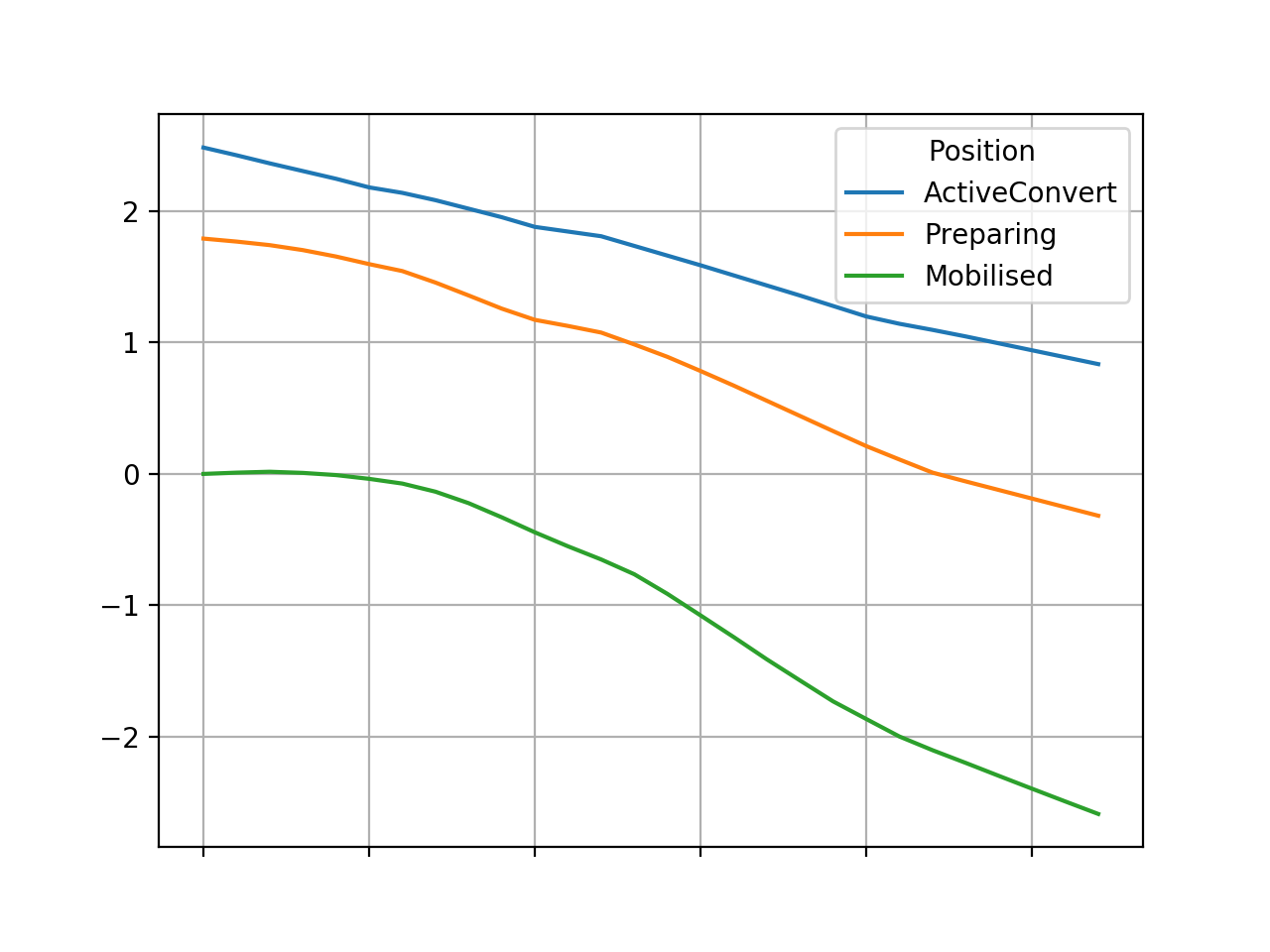}
            \caption{Position score through time}
        \end{subfigure}
        \begin{subfigure}[t]{0.475\textwidth}
            \centering
            \includegraphics[height=5cm]{Prior_coarse.pdf}
            \caption{Prior position probabilities}
        \end{subfigure}
        \hfill
        \begin{subfigure}[t]{0.475\textwidth}
            \centering
            \includegraphics[height=5cm]{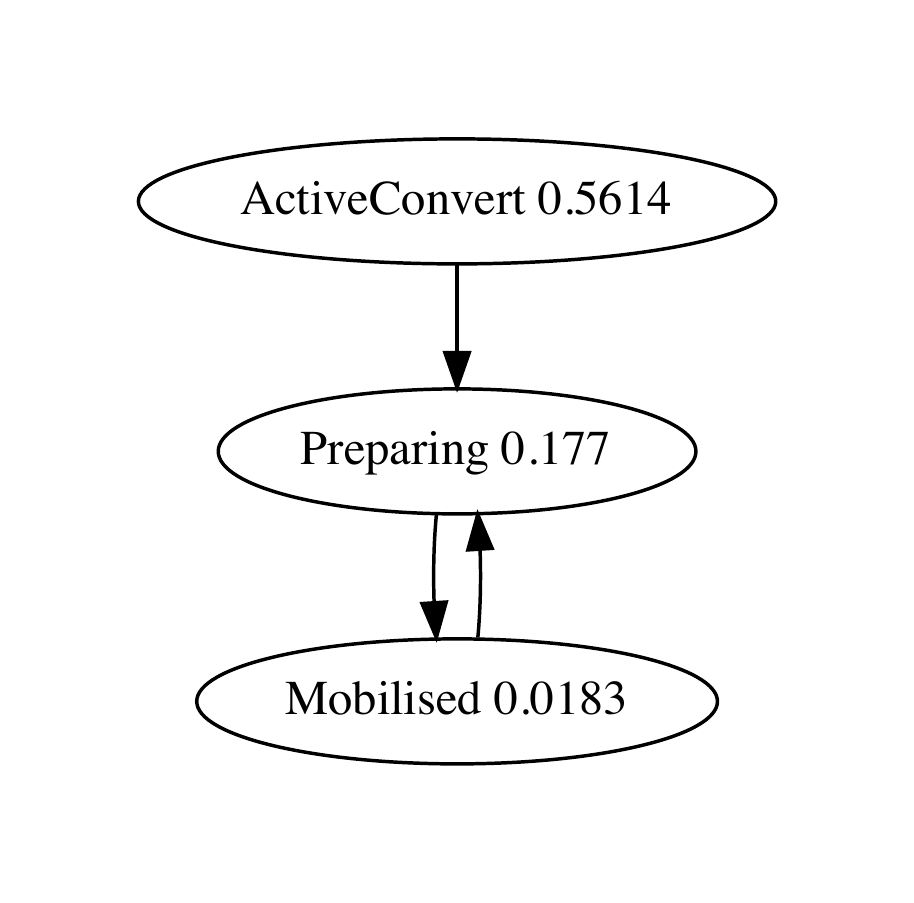}
            \caption{Posterior position probabilities} 
        \end{subfigure}
    \caption{Results for Scenario B using coarser RDCEG structure}
    \label{coarse4}
\end{figure}

\begin{figure}
    \centering
        \begin{subfigure}[t]{0.475\textwidth}
            \centering
            \includegraphics[width=\textwidth]{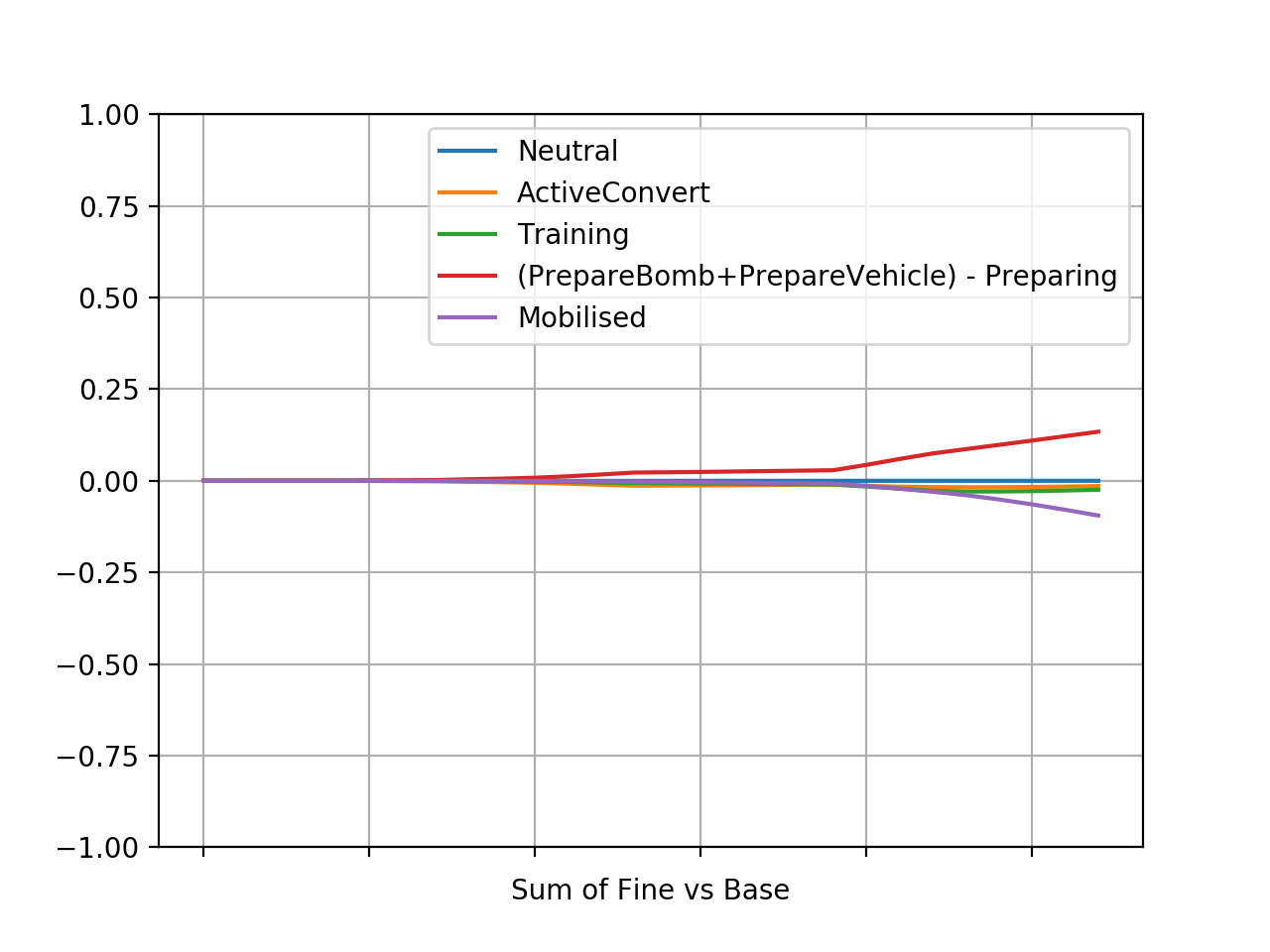}
            \caption{Refined RDCEG minus Base RDCEG state probabilities through time}
        \end{subfigure}
        \hfill
        \begin{subfigure}[t]{0.475\textwidth}
            \centering
            \includegraphics[width=\textwidth]{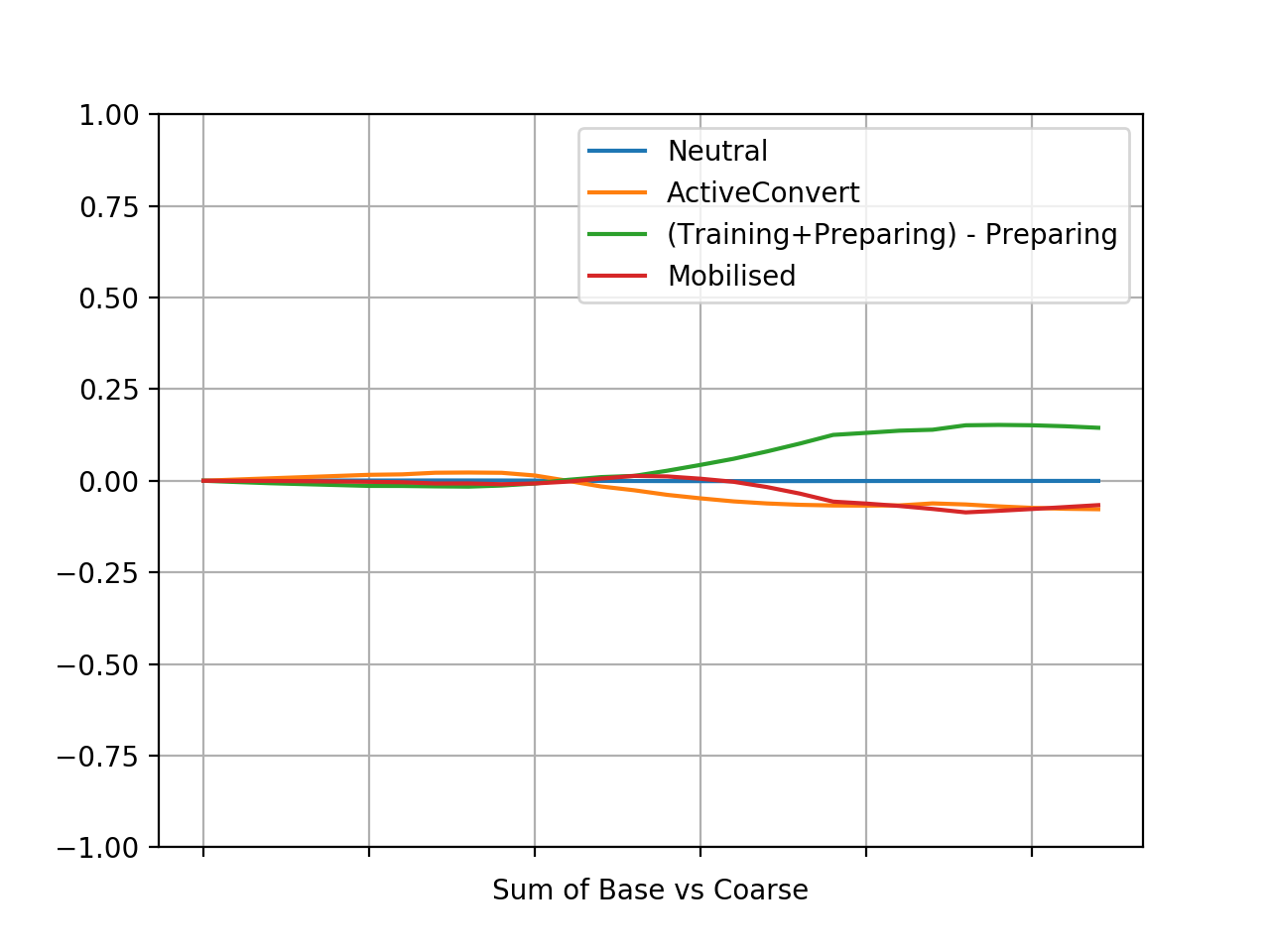}
            \caption{Base RDCEG minus Coarse RDCEG state probabilities through time}
        \end{subfigure}
        \centering
        \hfill
        \begin{subfigure}[t]{0.475\textwidth}
            \centering
            \includegraphics[width=\textwidth]{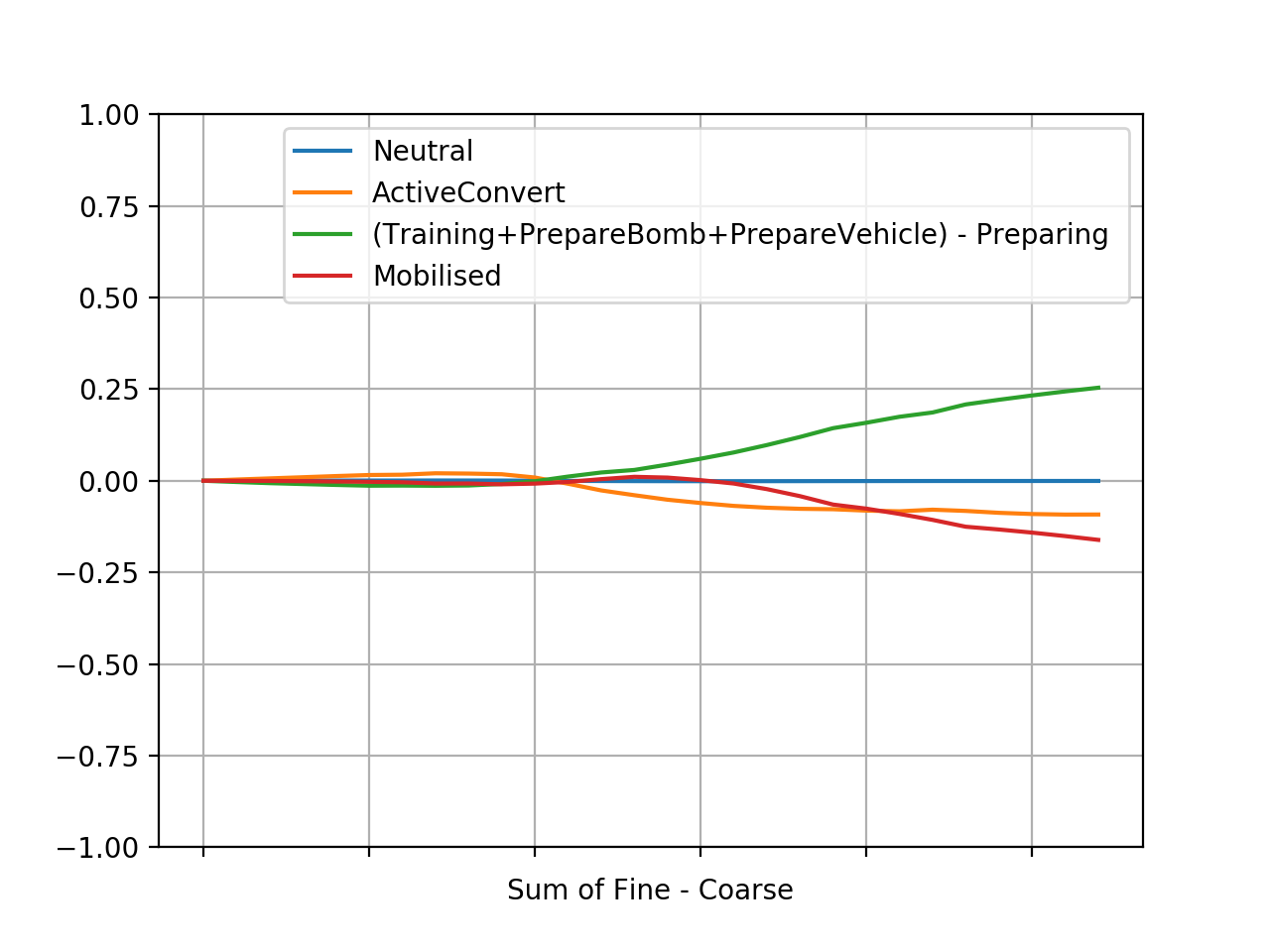}
            \caption{Refined RDCEG minus Coarse RDCEG state probabilities through time}
        \end{subfigure}        
    \caption{Analysis of state probabilities across RDCEG structure using data from Scenario A}
    \label{fine_base_coarse}
\end{figure}

We use the model as described in Section \ref{vehicle} as the base RDCEG structure, base prior state probabilities and base
holding distribution. We use Scenarios A and B from Appendix \ref{appendix_scens} to analyse the effect on the 
probability evolution through time under
changes in the RDCEG structure. The results are illustrated in Figures \ref{base3} to \ref{fine_base_coarse}
\FloatBarrier
\section{Sensitivity analysis} \label{appendix_sens}
Again we use the model as described in Section \ref{vehicle} as the base RDCEG structure, base prior state 
probabilities and base holding distribution. We use Scenarios A and B in Appendix \ref{appendix_scens} to 
analyse the effect on the probability evolution through time under changes in the state prior probabilities and changes
in the holding distribution parameter $\zeta$. Figure \ref{priorshiftcharts} and Tables \ref{priorShiftTable1} and \ref{priorShiftTable2} illustrate these impacts.
\vfill
\begin{figure}[H]
\centering
\begin{subfigure}[t]{0.475\textwidth}
    \includegraphics[width=\textwidth]{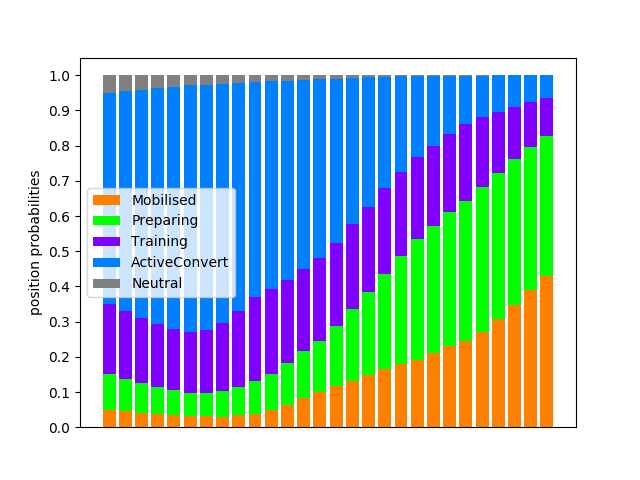}
    \caption{Original prior state probabilities}
\end{subfigure}
\hfill
\begin{subfigure}[t]{0.475\textwidth}
    \includegraphics[width=\textwidth]{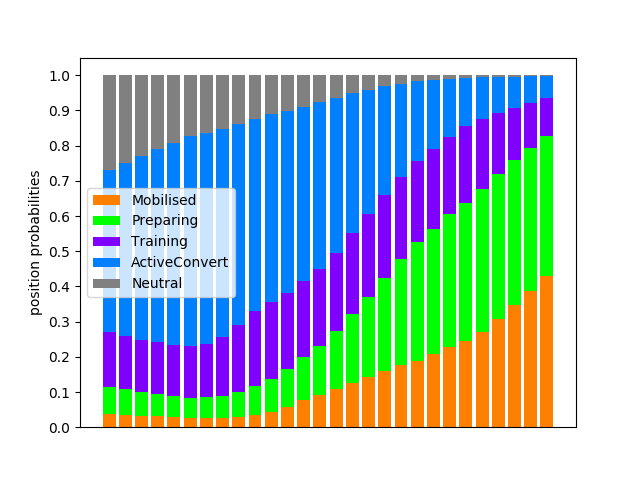}
    \caption{Increasing Neutral prior by 0.3 then renormalising}
\end{subfigure}
\centering
\begin{subfigure}[t]{0.475\textwidth}
    \includegraphics[width=\textwidth]{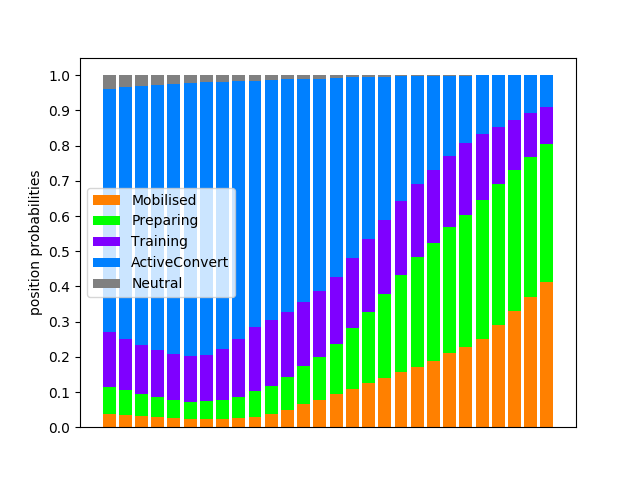}
    \caption{Increasing Active Convert prior by 0.3 then renormalising}
\end{subfigure}
\hfill
\centering
\begin{subfigure}[t]{0.475\textwidth}
    \includegraphics[width=\textwidth]{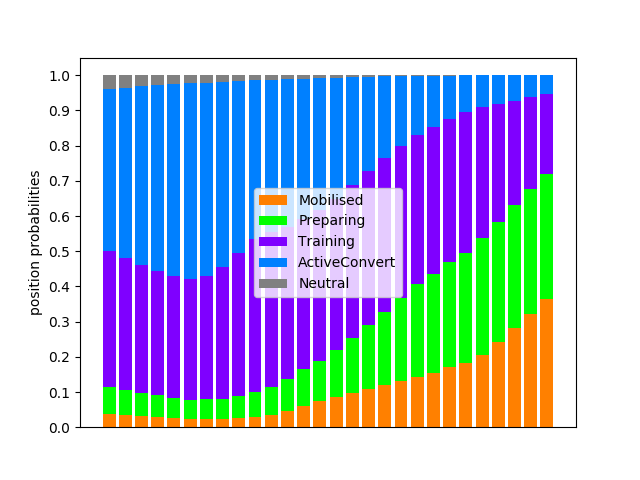}
    \caption{Increasing Training prior by 0.3 then renormalising}
\end{subfigure}
\hfill
\begin{subfigure}[t]{0.475\textwidth}
    \includegraphics[width=\textwidth]{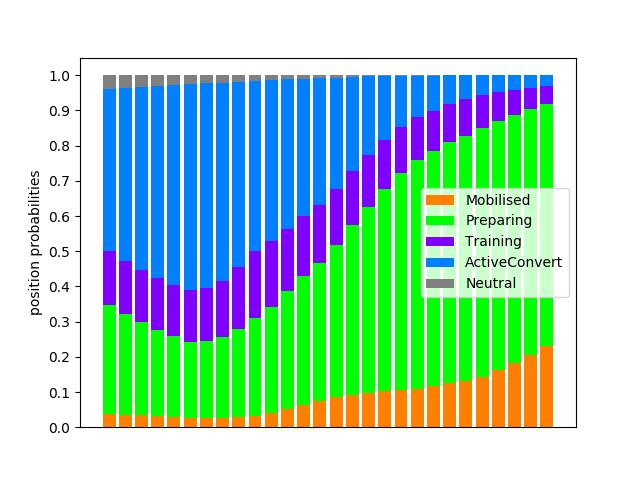}
\caption{Increasing Preparing prior by 0.3 then renormalising}
\end{subfigure}
\hfill
\centering
\begin{subfigure}[t]{0.475\textwidth}
    \includegraphics[width=\textwidth]{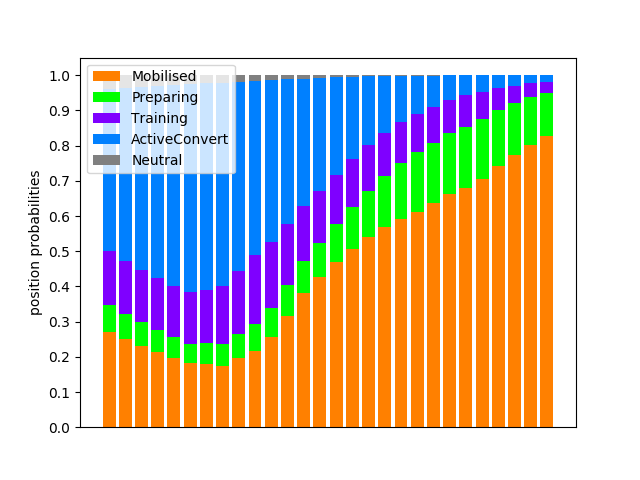}
    \caption{Increasing Mobilised prior by 0.3 then renormalising}
\end{subfigure}
\caption{Scenario A state prior probability sensitivity analysis: Moderate shifts}
\end{figure}
\begin{figure}
\ContinuedFloat
\centering
\begin{subfigure}[t]{0.475\textwidth}
    \includegraphics[width=\textwidth]{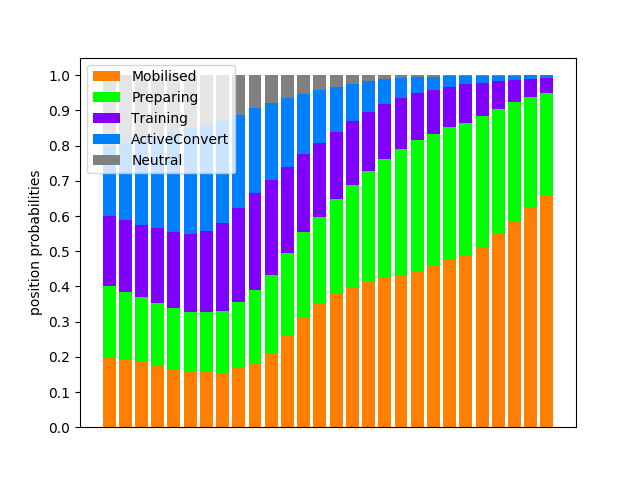}
    \caption{All state priors set equal}
\end{subfigure}
\hfill
\begin{subfigure}[t]{0.475\textwidth}
    \includegraphics[width=\textwidth]{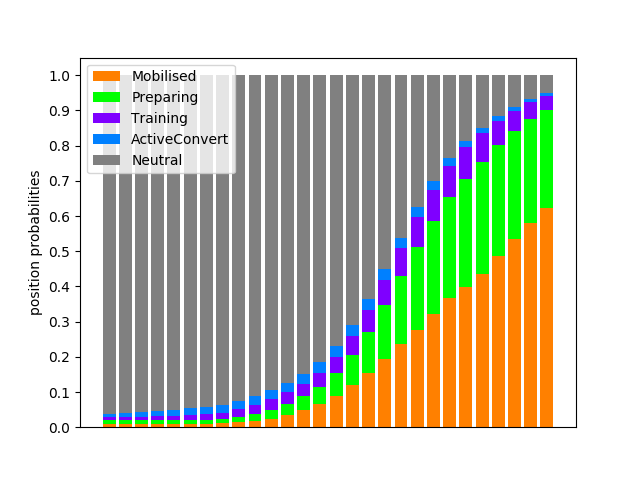}
    \caption{Increasing Neutral prior to 0.962}
\end{subfigure}
\centering
\begin{subfigure}[t]{0.475\textwidth}
    \includegraphics[width=\textwidth]{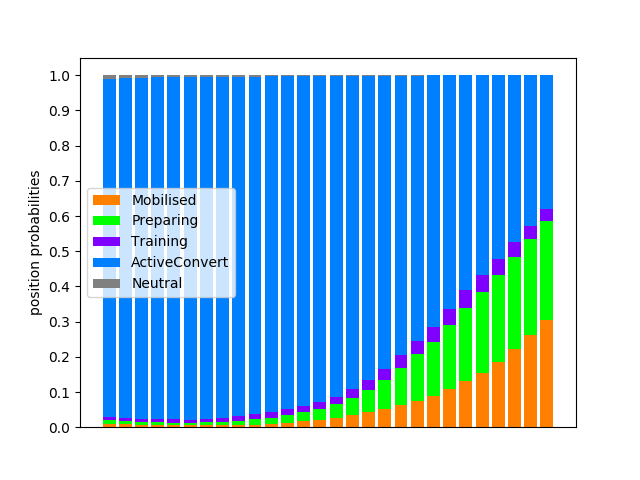}
    \caption{Increasing Active Convert prior to 0.962}
\end{subfigure}
\hfill
\centering
\begin{subfigure}[t]{0.475\textwidth}
    \includegraphics[width=\textwidth]{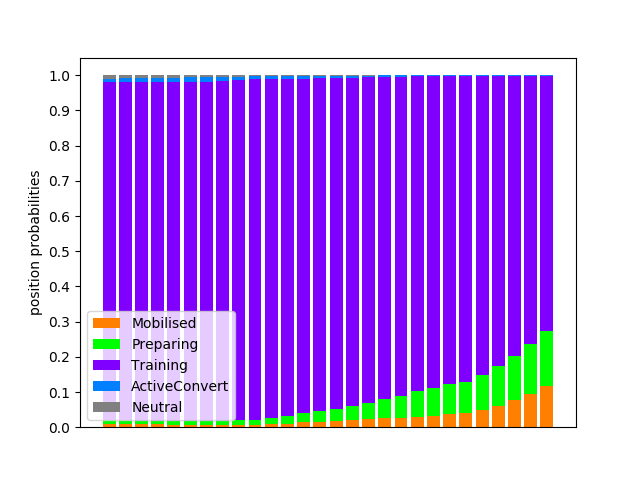}
    \caption{Increasing Training prior to 0.962}
\end{subfigure}
\hfill
\begin{subfigure}[t]{0.475\textwidth}
    \includegraphics[width=\textwidth]{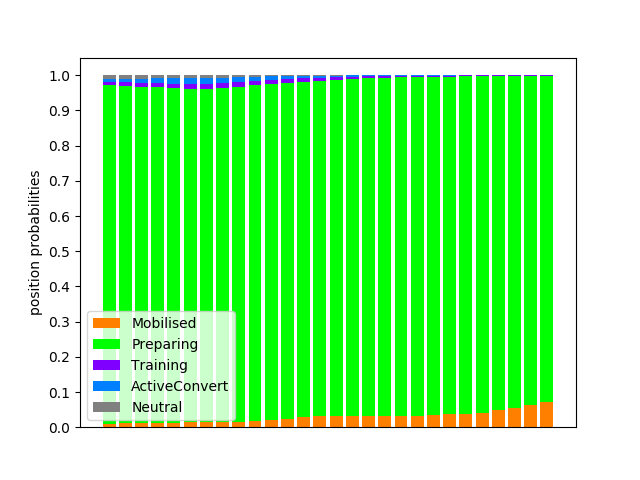}
\caption{Increasing Preparing prior to 0.962}
\end{subfigure}
\hfill
\centering
\begin{subfigure}[t]{0.475\textwidth}
    \includegraphics[width=\textwidth]{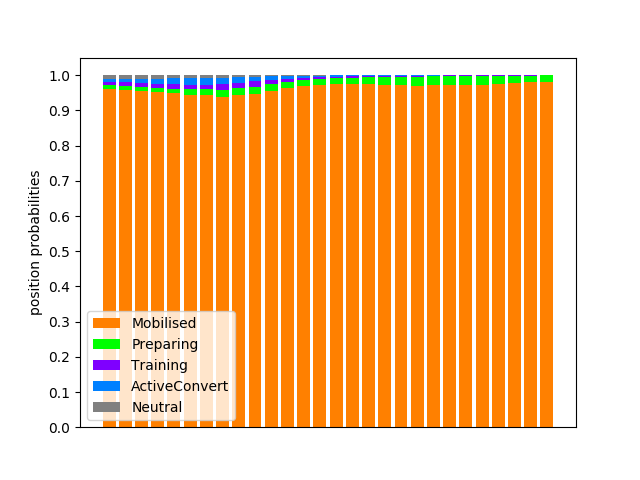}
    \caption{Increasing Mobilised prior to 0.962}
\end{subfigure}
\caption{Scenario A State prior probability sensitivity analysis: Extreme shifts}
\label{priorshiftcharts}
\end{figure}
\begin{table}
\begin{small}
\begin{tabular}{llrrrrrr}
\toprule
{} & SubScenario &  Prior &     t5 &    t10 &    t15 &    t20 &    t26 \\
State         &             &        &        &        &        &        &        \\
\midrule
Neutral       &           a &  0.050 &  0.028 &  0.016 &  0.006 &  0.001 &  0.000 \\
ActiveConvert &           a &  0.600 &  0.696 &  0.566 &  0.367 &  0.167 &  0.064 \\
Training      &           a &  0.200 &  0.178 &  0.235 &  0.242 &  0.221 &  0.108 \\
Preparing     &           a &  0.100 &  0.068 &  0.119 &  0.235 &  0.380 &  0.398 \\
Mobilised     &           a &  0.050 &  0.031 &  0.063 &  0.149 &  0.231 &  0.430 \\
Neutral       &           b &  0.269 &  0.165 &  0.100 &  0.041 &  0.010 &  0.002 \\
ActiveConvert &           b &  0.462 &  0.597 &  0.518 &  0.354 &  0.165 &  0.064 \\
Training      &           b &  0.154 &  0.153 &  0.215 &  0.234 &  0.219 &  0.108 \\
Preparing     &           b &  0.077 &  0.058 &  0.109 &  0.227 &  0.377 &  0.397 \\
Mobilised     &           b &  0.038 &  0.026 &  0.058 &  0.144 &  0.229 &  0.429 \\
Neutral       &           c &  0.038 &  0.020 &  0.012 &  0.005 &  0.001 &  0.000 \\
ActiveConvert &           c &  0.692 &  0.773 &  0.660 &  0.462 &  0.228 &  0.091 \\
Training      &           c &  0.154 &  0.133 &  0.184 &  0.205 &  0.203 &  0.105 \\
Preparing     &           c &  0.077 &  0.051 &  0.095 &  0.203 &  0.357 &  0.392 \\
Mobilised     &           c &  0.038 &  0.023 &  0.049 &  0.125 &  0.211 &  0.411 \\
Neutral       &           d &  0.038 &  0.022 &  0.012 &  0.004 &  0.001 &  0.000 \\
ActiveConvert &           d &  0.462 &  0.549 &  0.419 &  0.268 &  0.124 &  0.054 \\
Training      &           d &  0.385 &  0.350 &  0.431 &  0.438 &  0.405 &  0.226 \\
Preparing     &           d &  0.077 &  0.055 &  0.092 &  0.180 &  0.298 &  0.356 \\
Mobilised     &           d &  0.038 &  0.024 &  0.047 &  0.109 &  0.172 &  0.364 \\
Neutral       &           e &  0.038 &  0.023 &  0.012 &  0.004 &  0.001 &  0.000 \\
ActiveConvert &           e &  0.462 &  0.582 &  0.425 &  0.223 &  0.082 &  0.030 \\
Training      &           e &  0.154 &  0.149 &  0.177 &  0.147 &  0.108 &  0.051 \\
Preparing     &           e &  0.308 &  0.219 &  0.336 &  0.528 &  0.684 &  0.688 \\
Mobilised     &           e &  0.038 &  0.027 &  0.051 &  0.098 &  0.126 &  0.231 \\
Neutral       &           f &  0.038 &  0.023 &  0.011 &  0.003 &  0.001 &  0.000 \\
ActiveConvert &           f &  0.462 &  0.588 &  0.412 &  0.196 &  0.071 &  0.018 \\
Training      &           f &  0.154 &  0.151 &  0.171 &  0.129 &  0.094 &  0.031 \\
Preparing     &           f &  0.077 &  0.058 &  0.089 &  0.131 &  0.171 &  0.124 \\
Mobilised     &           f &  0.269 &  0.180 &  0.315 &  0.541 &  0.664 &  0.827 \\
\bottomrule
\end{tabular}
\caption{Posterior state probabilities over time, shown roughly every fifth time period (i.e. every 5 weeks), for moderate shifts in prior state probabilities}
\label{priorShiftTable1}
\end{small}
\end{table}
\begin{table}
\begin{small}
\begin{tabular}{llrrrrrr}
\toprule
{} & SubScenario &  Prior &     t5 &    t10 &    t15 &    t20 &    t26 \\
State         &             &        &        &        &        &        &        \\
\midrule
Neutral       &           g &  0.200 &  0.142 &  0.065 &  0.017 &  0.003 &  0.001 \\
ActiveConvert &           g &  0.200 &  0.300 &  0.197 &  0.086 &  0.029 &  0.008 \\
Training      &           g &  0.200 &  0.229 &  0.243 &  0.169 &  0.115 &  0.042 \\
Preparing     &           g &  0.200 &  0.170 &  0.236 &  0.313 &  0.376 &  0.293 \\
Mobilised     &           g &  0.200 &  0.158 &  0.260 &  0.414 &  0.477 &  0.656 \\
Neutral       &           h &  0.962 &  0.943 &  0.875 &  0.635 &  0.234 &  0.050 \\
ActiveConvert &           h &  0.010 &  0.020 &  0.026 &  0.032 &  0.022 &  0.008 \\
Training      &           h &  0.010 &  0.015 &  0.032 &  0.063 &  0.088 &  0.040 \\
Preparing     &           h &  0.010 &  0.011 &  0.032 &  0.116 &  0.289 &  0.279 \\
Mobilised     &           h &  0.010 &  0.010 &  0.035 &  0.154 &  0.366 &  0.623 \\
Neutral       &           i &  0.010 &  0.005 &  0.003 &  0.002 &  0.001 &  0.000 \\
ActiveConvert &           i &  0.962 &  0.972 &  0.945 &  0.864 &  0.665 &  0.379 \\
Training      &           i &  0.010 &  0.010 &  0.017 &  0.028 &  0.044 &  0.033 \\
Preparing     &           i &  0.010 &  0.008 &  0.021 &  0.065 &  0.180 &  0.283 \\
Mobilised     &           i &  0.010 &  0.005 &  0.013 &  0.042 &  0.110 &  0.304 \\
Neutral       &           j &  0.010 &  0.006 &  0.003 &  0.001 &  0.000 &  0.000 \\
ActiveConvert &           j &  0.010 &  0.013 &  0.008 &  0.005 &  0.002 &  0.001 \\
Training      &           j &  0.962 &  0.963 &  0.958 &  0.925 &  0.876 &  0.726 \\
Preparing     &           j &  0.010 &  0.012 &  0.022 &  0.047 &  0.084 &  0.155 \\
Mobilised     &           j &  0.010 &  0.007 &  0.010 &  0.023 &  0.037 &  0.118 \\
Neutral       &           k &  0.010 &  0.008 &  0.003 &  0.001 &  0.000 &  0.000 \\
ActiveConvert &           k &  0.010 &  0.017 &  0.008 &  0.003 &  0.001 &  0.000 \\
Training      &           k &  0.010 &  0.013 &  0.010 &  0.005 &  0.003 &  0.001 \\
Preparing     &           k &  0.962 &  0.948 &  0.956 &  0.958 &  0.959 &  0.927 \\
Mobilised     &           k &  0.010 &  0.014 &  0.023 &  0.033 &  0.037 &  0.072 \\
Neutral       &           l &  0.010 &  0.009 &  0.002 &  0.000 &  0.000 &  0.000 \\
ActiveConvert &           l &  0.010 &  0.018 &  0.007 &  0.002 &  0.001 &  0.000 \\
Training      &           l &  0.010 &  0.014 &  0.009 &  0.004 &  0.002 &  0.001 \\
Preparing     &           l &  0.010 &  0.016 &  0.018 &  0.020 &  0.025 &  0.018 \\
Mobilised     &           l &  0.962 &  0.944 &  0.963 &  0.974 &  0.972 &  0.981 \\
\bottomrule
\end{tabular}
\caption{Posterior state probabilities over time, shown roughly every fifth time period for i) all state priors set equal to 0.2 ii) each state prior set to 0.962 and the other states set close to 0.01 (rounded to three decimal places) to test behaviour with extreme priors}
\label{priorShiftTable2}
\end{small}
\end{table}
\begin{figure}[H]
\centering
\begin{subfigure}[t]{0.475\textwidth}
\includegraphics[width=\textwidth]{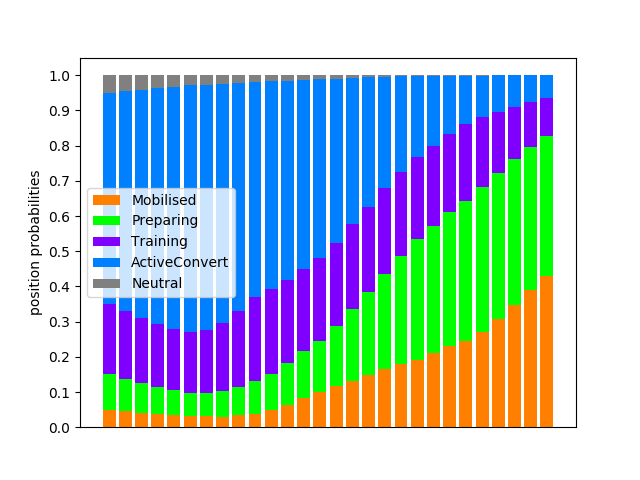}
\caption{$\zeta = 0.001$ as used in base scenario}
\end{subfigure}
\hfill
\begin{subfigure}[t]{0.475\textwidth}
\includegraphics[width=\textwidth]{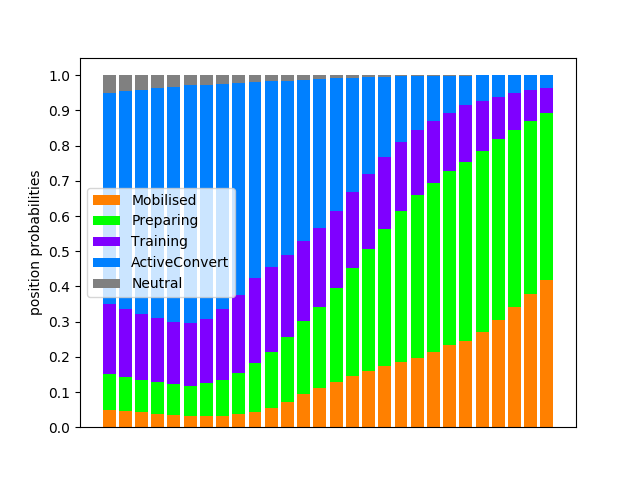}
\caption{$\zeta = 0.01$}
\end{subfigure}
\centering
\begin{subfigure}[t]{0.475\textwidth}
\includegraphics[width=\textwidth]{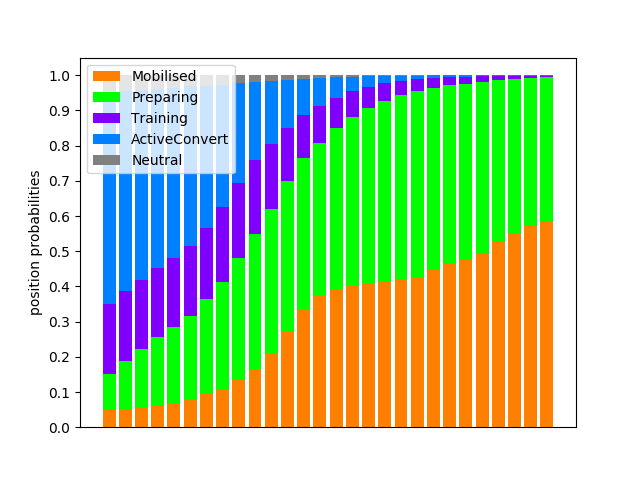}
\caption{$\zeta = 0.1$}
\end{subfigure}
\hfill
\begin{subfigure}[t]{0.475\textwidth}
\includegraphics[width=\textwidth]{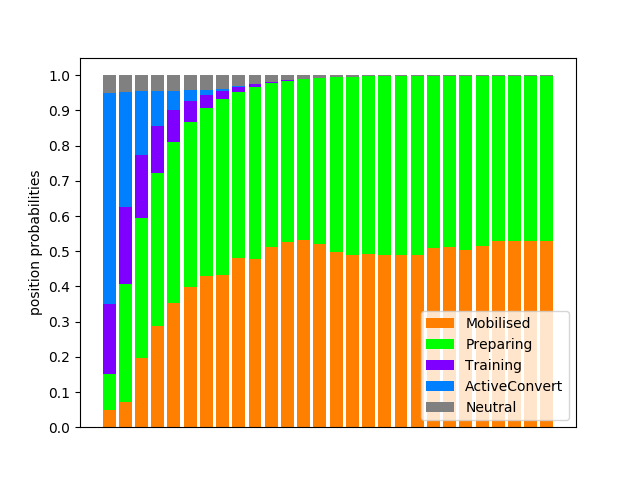}
\caption{$\zeta = 0.5$}
\end{subfigure}
\caption{Effect on Scenario A posterior probabilities through time of varying the holding time distribution $\zeta$}
\label{xishiftcharts}
\end{figure}
\begin{table}
\begin{tabular}{lllllllll}
\toprule
{} &          State &  Prior &     t0 &     t5 &    t10 &    t15 &    t20 &    t26 \\
$\zeta$    &                &        &        &        &        &        &        &        \\
\midrule
0.001 &        Neutral &   0.05 &  0.045 &  0.028 &  0.016 &  0.006 &  0.001 &  0.000 \\
   &  ActiveConvert &    0.6 &  0.625 &  0.696 &  0.566 &  0.367 &  0.167 &  0.064 \\
   &       Training &    0.2 &  0.193 &  0.178 &  0.235 &  0.242 &  0.221 &  0.108 \\
   &      Preparing &    0.1 &  0.091 &  0.068 &  0.119 &  0.235 &  0.380 &  0.398 \\
   &      Mobilised &   0.05 &  0.046 &  0.031 &  0.063 &  0.149 &  0.231 &  0.430 \\
$\zeta$    &          State &  Prior &     t0 &     t5 &    t10 &    t15 &    t20 &    t26 \\
0.01  &        Neutral &   0.05 &  0.045 &  0.028 &  0.015 &  0.005 &  0.001 &  0.000 \\
   &  ActiveConvert &    0.6 &  0.620 &  0.664 &  0.495 &  0.275 &  0.105 &  0.035 \\
   &       Training &    0.2 &  0.193 &  0.183 &  0.233 &  0.213 &  0.166 &  0.072 \\
   &      Preparing &    0.1 &  0.096 &  0.092 &  0.184 &  0.346 &  0.495 &  0.475 \\
   &      Mobilised &   0.05 &  0.046 &  0.033 &  0.072 &  0.161 &  0.233 &  0.417 \\
$\zeta$    &          State &  Prior &     t0 &     t5 &    t10 &    t15 &    t20 &    t26 \\
0.1   &        Neutral &   0.05 &  0.046 &  0.030 &  0.013 &  0.003 &  0.001 &  0.000 \\
   &  ActiveConvert &    0.6 &  0.567 &  0.404 &  0.138 &  0.030 &  0.005 &  0.001 \\
   &       Training &    0.2 &  0.198 &  0.201 &  0.150 &  0.060 &  0.023 &  0.005 \\
   &      Preparing &    0.1 &  0.138 &  0.272 &  0.427 &  0.497 &  0.505 &  0.408 \\
   &      Mobilised &   0.05 &  0.050 &  0.093 &  0.273 &  0.410 &  0.466 &  0.587 \\
$\zeta$    &          State &  Prior &     t0 &     t5 &    t10 &    t15 &    t20 &    t26 \\
0.5   &        Neutral &   0.05 &  0.048 &  0.041 &  0.014 &  0.003 &  0.001 &  0.001 \\
   &  ActiveConvert &    0.6 &  0.326 &  0.015 &  0.000 &  0.000 &  0.000 &  0.000 \\
   &       Training &    0.2 &  0.221 &  0.037 &  0.001 &  0.000 &  0.000 &  0.000 \\
   &      Preparing &    0.1 &  0.335 &  0.477 &  0.458 &  0.504 &  0.488 &  0.468 \\
   &      Mobilised &   0.05 &  0.071 &  0.429 &  0.526 &  0.492 &  0.511 &  0.530 \\
\bottomrule
\end{tabular}
\caption{Numeric data displayed to 3 decimal places shown for roughly every fifth time step varying the holding time distribution $\zeta$}
\label{xishifttable}
\end{table}
\FloatBarrier

\end{document}